%% file: neurips_2026.tex
\documentclass{article}

\usepackage[preprint]{neurips_2026}

\usepackage[utf8]{inputenc} % allow utf-8 input
\usepackage[T1]{fontenc}    % use 8-bit T1 fonts
\usepackage{hyperref}       % hyperlinks
\usepackage{url}            % simple URL typesetting
\usepackage{booktabs}       % professional-quality tables
\usepackage{amsfonts}       % blackboard math symbols
\usepackage{nicefrac}       % compact symbols for 1/2, etc.
\usepackage{microtype}      % microtypography
\usepackage{xcolor}         % colors

\newcommand{\sys}{\textsf{TrustedARI}\xspace}
\newcommand{\fsys}{$\mathcal{F}_{\textsf{sys}}$\xspace}

\newcommand{\mcpserver}{\textcolor{DarkRed}{\textsf{R}}\xspace}
\newcommand{\mcpclient}{\textcolor{DarkRed}{\textsf{A}}\xspace}
\newcommand{\toolserver}{\textcolor{DarkRed}{\textsf{S}}\xspace}
\newcommand{\shared}{\textcolor{DarkRed}{\textsf{X}}\xspace}
\newcommand{\public}{\textcolor{DarkRed}{\textsf{P}}\xspace}
\newcommand{\sid}{\mathsf{sid}}

\input{macro}

\title{\sys: Towards Trust-Native Agentic Routing Infrastructure for Agentic AI}

\author{%
  Qi Li \\
  Tsinghua University\\
  \texttt{li-q25@mails.tsinghua.edu.cn} \\
  \And
  Zhenhua Zou \\
  Tsinghua University\\
  \texttt{cbackyx@gmail.com} \\
  \And
  Shuo Li \\
  Tsinghua University\\
  \texttt{shuo-li22@mails.tsinghua.edu.cn} \\
  \And
  Mingwei Xu \\
  Tsinghua University\\
  \texttt{xumw@tsinghua.edu.cn} \\
  \And
  Zhuotao Liu \thanks{Corresponding author.} \\
  Tsinghua University \\
  \texttt{zhuotaoliu@tsinghua.edu.cn} \\
}

\begin{document}

\maketitle

\input{0.abstract}

% \input{intro_router}
\input{intro_tlsoracle}
% \input{1.introduction}
\input{2.preliminaries}

\input{3.problem}
\input{4.design}

\input{5.evaluation}
\input{6.discussion}
\input{7.related_work}
\input{8.conclusion}

\bibliographystyle{plain}
\bibliography{ref}

%%%%%%%%%%%%%%%%%%%%%%%%%%%%%%%%%%%%%%%%%%%%%%%%%%%%%%%%%%%%

\appendix

\input{appendix}

% \section{Technical appendices and supplementary material}
% Technical appendices with additional results, figures, graphs, and proofs may be submitted with the paper submission before the full submission deadline (see above). You can upload a ZIP file for videos or code, but do not upload a separate PDF file for the appendix. There is no page limit for the technical appendices. 

% Note: Think of the appendix as ``optional reading'' for reviewers. The paper must be able to stand alone without the appendix; for example, adding critical experiments that support the main claims to an appendix is inappropriate. 

% %%%%%%%%%%%%%%%%%%%%%%%%%%%%%%%%%%%%%%%%%%%%%%%%%%%%%%%%%%%%

% \newpage
% \input{checklist.tex}

\end{document}

%% file: macro.tex
\usepackage{tikz}
\usepackage{amsmath}
\usepackage{url}            % simple URL typesetting
\usepackage{booktabs}       % professional-quality tables
\usepackage{amsfonts}       % blackboard math symbols
\usepackage{bbm}
\usepackage{nicefrac}       
\usepackage{threeparttable}
\usepackage{multirow}
\usepackage{graphicx}
\usepackage{amsmath}
\usepackage{color}
\usepackage{adjustbox}
\usepackage[dvipsnames,svgnames,x11names,table]{xcolor}
\usepackage{subfigure}
\usepackage{makecell}
\usepackage{multicol}
\usepackage{listings}
\usepackage{wrapfig}
\usepackage{xspace}
\usepackage{enumitem}
\usepackage{makecell}
\usepackage{mathrsfs}

\usepackage[ruled,linesnumbered,noend,vlined]{algorithm2e} % For algorithms
\usepackage{algorithmic}

\usepackage{enumitem}
\usepackage{tikz}

\newcommand{\etal}{\hbox{{et al.}}\xspace}
\newcommand{\eg}{\hbox{{e.g.,}}\xspace}
\newcommand{\ie}{\hbox{{i.e.,}}\xspace}

\newcommand{\etc}{\hbox{{etc.}}\xspace}

\newcommand{\paraspace}{\vspace{0.01in}}
\newcommand{\parab}[1]{\paraspace\noindent{\bf #1.}}

\newcommand{\first}{\textsf{(i)}\xspace}
\newcommand{\second}{\textsf{(ii)}\xspace}
\newcommand{\third}{\textsf{(iii)}\xspace}

\newcommand{\simu}{$\mathcal{S}$\xspace}

\newcommand\revision[1]{{\color{black} #1}}
\newcommand{\anno}[1]{\textsf{\color{DarkBlue!60}{{\# #1}}}}
\newcommand{\tls}[1]{\textsf{#1}}

\usepackage{newfloat}
\usepackage{caption}

\DeclareFloatingEnvironment[fileext=frm,placement={!htbp},name=Functionality]{Functionality}
\DeclareFloatingEnvironment[fileext=frm,placement={!htbp},name=Protocol]{Protocol}
\DeclareFloatingEnvironment[fileext=frm,placement={!htbp},name=Application]{Application}
\usepackage[sets,operators]{cryptocode}

\usepackage{placeins}
\PassOptionsToClass{table, xcdraw}{xcolor}
\definecolor{lbcolor}{rgb}{0.95,0.95,0.95}
\usepackage{minted}

\usepackage[framemethod=TikZ]{mdframed}
\usepackage{lipsum}

\newdimen\linenumbersep

\newcommand{\linenumber}[1]{%
  \linenumbersep 4pt%
  \advance\linenumbersep\mdflength{innerleftmargin}%
  \advance\linenumbersep\mdflength{innerlinewidth}%
  \advance\linenumbersep\mdflength{middlelinewidth}%
  \advance\linenumbersep\mdflength{outerlinewidth}%
  \advance\linenumbersep\mdflength{linewidth}%
  \makebox[0pt][r]{{\rmfamily\tiny#1}\hspace*{\linenumbersep}}}

\usepackage{dashrule}

\mdfdefinestyle{FunctionalityFrame}{%
    linecolor=black,
    outerlinewidth=0.5pt,
    font=\normalsize,
    roundcorner=2pt,
    innertopmargin=2pt,
    innerbottommargin=2pt,
    innerrightmargin=4pt,
    innerleftmargin=3pt,
    backgroundcolor=lbcolor,
    nobreak=true
    }

\mdfdefinestyle{ProtocolFrame}{%
    linecolor=black,
    outerlinewidth=0.5pt,
    font=\normalsize,
    roundcorner=1pt,
    innertopmargin=2pt,
    innerbottommargin=2pt,
    innerrightmargin=4pt,
    innerleftmargin=3pt,
    nobreak=true
    }

\mdfdefinestyle{ApplicationFrame}{%
    linecolor=black,
    outerlinewidth=0.5pt,
    font=\normalsize,
    roundcorner=1pt,
    innertopmargin=2pt,
    innerbottommargin=2pt,
    innerrightmargin=4pt,
    innerleftmargin=3pt,
    nobreak=true
    }

\newcounter{subprot}
\renewcommand\thesubprot{\arabic{subprot}}

\usepackage{tikz}
\usetikzlibrary{matrix,arrows.meta,calc,positioning}
\usepackage{amssymb}

\definecolor{procgreen}{RGB}{0, 128, 0}
\definecolor{procblue}{RGB}{0, 139, 180}
\definecolor{procgray}{RGB}{128, 128, 128}

%% file: 0.abstract.tex
\begin{abstract}
% AI agents increasingly access external capabilities through Agentic Routing Infrastructure (ARI), yet its design introduces critical trust risks.
% Because ARI assembles and forwards invocations on the agent's behalf, it obtains plaintext access to sensitive queries and downstream responses.
% Meanwhile, agents lack a reliable mechanism to verify that ARI binds invocations to the intended endpoints or preserves invocation integrity when forwarding queries and relaying responses.
% To bypass the overhead of heterogeneous interfaces and fragmented subscriptions, 
AI agents increasingly access external models, tools, and services through Agentic Routing Infrastructure (ARI) to manage the overhead of heterogeneous interfaces and fragmented subscriptions.
Yet, the architecture of ARI introduces fundamental trust risks: it obtains plaintext access to agent queries and service responses, while leaving agents unable to verify that their queries are routed to intended service providers or that requests and responses remain untampered.
To address this problem, we present \sys, the first trust-native agentic routing infrastructure for agentic AI. 
Architecturally, \sys is built upon three core innovations: \first an ARI-adapted three-party TLS handshake that enables the agent and ARI to jointly authenticate the service provider through role-specific distribution of TLS key materials; \second a privacy-preserving query-construction protocol that allows the agent and ARI to collaboratively construct well-formed queries without exposing their respective private inputs;
% \second a privacy-preserving query construction protocol that allows the agent and ARI to collectively compute well-formed requests without disclosing each other's private inputs, 
% eliminates data leakage to ARI while ensuring end-to-end confidentiality and integrity of service invocations; 
% and \third a billing-compatible response integrity protocol that achieves verifiable billing in spite of such confidentiality guarantees. 
and \third a verifiable billing protocol that supports fair usage-based settlement while preserving the integrity and confidentiality of service responses.
%simultaneously achieves response integrity/confidentiality and fair billing. 

We implemented and extensively evaluated a prototype of \sys to validate its performance. 
Experiments confirm that \sys is highly efficient: our ARI-adapted handshake protocol reduces 
% connection setup latency by up to 49.81\% and 
communication overhead by 39.34\% compared to the existing three-party TLS handshake.
Furthermore, the privacy-preserving query-construction protocol imposes negligible overhead--averaging 0.19 seconds in computation time and 0.58 MB in communication costs--while the verifiable billing protocol speeds up proof generation by 28.20$\times$.
% by reducing cryptographic constraints by 33.26$\times$.
% while the billing-compatible response-integrity protocol achieves a $28.20\times$ speedup in proof generation while reducing the number of circuit constraints by $33.26\times$.
% Crucially, \sys is readily deployable through a lightweight agent-side compatibility layer, enabling integration without model retraining, tuning, or downstream service modifications.
Crucially, \sys is readily deployable without any modification to the service providers.
\end{abstract}

%% file: intro_tlsoracle.tex
% \newpage
\section{Introduction}
Artificial Intelligence (AI) agents are increasingly deployed as autonomous systems that execute complex, multi-step workflows via external models, tools, and services~\cite{wang2024survey}.
% However, scaling direct integrations between agents and individual service providers introduces severe deployment barriers across technical and economic dimensions. Technically, 
However, direct integration between agents and individual service providers imposes non-trivial deployment overhead, such as 
% is constrained by 
highly heterogeneous interfaces, long-tailed API maintenance~\cite{mo2025livemcpbench}, and regional network or policy restrictions. Economically, fragmented access to diverse models and tools imposes significant subscription costs on individual users for occasional task execution. 
To avoid these engineering frictions, % To jointly mitigate these hurdles, 
% AI ecosystems introduce \emph{Agentic Routing Infrastructure} (ARI) as an important intermediate layer, 
\revision{emerging AI ecosystems increasingly rely on an intermediate layer that we refer to as \emph{Agentic Routing Infrastructure} (ARI),}
which primarily manifests in two paradigms: \emph{LLM API Routing} (\eg OpenRouter~\cite{openrouter_2023}, LiteLLM~\cite{litellm}) to facilitate model invocation and \emph{Agent Tool Routing} (\eg Pipedream~\cite{Pipedream},  Zapier~\cite{Zapier}) \revision{for connecting various web services through the unified interfaces such as 
Model Context Protocol (MCP)~\cite{mcp_document}}. 
% A technical instantiation of the latter paradigm is Anthropic's Model Context Protocol (MCP) Server~\cite{mcp_document}.

Architecturally, ARI functions as an intermediary infrastructure positioned between the agent and service providers to abstract downstream service interfaces, assemble and forward client requests, and establish a consolidated, usage-based billing pipeline.
In a typical deployment, ARI holds the service-access metadata (\eg API keys, prepaid credits) required to invoke these external service capabilities and charges agents for mediated calls. 
% Given a target service (\eg an LLM endpoint or a web tool), the agent sends the request to ARI, which translates it into a concrete downstream API call, forwards the invocation, receives the downstream response, and relays it back to the agent.
Given a target service (\eg an LLM service or a web tool), the agent sends the request to ARI, which translates it into the service-specific invocation request and forwards it to the corresponding service provider.
After receiving the provider's result, ARI processes and returns the response to the agent.
% If the target service is bound to a user-owned account, the user should additionally delegate the corresponding account credentials to ARI. 
% Consequently, ARI becomes an intermediary that observes the agent’s plaintext request and response data, while preventing the agent from directly verifying the downstream execution.

\begin{figure}[t]
    \centering
    \includegraphics[width=0.65\columnwidth]{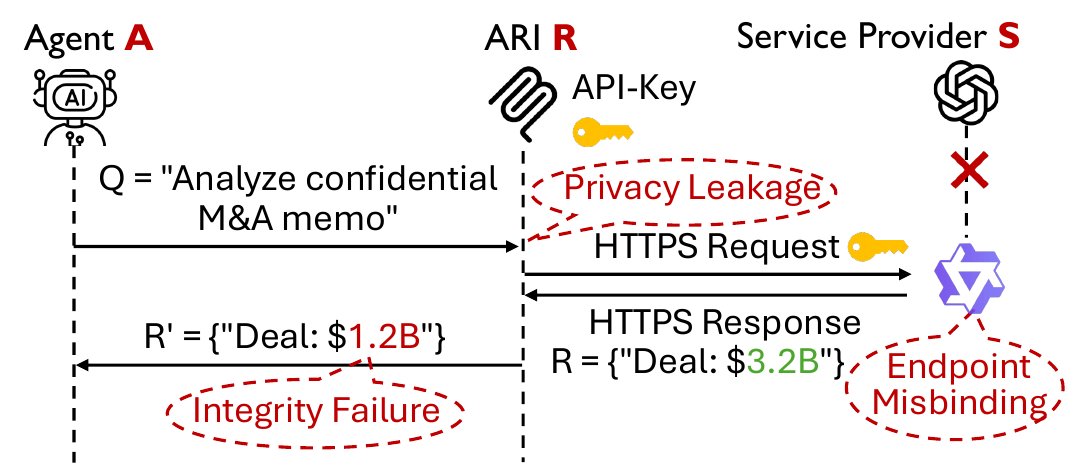}
    \caption{The architecture of ARI and its security risks.}
    \label{fig:router}
\end{figure}

\begin{figure}[t]
    \centering
    \includegraphics[width=0.65\columnwidth]{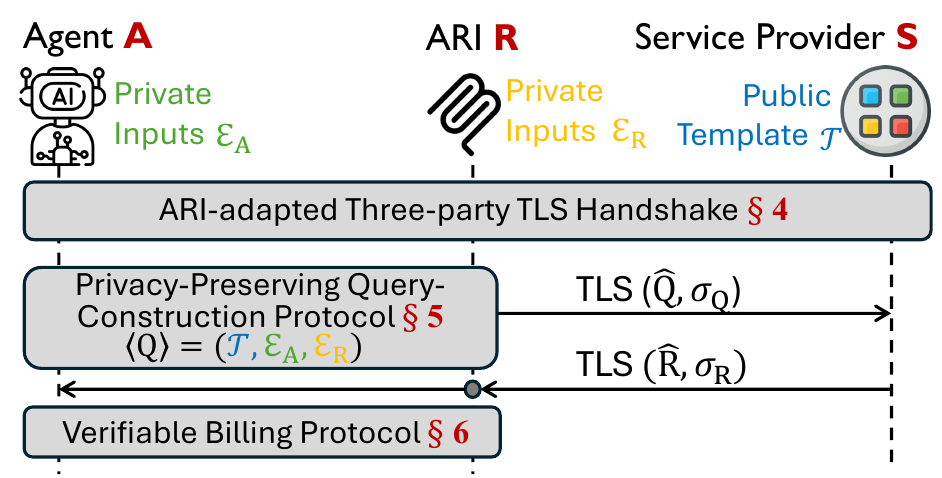}
    \caption{\revision{Overview of \sys.}}
    \label{fig:trustrouter}
\end{figure}

Despite this convenience, ARI introduces two fundamental trust issues: privacy leakage~\cite{zhao2025mcp,yao2025intentminer} and the lack of end-to-end integrity guarantees~\cite{li2025we}. 
% First, the routing workflow exposes both request and response payloads to the intermediary. 
% During a request, the router receives the agent's query, attaches router-held authentication material (\eg an API key), and forwards the resulting invocation to the downstream LLM.
% Upon receiving the response, the router reformats the LLM's raw output into a unified interface. Consequently, both the query and response are fully exposed to the router in application-layer plaintext.
\revision{Figure~\ref{fig:router} illustrates these risks using LLM API Routing as an example. 
% ARI receives the complete agent request and issues a separate invocation request to the selected LLM using ARI-held credentials (\eg an API key).
% As a result, both the agent's query and the LLM's response are exposed to ARI in application-layer plaintext, allowing the ARI to observe sensitive task inputs and model outputs.
ARI receives the complete agent request, constructs a separate LLM-facing request using ARI-held credentials, and relays the LLM's response. Thus, ARI observes both the agent query and the model output in plaintext.}
% Additionally, the agent lacks any mechanism to verify the integrity of the received responses.
% For instance, the agent cannot detect endpoint misbinding (\eg routing requests to a cheaper or lower-tier LLM while charging for a premium one) or guarantee result integrity, such as whether the returned response and billing-relevant metadata (\eg token usage) are faithfully derived from the actual downstream execution. 
Second, the agent lacks a reliable mechanism to verify the service integrity.
% the issued query and the received response.
% Specifically, the agent cannot ensure that ARI invokes the intended LLM, \ie a request intended for a premium LLM may be routed to a lower-tier model.
\revision{On the one hand, the agent cannot verify that ARI invokes the intended LLM provider, \ie a request intended for a premium model may be forwarded to a different provider offering lower-tier model services.}
On the other hand, the agent also cannot confirm that the query is forwarded unchanged or that the response is returned unchanged by the ARI.
% For instance, the agent cannot determine whether the router actually invoked the specified LLM endpoint, whether it silently substituted a cheaper or lower-tier model, or whether the returned content was modified before delivery.
% Notably, these fundamental privacy and integrity deficiencies are identically persistent within the \emph{Agent Tool Routing} paradigm.
The same privacy and integrity issues persist in the Agent Tool Routing paradigm, where the downstream endpoints are general web services instead of LLMs.
To address these problems, we introduce \sys, a trust-native agentic routing infrastructure for agentic AI.
% we introduce \sys, a privacy-preserving and verifiable routing protocol for agentic AI.
% The key insight of \sys is to decouple the ARI’s control-plane role from plaintext access in the data plane.
The key idea of \sys is to decouple ARI's participation in constructing and relaying a service request from its capability to observe or unilaterally control the invocation.
% To this end, as shown in Figure~\ref{fig:trustrouter}, \sys lets the agent and ARI jointly emulate a ``virtual TLS client'' via secure multiparty computation to establish a standard TLS connection to a downstream service provider. This three-party TLS session ensures that the agent's request is routed to the intended provider and that ARI cannot tamper with the agent's request and the service's response. 
% To this end, as shown in Figure~\ref{fig:trustrouter}, \sys enables the agent and the ARI to jointly execute the standard Transport Layer Security (TLS) protocol with the downstream service provider, using secure multiparty computation to act as a distributed ``virtual TLS client.''
\revision{To this end, \sys leverages secure multi-party computation to enable the agent and the ARI to execute jointly as a ``virtual client'', establishing a standard Transport Layer Security (TLS) session with the service provider, as illustrated in Figure~\ref{fig:trustrouter}. 
This joint execution securely encrypts the jointly-computed requests to guarantee privacy, while yielding cryptographic evidence directly tied to the downstream service to ensure end-to-end integrity.}
% Meanwhile, a privacy-preserving query protocol enables the agent and ARI to jointly construct service request without leaking secret information to each other. %  Crucially, \sys ensures verifiable billing in spite of such privacy assurance. 
% In spite of such integrity and privacy assurance, \sys remains compatible with the usage-based billing model adopted by most ARI.
The design of \sys draws inspiration from TLS-oracles~\cite{zhang2020deco, lauinger2025janus, xie2024lightweight, celi2023distefano, tan2023mpcauth}.
However, directly applying existing TLS-oracles to the ARI setting is infeasible, as the unique trust model and operational requirements of agentic routing introduce three fundamental challenges that \sys resolves with dedicated protocol designs.

% First, unlike existing TLS oracles where one party observes the complete traffic payloads, \sys must enforce distinct disclosure semantics: both the agent and ARI contribute private information to the query, but the agent requires the downstream identity and complete response, while ARI requires only billing-relevant data, such as the ``\texttt{token\_usage}'' field in an LLM API response. 
\revision{First, unlike prior TLS-oracle settings, where one party provides private inputs and receives the plaintext payloads, the ARI setting has a fundamentally different %role-specific 
interaction semantic. First, both the agent and ARI contribute private information to the query. Second, only one party (\ie the agent) is allowed to received the authenticated and encrypted responses, while the other party (\ie the ARI) is interested to verify certain billing-relevant data from the encrypted response.}  
% yet only the agent validate the identities of service providers and receive complete, untampered responses, whereas ARI learns only billing-relevant fields from the response, such as the \texttt{token\_usage} field in LLM API response.}
% To meet this asymmetric trust model, \sys introduces an \textit{ARI-adapted Three-party Handshake} (\S~\ref{sec:handshake}), strategically distributing key materials across the protocol stages. 
\revision{To realize this interaction semantic, \sys introduces an \textit{ARI-adapted Three-party Handshake} (\S~\ref{sec:handshake}) protocol, which distributes key material across protocol stages to enforce each party's intended view.} 
Handshake secrets are released to the agent so it can independently verify the identities of service providers to prevent mis-routing. The client application key is then secret-shared between both parties, allowing them to collaboratively construct encrypted requests. 
% Finally, the server application key is initially shared to bind ARI to the session, and later revealed to the agent to enable full response decryption and proof of billing-related data. 
The server application key is first derived in secret-shared form to bind ARI to the session, and ARI later releases its share to the agent, enabling the agent to decrypt the full response while limiting ARI's visibility only to billing-related fields.
% disclose only the required billing-related fields.
This dedicated key distribution not only fulfills the aforementioned interaction sematic, 
% \sys's privacy and integrity guarantees, 
but also eliminates much of the expensive multiparty computation and zero-knowledge proofs required by existing TLS oracles~\cite{celi2023distefano}.

Second, ARI-mediated requests encompass highly fragmented private inputs from both the agent (\eg user prompts) and ARI (\eg API keys), making it challenging to securely assemble a valid request without revealing any content-level or structure-level information.
To address this problem, \sys designs a \textit{privacy-preserving query-construction protocol} (\S~\ref{sec:query}) that models ARI-mediated invocation as secure template instantiation. A public \emph{template}, agreed upon by the agent and ARI, defines the request syntax and the ownership of different fields in the request. 
During template instantiation, the agent and ARI privately compute and assemble the ownership-tagged segments from their respective private inputs, while hiding both private field values and structural information (\ie the length of a field in the query).
Finally, leveraging the secret-shared TLS client-side application key, the agent and ARI derive the TLS-encrypted records of the assembled and well-structured requests that can be natively processed by the service provider.

% Third, ARI's usage-based settlement is incompatible with the confidentiality and integrity of service responses. 
\revision{Third, ARI's usage-based settlement becomes non-trivial given the confidentiality and integrity guarantees of responses.}
Specifically, under our key schedule, the agent obtains the complete TLS-authenticated response, while ARI observes only encrypted response records.
Consequently, ARI cannot directly inspect billing-relevant fields (\eg the \texttt{token\_usage} field) embedded in the response, breaking its usage-based settlement model.
To support fair settlement under the confidentiality/integrity guarantees,
\sys designs a \textit{verifiable billing protocol} (\S~\ref{sec:response}).
The protocol requires the agent to report the pre-declared billing-relevant fields and provide a zero-knowledge proof that these values are faithfully extracted from the TLS-authenticated response.
% \sys substantially reduces proving cost by implementing ZKP-friendly circuits that only prove the response-packet prefix or suffix containing the target fields, rather than generating proof over the entire packet~\cite{zhang2020deco, angel2026coral}.
\sys substantially reduces proving cost by designing a ZKP-friendly circuit that attests only a small response window containing the billing fields, rather than the entire response packet~\cite{zhang2020deco,angel2026coral}.

\parab{Contribution} The main contribution of this paper is the design, implementation and evaluation of \sys, 
the first trust-native agentic routing infrastructure for agentic AI.
We provide formal specifications of our protocols with rigorous security proofs. 
We implement a prototype of \sys in approximately 9000 lines of C++ and 3000 lines of Go code, and perform extensive evaluations. 
Our evaluation confirms that \sys is highly efficient compared to baselines constructed from existing approaches~\cite{celi2023distefano, zhang2020deco, grubbs2022zero, angel2026coral}.  
Specifically, the ARI-adapted handshake reduces the connection setup latency to service providers by up to 50.47\% compared to baselines, while cutting communication overhead by 39.34\%. 
Meanwhile, the privacy-preserving query-construction protocol introduces minimal overhead, adding only 0.19 seconds (14.29\%) in computation time and 0.58 MB (1.36\%) in communication costs. 
Furthermore, the verifiable billing protocol achieves a $28.20\times$ speedup in proof generation with a $33.26\times$ reduction in circuit constraints.
% Furthermore, the verifiable billing protocol achieves a $28.20\times$ speedup in proof generation while reducing the number of circuit constraints by $33.26\times$.
% Crucially, \sys is readily deployable as the common production LLMs can natively generate \sys-compatible service invocations without any model training or tuning.
Crucially, \sys is readily deployable without any modification to the service providers.
% through a lightweight agent-side compatibility layer that translates existing LLM API requests and standard tool calls into \sys-compatible templates, enabling integration without 

%% file: 2.preliminaries.tex
\section{Preliminaries}
% \section{Background and Cryptographic Primitives}
\label{sec:background}

\parab{Transport Layer Security (TLS) 1.3}
\revision{As the foundation for secure communication, TLS ensures the authentication of service providers and the confidentiality and integrity of data transmission.} We focus on TLS 1.3~\cite{rfc8446} because its cryptographic design—also adopted by the now-ubiquitous QUIC protocol~\cite{rfc9000}—has been reported to secure over 96\% of web traffic~\cite{tlsadoption}. Structurally, TLS 1.3 establishes keys via a handshake layer and protects application data using an AEAD scheme $\mathcal{F}_\textsf{AEAD}$ in the record layer. We adopt the TLS 1.3 notation from Dowling \etal~\cite{dowling2021cryptographic} in the paper.

% \subsection{Internet-Native Payments: x402 and Escrow Scheme}
% The x402 protocol~\cite{x402_escrow_839} is an emerging internet-native payment standard designed for autonomous Machine-to-Machine (M2M) transactions. To support usage-based services where exact costs are unknown a priori (\eg LLM token consumption), x402 introduces a pre-funded settlement model. By leveraging specialized escrow smart contracts such as TrustEngine~\cite{trustengine_sol}, the protocol enables secure fund locking and atomic reconciliation based on post-execution usage evidence, providing a trustless and decentralized financial foundation for agentic AI.
% \parab{Internet-Native Payments: x402 and Escrow Scheme}
% The x402 protocol~\cite{x402_escrow_839} is an emerging internet-native payment standard designed for autonomous Machine-to-Machine (M2M) transactions via a pre-funded settlement model. Utilizing escrow smart contracts like TrustEngine~\cite{trustengine_sol}, it enables secure fund locking and atomic reconciliation based on post-execution usage evidence. This provides a decentralized, trustless financial foundation for usage-based agentic AI.

% \subsection{Cryptographic Primitives}
% \section{Cryptographic Primitives}

\parab{Secure Multiparty Computation (MPC) on Secret-Shared Data} \sys builds on standard two-party computation (2PC) primitives, where intermediate values are represented as secret shares held by the agent and the ARI. 
% We use additive secret sharing over the ring $\mathbb{Z}_{2^k}$: for a integer $x \in \mathbb{Z}_{2^k}$, the agent holds $\langle x \rangle_0$ and the MCP server holds $\langle x \rangle_1$ such that $\langle x \rangle_0 + \langle x \rangle_1 = x$ (mod $2^k$). We also use bitwise (XOR) sharing for $k$-bit strings: for $x \in \{0,1\}^k$, the parties hold  $\langle x \rangle_0$, $\langle x \rangle_1 \in\{0,1\}^k$ (both bitwidth $k$) such that $\langle x \rangle_0 \oplus \langle x \rangle_1 = x$, where $x$ also has bitwidth $k$. 
We employ additive secret sharing over $\mathbb{Z}_{2^k}$, where $x = \langle x \rangle_0 + \langle x \rangle_1 \pmod{2^k}$, and bitwise (XOR) sharing for $k$-bit strings, where $x = \langle x \rangle_0 \oplus \langle x \rangle_1$. 
% We leverage share conversion primitives to transition between these domains, specifically $\mathcal{F}_{\textsf{A2B}}$ for arithmetic-to-bitwise and $\mathcal{F}_{\textsf{B2A}}$ for bitwise-to-arithmetic conversions.
Share conversion primitives, $\mathcal{F}_{\textsf{A2B}}$ and $\mathcal{F}_{\textsf{B2A}}$, facilitate transitions between these domains. 
% When unambiguous, we write $\langle x \rangle$ to denote that $x$ is secret-shared between the two parties. 
% For clarity in the context of the TLS handshake, we occasionally denote the shares held by the agent (\mcpclient) and the MCP server (\mcpserver) as $\textsf{x}^\textsf{c}$ and $\textsf{x}^\textsf{s}$, respectively, particularly when $\textsf{x}$ represents session keys or intermediate cryptographic secrets.
\revision{We denote shared values as $\langle x \rangle$, or as $x^{\mathsf{A}}$ and $x^{\mathsf{R}}$ for shares held by the agent and ARI when describing TLS cryptographic secrets.} 
% Our MPC backend supports standard arithmetic over additive shares (e.g., addition and multiplication) and common Boolean/control primitives (\eg comparison and the two-way multiplexer($\mathcal{F}_{\textsf{MUX}_2}$)~\cite{rathee2020cryptflow2}).
Our backend supports standard arithmetic, Boolean, and control-flow primitives (\eg comparison and two-way multiplexer ($\mathcal{F}_{\textsf{MUX}_2}$)~\cite{rathee2020cryptflow2}).

\parab{Garbled Circuit (GC)} 
% For Boolean-heavy subroutines, we use Yao’s garbled circuits (GC)~\cite{yao1982protocols, yao1986generate}, which are best-suited for bitwise operations. GC protocols are naturally asymmetric: the garbler constructs the garbled circuit, and the evaluator executes it. To preserve composability with our MPC design, we conceptually apply an output-masking technique so that the circuit evaluation yields secret-shared outputs: the evaluator supplies a random mask $r$, the garbled circuit outputs $y + r$ mod $2^k$ (or $y \oplus r$ for bitwise outputs), and the two parties interpret the resulting values as additive (or XOR) shares of $y$. For simplicity, we treat GC subroutines as taking secret-shared inputs and producing secret-shared outputs, and omit this masking detail in the protocol description.
For bitwise-heavy operations, we employ Yao’s garbled circuits~\cite{yao1982protocols, yao1986generate}. To ensure composability with the secret-sharing framework, we adopt an output-masking technique: the evaluator provides a random mask $r$, and the circuit computes $y \oplus r$ (or $y + r \pmod{2^k}$). The parties thus obtain XOR (or additive) shares of the actual result $y$ 
% without the garbler learning the plaintext. 
without revealing the plaintext output to either party.
% For brevity, we treat GC as a modular primitive that natively consumes and produces secret shares.
For brevity, we treat GC as taking secret-shared inputs and producing secret-shared outputs, and omit this masking detail in the protocol description.

\parab{Oblivious Transfers (OT)} We use 1-out-of-2 oblivious transfer as a core building block. In OT, the sender holds $(m_0, m_1)$, the receiver chooses $b \in \{0, 1\}$ and learns $m_b$, while the sender learns nothing about $b$ and the receiver learns nothing about $m_{1-b}$. To reduce cost, we use IKNP OT extension~\cite{ishai2003extending} to amortize a small number of base OTs into many efficient OTs.

\parab{Zero-knowledge Proofs} 
% We use zero-knowledge arguments of knowledge (ZK-AoK) to prove the correctness of selected statements without revealing witnesses. Formally, an argument of knowledge for an NP relation $\mathcal{R}$ is a protocol between a computationally-bounded prover $\mathcal{P}$ and a verifier $\mathcal{V}$. At the end of the protocol, $\mathcal{V}$ is convinced by $\mathcal{P}$ that there exists a witness $w$ such that $(x;w)\in \mathcal{R}$ for some input $x$, without disclosing $w$. The protocol satisfies the standard properties of (i) correctness, (ii) knowledge soundness, and (iii) zero knowledge. We focus on arguments of knowledge which have the stronger property that if the prover convinces the verifier of the statement's validity, then the prover must know $w$. To minimize communication overhead, we adopt Non-Interactive Zero-Knowledge (NIZK) proofs, which allow the prover to convince the verifier in a single message. 
We employ zero-knowledge arguments of knowledge (ZK-AoK) to prove statements regarding an NP relation $\mathcal{R}$ without revealing the witness $w$. Formally, for a public input $x$ and witness $w$ such that $(x, w) \in \mathcal{R}$, a prover $\mathcal{P}$ convinces a verifier $\mathcal{V}$ of the statement's validity. The protocol satisfies completeness, zero-knowledge, and knowledge soundness, which ensures that any prover accepted with non-negligible probability can be used to extract a valid witness.
% which ensures that a successful $\mathcal{P}$ must actually possess the witness $w$. 
See formal definitions in Appendix~\ref{sec:zkp}.
% In \sys, ZK proofs support the verifiable billing protocol by attesting that a small, tool-specific set of billing-relevant fields is correctly extracted from the decrypted response, without opening the full response. 
In our system, we instantiate the proving system with Plonk~\cite{gabizon2019plonk}, a widely adopted NIZK construction.

%% file: 3.problem.tex
\section{Problem Statement}
\label{sec:problem}

In this section, we introduce the ideal functionality of \sys (i.e., \fsys, as shown in Functionality~\ref{func:fsys}) to formally capture its input setting, computation goal, threat model, and privacy guarantee.

There are three participating entities: the agent (\mcpclient), the ARI (\mcpserver), and the downstream service provider (\toolserver). 
Prior to protocol execution, the agent and ARI agree on a designated service provider
and invocation specification, which together fix
% , which fixes
a public query template $\mathcal{T}$ and a billing-related response field selector $\varphi$. 
Structurally, $\mathcal{T}$ defines the public skeleton layout, field formatting constraints, and padding boundaries of the query request.
Formally, the public query template is defined as a sequence of segment descriptors $\mathcal{T}=[\tau_1,\ldots,\tau_L]$, where each descriptor $\tau_i=(\theta_i,h_i)$ specifies the segment category and ownership $\theta_i$ and its public padded length $h_i$.
The template $\mathcal{T}$ is public to both parties in advance, whereas the specific content populated within the variable slots represents private inputs.
Under this setting, the agent and ARI provide their respective indexed entry sets
(\ie field contents) $\mathcal{E}_{\textsf{A}}$ and $\mathcal{E}_{\textsf{R}}$.
Each entry is a pair $(i,\tilde{s},g)$, where $i$ identifies the segment position $\tau_i$ in the template and $\tilde{s}$ is the content string padded to the public length $h_i$, and $g$ is the true content length.
% The padding hides the true length of the underlying content before assembly.
The padding hides the true length of each party's private content from the other party during secure assembly.
The indices that appear in $\mathcal{E}_{\textsf{A}}$ or $\mathcal{E}_{\textsf{R}}$ are determined by the segment ownership encoded in $\theta_i$.
% These entries correspond to query parameters or real-world credentials,
These entries correspond to agent-side query parameters or ARI-side service-access metadata,
depending on the downstream service specification.
The service provider is modeled as a standard TLS-protected external service; % and therefore we omit its actions.
we only model its TLS transcript interface and omit its service-specific internal logic.

The lifecycle of \fsys proceeds sequentially across three phases: \emph{handshake}, \emph{query}, and \emph{response}. We use $\sid$ to denote a session identifier that links messages across these phases within the same service invocation.
At the handshake stage, ARI initiates the session by submitting a proposed binding $(\sid,\mathsf{id}_{\toolserver})$ to \fsys, where $\sid$ identifies the invocation instance and $\mathsf{id}_{\toolserver}$ denotes the target service provider. \fsys forwards this proposed binding to the agent for independent validation. 
The agent checks whether $\mathsf{id}_{\toolserver}$ matches the service provider agreed upon with ARI. 
If the check fails, \fsys outputs $\bot$ and aborts; otherwise, \fsys records the binding $(\sid,\mathsf{id}_{\toolserver})$ for subsequent query and response processing.

At the query stage, after receiving $\mathcal{E}_{\textsf{A}}$ from the agent and $\mathcal{E}_{\textsf{R}}$ from ARI, the procedure
% $\textsf{Assemble}$ instantiates the public template by removing the padding from each indexed padded content $\tilde{s}$ and placing the recovered content into the segment indexed by $i$
% $\textsf{Assemble}$ instantiates the template by privately unpadding each indexed content $\tilde{s}$ and placing the recovered value into segment $i$, producing the plaintext query $Q$.
$\textsf{Assemble}$ instantiates the template by truncating each padded content $\tilde{s}$ to its true length $g$ and placing the recovered value into segment $i$, producing the plaintext query $Q$.
Then, \fsys computes the TLS-authenticated query transcript $(\sid,\hat{Q},\sigma_Q)$ for $Q$ and outputs it to both the agent and ARI. Finally, the ARI forwards $(\sid,\hat{Q},\sigma_Q)$ to the service provider.

At the response stage, the service provider returns a TLS response transcript $(\sid,\hat{R},\sigma_R)$. Then, \fsys validates its correctness. If the validation fails, \fsys outputs $\bot$ and aborts. Otherwise, \fsys forwards $(\sid,\hat{R},\sigma_R)$ to both parties and outputs the plaintext response $(\sid, R)$ only to the agent. Finally, when the agent submits the billing-relevant value $v$, \fsys checks whether $R[\varphi]=v$. If the check passes, \fsys outputs $(v,1)$ to the ARI; otherwise, it outputs $(v,0)$.

\begin{Functionality}[!t]
  \begin{mdframed}[style=FunctionalityFrame, 
        align=center, 
        userdefinedwidth=\linewidth,
        innertopmargin=1pt,
        innerbottommargin=1pt,
        ]
      \procedureblock[
      mode=text,
      headlinesep=0pt,
      bodylinesep=0.2\baselineskip
      ]
      {\vspace*{-3pt}\centering \textbf{Functionality } \fsys interacts with \mcpclient, \mcpserver and \toolserver}{%
          \begin{minipage}{\linewidth}
          \raggedright
          % \vspace{0.1\baselineskip}
          % --- 0. INPUT ---
          \noindent\textbf{Input.}
          % \mcpclient holds private field contents $\{\mathcal{E}_{\textsf{A}}\}$.
          % \mcpserver holds private field contents $\{\mathcal{E}_{\textsf{R}}\}$.
          \mcpclient and \mcpserver hold their respective indexed padded field contents $\mathcal{E}_{\textsf{A}}$ and $\mathcal{E}_{\textsf{R}}$.
          \mcpclient and \mcpserver agree on a target service provider \toolserver, which fixes a public query template $\mathcal{T}$ and a billing-related field selector $\varphi$.
          \toolserver is modeled as a standard TLS server. \\
          % so we omit \toolserver's actions. 
          % --- 1. HANDSHAKE ---
          \noindent\textbf{Handshake:}
          \begin{itemize}[leftmargin=1em, noitemsep, topsep=1pt]
              \item Upon receiving $(\sid, \mathsf{id}_{\toolserver})$ from \mcpserver, \fsys sends $(\sid, \mathsf{id}_{\toolserver})$ to \mcpclient.
              \item \mcpclient independently validates that the session is bound to the agreed service provider \toolserver; if it fails, \fsys outputs $\bot$ and aborts. Otherwise, \fsys records $(\sid, \mathsf{id}_{\toolserver})$ binding.
          \end{itemize}
          % --- 2. QUERY ---
          \noindent\textbf{Query:}
          \begin{itemize}[leftmargin=1em, noitemsep, topsep=1pt]
              \item Upon receiving $\mathcal{E}_{\textsf{A}}$ from \mcpclient and $\mathcal{E}_{\textsf{R}}$ from \mcpserver, \fsys computes $Q \leftarrow \textsf{Assemble}(\mathcal{T}, \mathcal{E}_\textsf{A}, \mathcal{E}_{\textsf{R}})$.
              \item \fsys outputs the TLS-protected query transcript $(\sid, \hat{Q}, \sigma_Q)$ to \mcpclient and \mcpserver; \mcpserver forwards $(\sid, \hat{Q}, \sigma_Q)$ to \toolserver.
          \end{itemize}
          % --- 3. RESPONSE ---
        \noindent\textbf{Response:}
        \begin{itemize}[leftmargin=1em, noitemsep, topsep=1pt]
            \item Upon receiving a TLS-protected response transcript $(\sid$, $\hat{R}$, $\sigma_R)$ from \toolserver, \fsys verifies its correctness; if it fails, \fsys outputs $\bot$ and aborts.
            \item Otherwise, \fsys forwards $(\sid, \hat{R}, \sigma_R)$ to both \mcpclient and \mcpserver, and outputs the plaintext response $(\sid, R)$ only to \mcpclient.
            \item Upon \mcpclient submitting a declared billing value $v$, \fsys outputs $(v,1)$ to \mcpserver if $R[\varphi] = v$, and $(v,0)$ otherwise.
        \end{itemize}
          \end{minipage}
      }
  \end{mdframed}
  \caption{Ideal Functionality \fsys.}
  \label{func:fsys}
\end{Functionality}

\parab{Threat Model}
In \sys, we assume that the downstream service providers (\eg LLM providers like OpenAI or tooling providers like GitHub) are honest. % whose endpoints are protected by standard TLS and faithfully process well-formed requests and return the corresponding responses.
Attacks originating from these service providers (\eg tool poisoning attacks~\cite{wang2025mcptox}, indirect prompt injection~\cite{chang2026overcoming}, and tool output attacks~\cite{zhao2025mcp}) are therefore outside the scope of this work.

%participants during protocol execution: 
We model both the agent and ARI as semi-honest but curious: 
they will not deviate arbitrarily from the protocol, but may try to infer each other's private information or selfishly reap more economic benefits
\revision{by choosing well-formed but self-serving input values at the protocol interfaces}.
% their local views.
% At the same time, both parties may have economic incentives in ARI-mediated interactions.
Specifically, a self-interested agent may try to infer ARI's private inputs, such as API keys, or reduce its payment by under-reporting token usage or disputing legitimate charges.
Meanwhile, a self-interested ARI may try to monetize sensitive agent data by learning private requests or prompts, increase profit by routing requests to unintended lower-tier service providers, or exaggerate usage for overcharging.
Essentially, \sys follows a ``trust-but-verify'' paradigm. The agent verifies ARI's routing behavior by authenticating the intended service provider and checking that the returned responses are authentic. Meanwhile, ARI verifies agent-side metering through the verifiable billing protocol, in which the agent provides a zero-knowledge proof that the reported billing fields are faithfully extracted from the encrypted response returned by the service provider. 
In \S\ref{sec:discussion}, we discuss how \sys can be extended toward a malicious-security model where the participants can arbitrarily deviate from the protocol. 
% These behaviors motivate our focus on privacy and accountability under a semi-honest execution model, while treating arbitrary malicious deviations from the protocol as out of scope.

\parab{Private Data}
The private data includes: \first the agent-side field contents $\mathcal{E}_{\textsf{A}}$ and the ARI-side field contents $\mathcal{E}_{\textsf{R}}$ used for query construction; \second the plaintext response $R$ released only to the agent. Additionally, all protocol-internal states computed from the aforementioned private data, including secret-shared intermediate values, must be kept secret throughout the process.

\parab{Public Data}
The public data includes: \first the session identifier $\sid$, the target service provider $\mathsf{id}_{\toolserver}$, the query template $\mathcal{T}$, and the billing-related field selector $\varphi$; \second the TLS-authenticate query and response transcripts $(\sid,\hat{Q},\sigma_Q)$ and $(\sid,\hat{R},\sigma_R)$; \third the declared billing value $v$ and the verification result indicating whether $R[\varphi]=v$.

%% file: 4.design.tex
% \section{System Design}
% \label{sec:design}

% This section presents the design of \sys, a privacy-preserving and verifiable Model Context Protocol for agentic AI. 
% We first introduce an MCP-adapted key schedule between the agent and the MCP server that selectively distributes TLS secrets to enforce asymmetric privileges separation. Building upon this, we describe a privacy-preserving tool-query protocol that securely assembles and encrypts requests from the agent and the MCP server's private inputs. Finally, we present a verifiable billing protocol leveraging zero-knowledge proofs to ensure fair metering on encrypted responses without compromising performance.

\input{handshake}

% \clearpage
\section{Privacy-Preserving Query-Construction}
\label{sec:query} 
\begin{figure}[t]
    \centering
    \includegraphics[width=0.65\columnwidth]{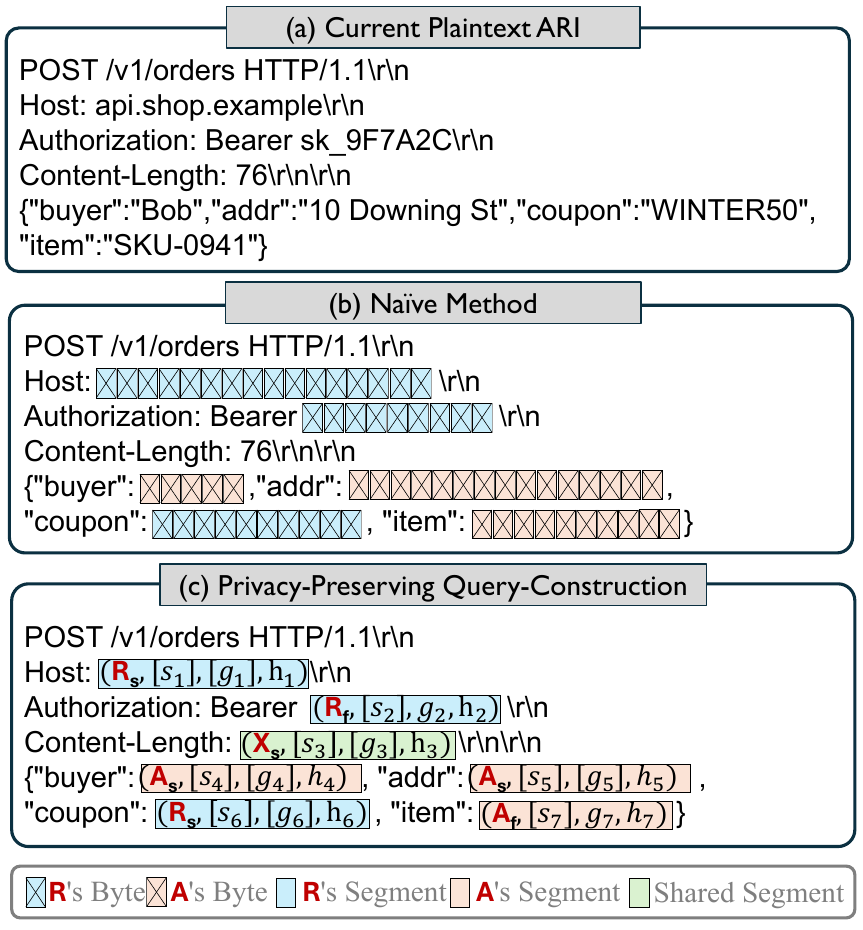}
    \caption{\revision{An example of a query template and its cryptographic abstraction for Agentic Tool Routing.}}
    \label{fig:template}
\end{figure}

% In conventional MCP implementations, as illustrated in Figure~\ref{fig:template}(a), the MCP server typically receives the agent's query to its intended tool in plaintext to perform necessary template filling and encryption. This inevitably discloses the proprietary agent data (\eg private prompts or authentication tokens) to the MCP server. 
% In this section, we introduce how the agent and MCP server can collaboratively generate a privacy-preserving query, where each party contributes partial information without revealing its private data to the other.
% Conventional MCP implementations require the agent to send its plaintext queries to the server for template filling and encryption, exposing sensitive data to an untrusted entity.
\revision{
% Conventional MCP implementations construct downstream tool requests at the MCP server, which exposes sensitive execution context such as credentials and tool queries in plaintext. 
% To prevent such exposure, \sys abstracts request construction as a collaborative filling process over a public tool-specific template, whose structural syntax and field ownership are known a priori; only the contents of designated sensitive fields are treated as private inputs. 
% We present a privacy-preserving query protocol that enables the agent and the server to collaboratively construct the final encrypted request without revealing any private inputs to each other.
In current ARI-mediated invocations, the agent sends its request context to ARI in plaintext and relies on ARI to complete the service invocation by adapting the request to the provider-specific format and adding any necessary ARI-held invocation data (\eg credentials), and forwarding it to the service provider.
This plaintext request-construction step exposes sensitive agent-side information (\eg private prompts in LLM routing, or sensitive tooling parameters)
% while forcing the agent to trust that ARI faithfully carries these inputs into the final request sent to the service provider.
and leaves the agent without cryptographic assurance that these fields are processed unchanged by ARI when constructing the provider-facing request.
Therefore, \sys introduces a privacy-preserving query-construction protocol that enables the agent and ARI to collaboratively construct the final encrypted request without revealing any private inputs to each other.

% For clarity, we consider a managed e-commerce checkout tool deployed via a commercial MCP server that aggregates downstream marketplace APIs in Figure~\ref{fig:template}. 
% The MCP server supplies the configured tool endpoint, its API credentials, along with enterprise-negotiated promotional incentives (\ie \texttt{coupon}). 
% The agent supplies the private buyer identities (\ie \texttt{buyer}), delivery destinations (\ie \texttt{shipto}), and the targeted product (\ie \texttt{item\_id}). 
% The downstream tool expects a standard HTTPS request whose header and payload naturally interleave public protocol syntax, server-owned fields, and agent-owned fields.
For clarity, we use \emph{Agent Tool Routing} as a running example, where an agent relies on ARI to invoke a managed e-commerce checkout service on its behalf, as illustrated in Figure~\ref{fig:template}.
In this example, ARI supplies the service-specific endpoint identifier, the required API credentials, and enterprise-negotiated promotional incentives (\ie \texttt{coupon}).
The agent supplies private buyer identities (\ie \texttt{buyer}), delivery destinations (\ie \texttt{addr}), and the target product (\ie \texttt{item}).
}
\subsection{Naive Method}
\label{subsec:query:naive}
We first consider a simple content-hiding approach for constructing requests. 
% Recall that the TLS Record protocol utilizes the AES-GCM (specifically AES-CTR for encryption) to generate a keystream from the session key $\textsf{tk}_\textsf{capp}$. 
Recall that, with an AES-GCM ciphersuite, TLS record encryption generates CTR-mode keystream blocks from the client application traffic key $\textsf{tk}_\textsf{capp}$.
For a plaintext block $p_i$, the keystream $k_i$ and resulting ciphertext $c_i$ are derived as:
\begin{equation}
\label{equ:aes_ctr}
% k_i = \textsf{AES}(\textsf{tk}_\textsf{capp}, \textsf{IV}_i); \quad c_i = k_i \oplus p_i
k_i = \textsf{AES}(\textsf{tk}_\textsf{capp}, \textsf{ctr}_i); \quad c_i = k_i \oplus p_i
\end{equation}
% where $\textsf{IV}_i$ represents the per-block initialization vector. 
where $\textsf{ctr}_i$ denotes the counter used for the $i$-th plaintext block.
Since $\langle \textsf{tk}_\textsf{capp} \rangle$ is secret-shared, both parties can execute a 2PC protocol to compute keystream shares $\langle k_i \rangle$.

% As shown in Figure~\ref{fig:template}(b), the MCP server could then prepare a \emph{query template} with predefined fields, including public (\public), MCP-server-owned (\mcpserver), and agent-owned (\mcpclient) chunks. 
% Each chunk $C_i$ has three attributes: owner $o_i$, content $s_i$ and its public length $h_i$.
% For instance, ``\texttt{alice}'' is a 5-bytes agent-owned chunk $C_i = \{o_i:\mcpclient, s_i:\text{``\texttt{alice}''}, h_i:5\}$, and ``\texttt{","password":"}'' is a 14-bytes public chunk $C_j = \{o_j:\public, s_j:\text{``\texttt{","password":"}''}, h_j:14\}$.
% This allows both parties to XOR their private data chunks with the corresponding keystream shares. 
% However, this strawman design leaks structural information. Because AES operates bitwise on fixed positions, both parties must define the exact offsets and lengths for each chunk within the tool request, which reveals critical metadata such as the precise length of a password. 
\revision{
As shown in Figure~\ref{fig:template}(b), a naive method is to instantiate the request skeleton as an ownership-labeled byte template with fixed byte offsets. The public syntax bytes are fixed by the template.
For an ARI-owned field, such as the \texttt{coupon} value \texttt{"WINTER50"}, the ARI places 10 bytes of plaintext at the designated positions, while the agent contributes an equal-length all-zero byte string as a placeholder. 
Conversely, for an agent-owned field, such as the \texttt{buyer} value \texttt{"Bob"}, the agent places the plaintext bytes and the ARI contributes equal-length zeros. 
Given the keystream shares generated under the secret-shared TLS key, the parties can XOR their byte contributions with the corresponding keystream shares to obtain ciphertext shares, which are then opened as a standard TLS ciphertext. 
However, this strawman protects only byte contents: because every private field must occupy predetermined positions with predetermined lengths, the template itself leaks structural metadata such as the exact length of a recipient address or coupon code.}
\subsection{Protocol Design}
\label{subsec:query:ours}
% To address the problem, our protocol enables collaborative construction of a tool-call query while hiding not only the actual content of private data chunks, but also their precise boundaries and lengths.
% As illustrated in Figure~\ref{fig:template}(c), to hide the precise length of private content, we set each chunk to a predetermined maximum size, such that the actual string is obfuscated within a fixed-length padded buffer. Formally, we define the public query template $\mathcal{T} = [C_1, \dots, C_L]$ consisting of chunks $C_i = (o_i, h_i)$, where $o_i$ denotes the owner and $h_i$ denotes the padded length. 
% Each chunk is obfuscated as a segment $E_i$, represented as a 4-tuple $(\theta_i, \langle s_i \rangle, \langle g_i \rangle, h_i)$, where $\theta_i$ is the category (including ownership and the segment attribute), $\langle s_i \rangle$ and $\langle g_i \rangle$ are the secret shares of the content and its actual length, and $h_i$ is the publicly known padded length.

To address this problem, our protocol enables collaborative query-construction while hiding not only the content of private fields, but also their precise boundaries and lengths. 
As illustrated in Figure~\ref{fig:template}(c), \sys represents each variable field as a padded segment with a public maximum length, so that the actual string is hidden inside a fixed-length buffer. 
Formally, the public query template is defined as a sequence of segment descriptors 
\(\mathcal{T}=[\tau_1,\ldots,\tau_L]\), where each descriptor \(\tau_i=(\theta_i,h_i)\) specifies the segment category \(\theta_i\) and its public padded length \(h_i\). 
An instantiated segment is represented as 
\(E_i=(\theta_i,\langle s_i\rangle,\langle g_i\rangle,h_i)\), where \(\langle s_i\rangle\) and \(\langle g_i\rangle\) are secret shares of the padded content and its actual length, respectively. 
For single-owner fields, the owner locally prepares the plaintext value and its length before secret-sharing them; for public syntax, the content and length are both public.

% We classify the segments into three distinct categories.
% Public segments (\public) consist of static protocol fields where the true length $g$ is identical to the padded length $h$. 
% A fixed-length private segment (including agent-owned $\mcpclient_f$, MCP-server-owned $\mcpserver_f$, and secret-shared $\shared_f$ fixed-length private segments) represents private fields with public lengths.
% For instance, API keys often have a 
% fixed length, allowing $g = h$ to be treated as a public constant. Such a segment can be represented as a server-owned fixed-length segment $E_1 = \{\theta_1:$$\mcpserver_f$$,s_1:\text{``\texttt{sk-api-key}''}$$,g_1:10,$$ h_1:10\}$, as shown in Figure~\ref{fig:template}(c).
% While $\mcpclient_f$ segments and $\mcpserver_f$ segments are initially owned by a single party, secret-shared segments $\shared_f$ result from the concatenation of segments from different owners, meaning neither the agent nor the server should possess the full knowledge of the actual content or the true length.  
% Structure-hidden private segments (including agent-owned $\mcpclient_s$, MCP-server-owned $\mcpserver_s$, and secret-shared $\shared_s$ structure-hidden private segments) represent private fields where the length $g$ is also treated as a secret.
% For instance, the password can be represented as an agent-owned structure-hidden segment $E_3=$$\{\theta_3:$$\mcpclient_s,$$s_3:\text{``\texttt{real-pw@}''},$$g_3:8$$,h_3:20\}$. 
% Before assembling, 
% the owner pads content $s$ to its public length $h$ with null bytes (\ie ``0x00'') to hide the actual length $g$.

We classify the segments into three categories. 
Public segments (\public) contain static protocol syntax or JSON delimiters, where the true length \(g\) equals the padded length \(h\). 
Fixed-length private segments include agent-owned \(\mcpclient_f\), ARI-owned \(\mcpserver_f\), and secret-shared \(\shared_f\) segments whose true lengths are public constants. 
For example, the ARI-owned bearer token in Figure~\ref{fig:template} can be modeled as an \(\mcpserver_f\) segment when the credential format has a fixed length, while the agent-owned \texttt{item} field can be modeled as a \(\mcpclient_f\) segment when it follows a fixed-format identifier. 
Structure-hidding private segments include \(\mcpclient_s\), \(\mcpserver_s\), and \(\shared_s\) segments whose true lengths are secret-shared. 
For example, the buyer identity and address are \(\mcpclient_s\) segments, and the ARI-selected \texttt{coupon} can be represented as an \(\mcpserver_s\) segment if the server does not want to reveal the coupon code or its length.
Before assembly, each owner pads its content to the public length \(h\) with null bytes (\ie ``0x00'') to hide the actual length \(g\).

\begin{figure}[!t]
    \centering 
    \subfigcapskip=-2pt
    \subfigure[LDC]{
        \label{fig:ldc}
        \includegraphics[width=0.39\columnwidth]{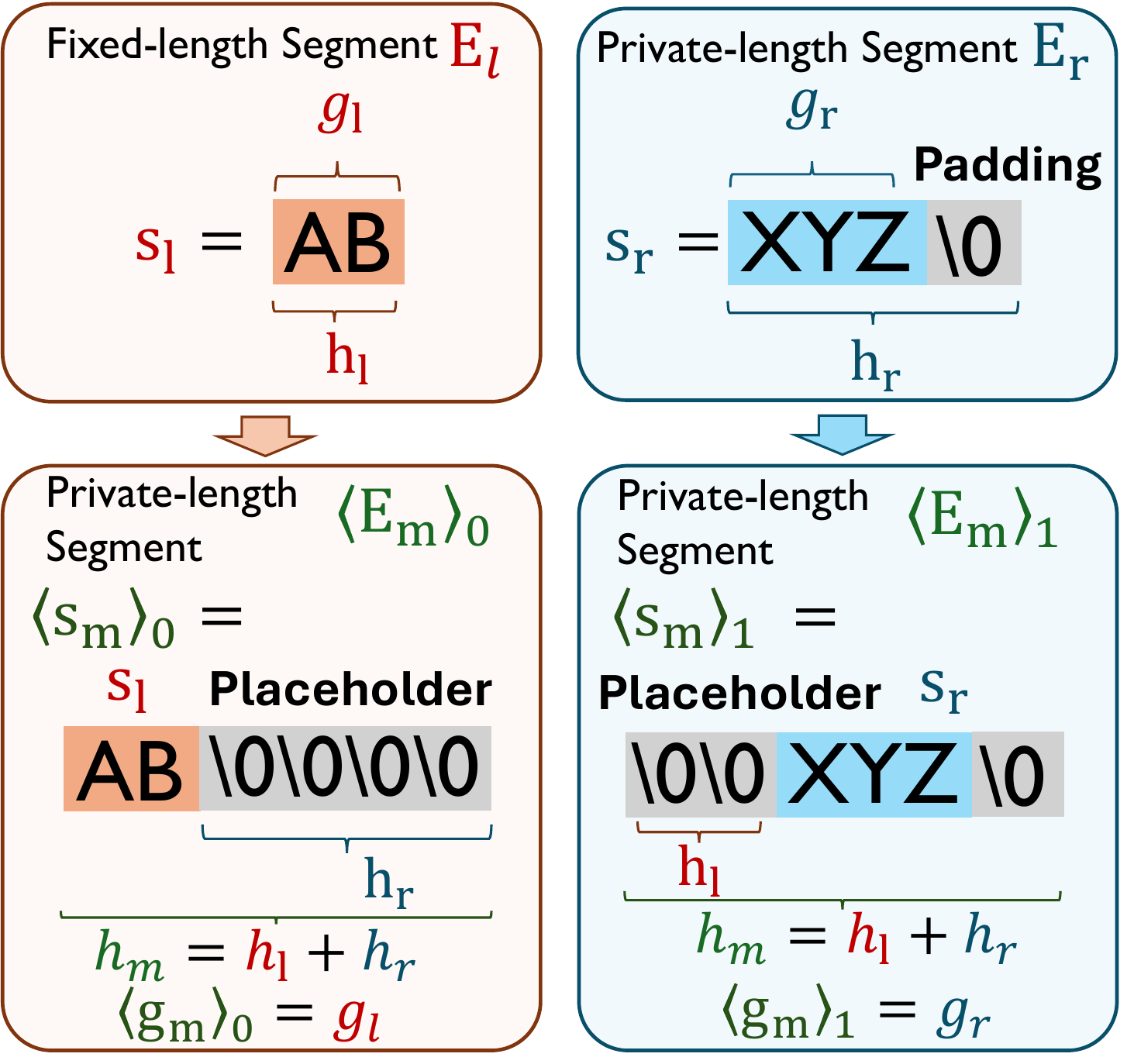}}
    % \hfill
    \subfigure[SHC]{
        \label{fig:shc}
        \includegraphics[width=0.4\columnwidth]{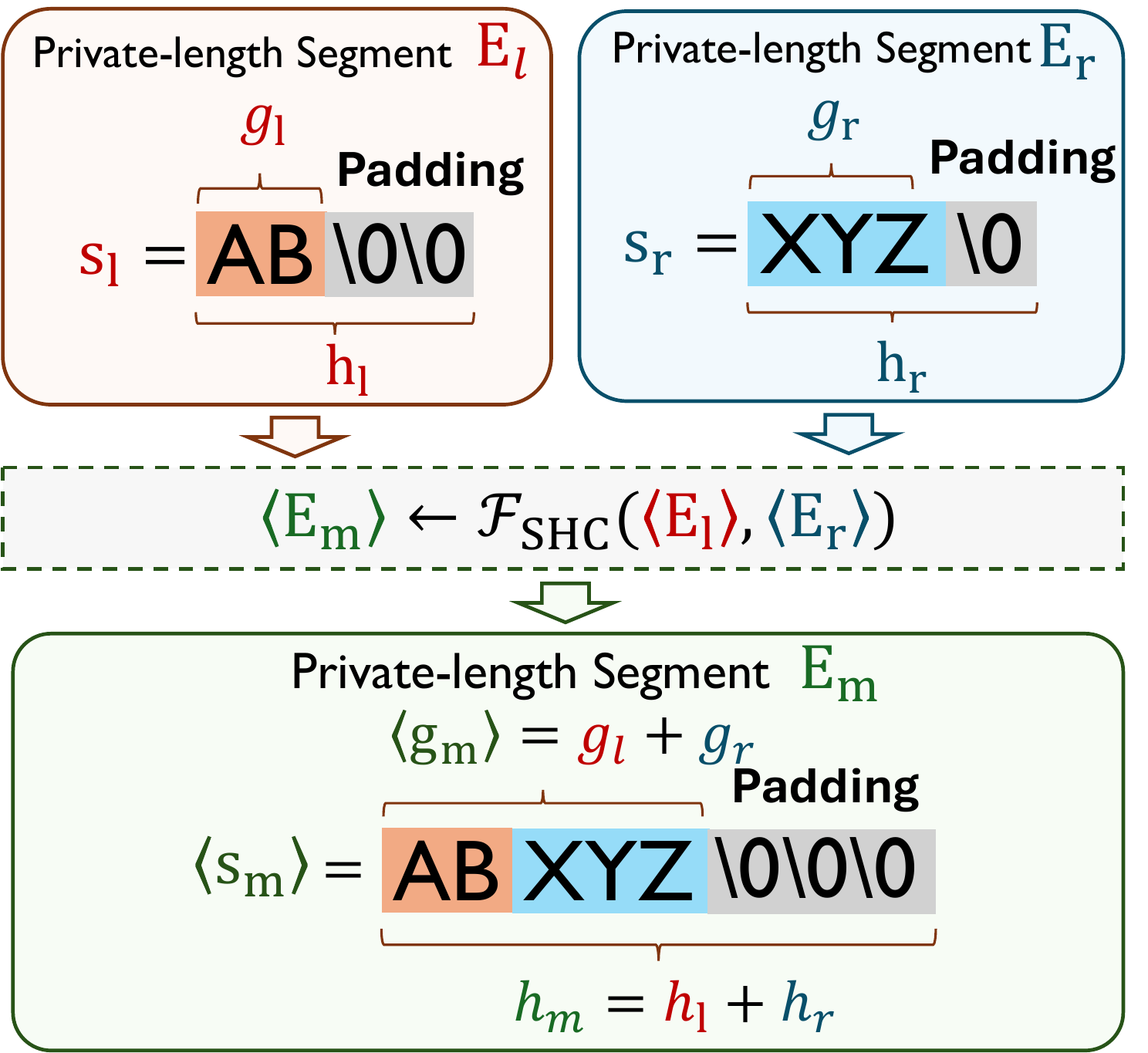}}
    \caption{An example of the local deterministic concatenation (LDC) and the structure-hiding concatenation (SHC).}
 \label{fig:concat_method}
\end{figure}

To assemble these three types of segments into a valid request, \sys must ensure that private segments are concatenated correctly even when their precise lengths are hidden. As illustrated in Figure~\ref{fig:concat_method}, we define two distinct concatenation paths based on whether the preceding segment’s length is public.

\parab{Local Deterministic Concatenation (\textsf{LDC})} 
% If the preceding segment has a public length, as is the case for public segments ($\public$) or fixed-length private segments ($\mcpclient_f, \mcpserver_f, \shared_f$), the starting offset of the subsequent segment is deterministic. In these scenarios, the agent and the MCP server can independently compute the relative positions of their respective shares. This allows for local concatenation without any communication overhead, effectively preserving the efficiency of the standard protocol for non-variable fields.
We first consider the case where the preceding segment $E_l$ has a public length, as illustrated in Figure~\ref{fig:concat_method}(a). %the preceding segment $S_l$ has a public length ($g_l = h_l$), such as a public segment (\public) or a fixed-length private segment ($\mcpclient_f, \mcpserver_f, \text{X}_f$). 
In this scenario, since the boundary of content $s_l = \text{``\texttt{AB}''}$ is known, the starting position for joining the content $s_r = \text{``\texttt{XYZ0}''}$ of segment $E_r$ is also deterministic. Here, ``\texttt{0}'' denotes a null byte.
Thus, to obtain the secret shares of the merged segment $E_m$, one party simply appends a $h_r$-byte placeholder (\ie ``0x00'') to $s_l$, while the other party prepends a $h_l$-byte placeholder to $s_r$. 
Consequently, both parties can compute the shares locally without any 2PC protocols. 
%aligns their respective shares of $S_r$ by appending a $h_r$-byte placeholder to $s_l$ or prepending a $h_l$-byte placeholder to $s_r$. 
The resulting segment $E_m$ is secret-shared by both parties: one holds $\langle E_m \rangle_0 = \{\theta_m, s_l \| \textsf{0x00}^{h_r} , g_l, h_l+h_r\}$ and the other one holds $\langle E_m \rangle_1 = \{\theta_m,  \textsf{0x00}^{h_l} \|s_r , g_r, h_l+h_r\}$. 
Thus, $E_m 
% = \langle E_m \rangle_0 \boxplus \langle E_m \rangle_1 
= \{\theta_m:\shared_s,s_m: (s_l \| \textsf{0x00}^{h_r}) \oplus (\textsf{0x00}^{h_l} \|s_r)=\text{``\texttt{ABXYZ0}''}, g_m: g_l + g_r=5, h_m : h_l + h_r=6\}$.
% \fixme{$\langle E_m \rangle_0 \boxplus \langle E_m \rangle_1 = \{\_, s_l \| \textsf{0x00}^{h_r} , g_l, h_l+h_r\} \boxplus \{\_,  \textsf{0x00}^{h_l} \|s_r , g_r, h_l+h_r\}=\{\shared_s,\langle s_m \rangle, \langle g_l + g_r \rangle, h_l + h_r\}$}.}
% \revision{$(\shared_s,\langle s_m \rangle, \langle g_m \rangle, h_m)$.}}
%updated with $h_m = h_l + h_r$ and $\langle g_m \rangle = \langle g_l \rangle + \langle g_r \rangle$, both of which are computed locally without any communication.}

\parab{Structure-Hiding Concatenation (\textsf{SHC})} 
% Nevertheless, the challenge arises when a segment has a secret-shared length $g$ (\eg variable-length private segments $\mcpclient_s, \mcpserver_s, \shared_s$). In this case, the displacement of all trailing data becomes a sensitive value, as concatenating at a public padded boundary $h$ would introduce malformed "gaps" that reveal structural metadata and break protocol compatibility. To address this, we design a Structure-Hidden Concatenation (\textsf{SHC}) protocol.
In contrast, when the length of the preceding segment $E_l$ is shared between two parties, naively joining two segments introduces malformed null bytes within the resulting segment. For instance, in Figure~\ref{fig:concat_method}(b), appending ``\texttt{XYZ0}'' to ``\texttt{AB00}'' yields ``\texttt{AB00XYZ0}'', which introduces two null bytes between the actual contents ``\texttt{AB}'' and ``\texttt{XYZ}''.
Therefore, we require a two-party computation protocol, named the Structure-Hiding Concatenation (SHC) protocol $\Pi_\textsf{SHC}$, to extract and merge the real content, while shifting all these padded null bytes to the end of the joined segment (\ie ``\texttt{ABXYZ000}''). 
% The resulting segment $E_m$ is represented as \revision{$(\shared_s,\langle s_m \rangle, \langle g_m \rangle, h_m)$}.}
% a secret-shared length $\langle g_l \rangle$ hidden within its padding $h_l$, such as in structure-hidden segments ($\mcpclient_s, \mcpserver_s, \shared_s$), the parties cannot determine where $s_l$'s actual content ends. 
% As shown in Figure~\ref{fig:concat_method}(b), a naive concatenation at the public boundary $h_l$ would introduce malformed gaps of null bytes between the actual data. To address this, we require a primitive $\mathcal{F}_{\text{SHC}}$ that extracts and merges the content (\eg ``AB'' and ``XYZ'') while shifting all padding to the end of the new segment $S_m$, resulting in a hidden length $\langle g_m \rangle = \langle g_l \rangle + \langle g_r \rangle$. We design a 2PC-based SHC protocol that leverages \emph{blind rotation} to resolve these padded boundaries.}

% \begin{figure}[t]
%     \centering
%     \includegraphics[width=\columnwidth]{figure/fig-template-big.pdf}
%     \caption{The structure-hidden concatenation protocol $\Pi_\textsf{SHC}$.}
%     \label{prot:concat-optm}
% \end{figure}

\input{algo/fig-template}

% The Structure-Hidden Concatenation Protocol $\Pi_\textsf{SHC}$ is summarized Protocol~\ref{prot:concat-optm}. 
Formally, $\Pi_\textsf{SHC}$ securely merges a left segment $E_l = (\theta_l, \langle s_l \rangle, \langle g_l \rangle, h_l)$ and a right segment $E_r= (\theta_r$, $\langle s_r \rangle$, $\langle g_r \rangle$, $h_r)$ into a secret-shared result $E_m = (\shared_s, \langle s_m \rangle, \langle g_m \rangle, h_m)$. We summarize it in Protocol~\ref{prot:concat-optm}.
% the secure merging of a left segment $\langle s_l, g_l, h_l \rangle$ and a right segment $\langle s_r, g_r, h_r \rangle$ into a secret-shared result $\langle s_m, g_m \rangle$ and the public $h_m = h_l + h_r$. 
As shown in Line 2--3, $\Pi_\textsf{SHC}$ first prepares extended vectors $\langle v_a \rangle$ and $\langle v_0 \rangle$ by %strategically 
padding the shares of $s_l$ and $s_r$ to a total length $h_m = h_l + h_r$. To eliminate these null bytes between the actual contents, the parties compute a secret shift distance $\langle d \rangle = h_l - \langle g_l \rangle \pmod{2^k}$. 
In Line~6, we invoke the $\Pi_\textsf{BlindRotate}$, which utilizes $\mathcal{F}_{\textsf{MUX}_2}$ to perform an oblivious left-shift of $\langle v_0 \rangle$ by the secret distance $d$. 
This ensures that the two segments are seamlessly joined at the sensitive boundary $g_l$ via a final XOR sum $\langle v_a \rangle \oplus \langle v_b \rangle$, resulting in a correctly compacted segment $E_m$ where the total length $\langle g_m \rangle = \langle g_l \rangle + \langle g_r \rangle$ remains hidden from both parties.
We defer the detailed $\Pi_\textsf{BlindRotate}$ in Appendix~\ref{sec:appendix:br}.

% We maintain the content length $g$ as an arithmetic share, because arithmetic sharing only requires a single $\mathcal{F}_\textsf{A2B} $ conversion within $\Pi_\textsf{SHC}$, whereas bitwise sharing need twice $\mathcal{F}_\textsf{B2A}$ conversions.
We maintain the content length $g$ as an arithmetic share. This choice requires only a single $\mathcal{F}_\textsf{A2B}$ conversion within $\Pi_\textsf{SHC}$, whereas bitwise sharing would require two $\mathcal{F}_\textsf{B2A}$ conversions.
% to perform length additions. 
We provide a detailed performance evaluation of different sharing domains in \S~\ref{subsubsec:arith}.

% The protocol first localizes the right segment to its maximum possible offset. To eliminate the gap between the end of the actual content in $s_l$ and the start of $s_r$, the parties compute a secret shift distance $d = h_l - g_l \pmod{2^k}$ within the MPC circuit. We then invoke a BlindRotate subprotocol, which utilizes an OT-based shift operation to perform an oblivious left-shift of the right segment shares. This ensures the two segments are seamlessly joined at the sensitive boundary $g_l$ without either party learning the value of $g_l$ or the content of the other's segments. 
% By recursively applying \textsf{SHC}, the parties can construct complex payloads where the precise location and length of every sensitive input remain hidden.

% \begin{figure}[t]
%     \centering
%     \includegraphics[width=\columnwidth]{figure/fig-query-big.pdf}
%     \caption{The privacy-preserving query protocol $\Pi_\textsf{query}$.}
%     \label{prot:template}
% \end{figure}

\parab{Collaborative Query Assembly and Encryption} 
Based on $\Pi_\textsf{LDC}$ (\ie the \textsf{LDC} protocol) and $\Pi_\textsf{SHC}$, the agent and ARI collaboratively assemble the full request by following an assembly plan $\Omega$. As detailed in Protocol $\Pi_\textsf{query}$ in Protocol~\ref{prot:mcpquery}, $\Omega$ defines a series of iterative assembly steps $\omega = (i, k, j)$, where the parties merge previously assembled segments $E_{[i,k]}$ and $E_{[k+1,j]}$ into a larger range $E_{[i,j]}$. For each step, the protocol adaptively selects the concatenation primitive: 
% if the preceding segment has a public length ($g_l = h_l$), 
if the preceding segment has a public true length according to the template metadata,
it triggers the efficient $\Pi_\textsf{LDC}$ update; otherwise, it invokes $\Pi_\textsf{SHC}$ protocol to handle hidden boundaries. 
Once the iterative assembly is complete, the parties obtain the final secret-shared query string $\langle Q \rangle$ by extracting the content part of the final segment $E_{[1,L]}$. 
% $\langle Q  \rangle = S_{[1,L]}.\langle s \rangle$. }

% With the structure-hidden request $\langle p \rangle$ and the client application key $\langle \textsf{tk}_\textsf{capp} \rangle$ represented as bitwise secret shares, the agent and MCP server jointly execute a 2PC-based AEAD circuit to derive keystream shares $\langle k_i \rangle = AES(\langle \textsf{tk}_\textsf{capp} \rangle, \textsf{IV}_i)$. Each ciphertext block is reconstructed as $\theta_i = \langle k_i \rangle_0 \oplus \langle k_i \rangle_1 \oplus \langle p_i \rangle_0 \oplus \langle p_i \rangle_1$. 
Once the final secret-shared query string $\langle Q \rangle$ is obtained, the agent and ARI jointly execute a 2PC-AEAD circuit $\Pi_\textsf{2PC-AEAD}$ using their respective shares of the query and the client application key $\langle \textsf{tk}_{\textsf{capp}} \rangle$. Internally, the parties securely derive keystream shares $\langle k \rangle = \text{AES}(\langle \textsf{tk}_{\textsf{capp}} \rangle, \text{IV})$ through a 2PC evaluation, then they compute ciphertext $\hat{Q} = \langle k \rangle \oplus \langle Q \rangle$ and its GMAC tag $\sigma_Q$.
% allowing the ARI to reconstruct the final ciphertext as $\hat{Q} = \langle k \rangle \oplus \langle Q \rangle$ and its GMAC tag $\sigma$.
The ciphertext and tag are then opened to both parties; ARI then forwards them to the service provider as a standard TLS-compliant record.
Due to the page limit, we defer the security proof to Appendix \ref{sec:appendix:query-proof}.
\input{algo/fig-query}
\subsection{Extensions}
\label{subsec:query:optm}

\parab{Concatenation Planning} 
Repeated use of $\Pi_\textsf{SHC}$ introduces a ``structural contamination'' problem. 
% Specifically, once a newly assembled range has a secret-shared true length,  (\eg $\langle E_{[1,2]} \rangle \leftarrow \Pi_\text{SHC}(\langle E_{[1,1]} \rangle, \langle E_{[2,2]}\rangle)$), its end boundary is hidden from both parties. 
% % Consequently, any subsequent segments $E_{[3,3]}, \dots, E_{[L,L]}$ must also be processed through $\Pi_\textsf{SHC}$ to be appended to $E_{[1,2]}$, even if they possess public or fixed-length segments, as their starting offsets are no longer deterministically known.
% % Consequently, every succeeding assembly operation must be executed via $\Pi_\textsf{SHC}$ rather than the more lightweight, efficient $\Pi_\textsf{LDC}$. 
% As a result, appending any subsequent range to \(E_{[1,2]}\) must use $\Pi_\textsf{SHC}$, even if that range contains only public or fixed-length segments, because its starting offset depends on the hidden end boundary of \(E_{[1,2]}\).
\revision{Specifically, once a newly assembled range has a secret-shared true length, its end boundary becomes hidden from both parties. 
For example, after 
\(\langle E_{[1,2]} \rangle \gets \Pi_\textsf{SHC}(\langle E_{[1,1]} \rangle, \langle E_{[2,2]} \rangle)\), 
any later range (\eg $E_{[3,3]}$) must be appended to \(E_{[1,2]}\) using \(\Pi_\textsf{SHC}\), even if it contains only public or fixed-length segments, because its placement depends on the hidden end boundary of \(E_{[1,2]}\).}
To minimize the cumulative overhead, we observe that the complexity of a single $\Pi_\textsf{SHC}$ operation is $\mathcal{O}((h_l+h_r)\log{}(h_l+h_r))$. Therefore, we formulate the request assembly as an optimization problem to find a concatenation order that minimizes the total cost.

We define $\Gamma(i, j)$ as the minimum cost to assemble a contiguous sequence of segments from index $i$ to $j$. The objective is to minimize the total cost $\Gamma(1, L)$ for $L$ initial segments, formulated as $\Gamma(i, j) = \min_{i \leq k < j} \left\{ \Gamma(i, k) + \Gamma(k+1, j) + \Delta(i, k, j) \right\}$,
% formulated as the following recurrence:}
% \begin{equation}
% \label{equ:dp}
%     \Gamma(i, j) = \min_{i \leq k < j} \left\{ \Gamma(i, k) + \Gamma(k+1, j) + \Delta(i, k, j) \right\}
% \end{equation}
where $\Delta(i, k, j)$ represents the marginal cost of merging segment ranges \(E_{[i,k]}\) and \(E_{[k+1,j]}\).
% We set $\Delta$ to zero if the operation qualifies as $\Pi_\textsf{LDC}$; otherwise, it is assigned the estimated 2PC communication and computation cost of the corresponding $\Pi_\textsf{SHC}$. 
% We set $\Delta(i,k,j)=0$ when the left range \(E_{[i,k]}\) has a public true length according to the template metadata; otherwise, \(\Delta\) is the estimated communication and computation cost of the corresponding $\Pi_\textsf{SHC}$.
\revision{We set \(\Delta(i,k,j)=0\) when the left range \(E_{[i,k]}\) has a public true length according to the template metadata, so the merge can be performed locally using the efficient \(\Pi_\textsf{LDC}\) primitive. 
Otherwise, \(\Delta(i,k,j)\) is set to the estimated communication and computation cost of the corresponding \(\Pi_\textsf{SHC}\) operation.}
By solving this optimization via dynamic programming, \sys generates an optimal plan $\Omega$ that 
% effectively minimizes the number of invocations 
minimizes the estimated total cost
of expensive $\Pi_\textsf{SHC}$.
\revision{\parab{Handling Protocol-Specific Metadata} 
Some protocols require metadata that depends on private field lengths. 
For example, an HTTP \texttt{Content-Length} header must encode the exact body length as an ASCII decimal string, while this length may depend on secret-shared payload segments. 
To handle such cases, \sys treats the metadata field as a secret-shared segment and computes its value with a small MPC metadata-generation primitive. 
% For \texttt{Content-Length}, the parties first compute the secret-shared body length from the relevant segment lengths and then use an Integer-to-String primitive, $\Pi_\textsf{I2S}$, to obtain a secret-shared ASCII representation that can be assembled into the request. 
% This allows the final request to satisfy protocol-specific parsing requirements without revealing the computed metadata during construction.
For \texttt{Content-Length}, the parties compute $\langle g_{\mathsf{body}}\rangle =
\sum_{i\in \mathcal{I}_{\mathsf{body}}} \langle g_i\rangle$,
where \(\mathcal{I}_{\mathsf{body}}\) denotes the body segments, and convert this arithmetic-shared integer into an XOR-shared ASCII decimal string using an Integer-to-String primitive \(\Pi_\textsf{I2S}\). 
The resulting string is then assembled into the request like other secret-shared segments, without revealing the computed body length during construction. 
We defer the details of $\Pi_\textsf{I2S}$ to Appendix~\ref{sec:appendix:i2s}.}

% Notably, while both the \textsf{BlindRotate} and \textsf{I2S} are inherent in bitwise-shared values, 

% We maintain the content length $g$ as an arithmetic secret share. This is because the arithmetic sharing only requires a single \textsf{A2B} conversion within the \textsf{SHC} protocol, whereas bitwise sharing would necessitate twice \textsf{B2A} conversions to perform length additions. We provide a detailed performance evaluation of these competing sharing domains in Section~\ref{sec:eval}.

% == long version ==
% \input{content-length}
% \parab{Extensibility to Other Application-layer Protocols} 
% Although we explain our query construction protocol for HTTP requests, 
% % our methodology is universally applicable to any application-layer protocols. 
% our methodology extends to other application-layer protocols whose messages can be expressed as templates of public syntax and private fields.
% This is because our method is based on protocol-agnostic structural abstraction: decomposing the packet into discrete segments based on ownership and privacy requirements. 
% % Given a packet, regardless of its format, it is always feasible to design a corresponding segment assembly plan 
% For such protocols, one can instantiate a corresponding segment assembly plan
% so that the agent and ARI can collectively construct well-formed messages in a privacy-preserving manner. 

% == short version ==
\revision{\parab{Beyond HTTP}
Although we instantiate \sys with HTTP requests, the segment-based abstraction is not HTTP-specific. 
It applies to any application-layer message that can be decomposed into public syntax and private fields. 
For such protocols, developers can define protocol-specific segment descriptors and metadata-generation primitives, while reusing the same LDC/SHC-based assembly framework.}

\section{Verifiable Billing Protocol}
\label{sec:response}

% \subsection{Circuit Design}
% \label{subsec:response:circuit}

% \input{template}

% Commercial MCP servers often adopt usage-based billing models (\eg per-invocation or per-token billing), where critical billing metadata (\eg via \emph{``status''} or \emph{``usage''} field) is embedded in the tool's response. However, in \sys, only the agent holds the response decryption key $\textsf{tk}_\textsf{sapp}$ to prevent the MCP server from reading the response. 
% This creates a problem of usage under-reporting by the agent.  
% To address this problem, \sys implements a verifiable billing protocol based on NIZK proofs. The protocol requires the agent to provide a proof $\pi$ to demonstrate that its reported billing metrics are faithfully extracted from the TLS-authenticated ciphertext. 
\revision{
% ARI usually adopts usage-based billing settlement (\eg per-invocation or per-token billing), where critical billing metadata (\eg the \texttt{token\_usage} fields) is embedded within the service responses. 
ARI-mediated invocations often use usage-based billing (\eg per-invocation or per-token billing), where metering fields (\eg \texttt{success} in tool response or \texttt{token\_usage} in LLM response) are embedded in the service responses. 
However, because \sys restricts response decryption exclusively to the agent via $\textsf{tk}_\textsf{sapp}$, the ARI remains blind to these fields and must rely entirely on self-reported metering by the agent. 
This information asymmetry 
% creates an inherent economic incentive for agents 
allows an economically motivated agent to under-report usage (\eg by claiming that a billable invocation failed or deflating token consumption). 
To be compatible with the existing settlement conventions, %without sacrificing privacy, 
\sys implements a verifiable billing protocol based on ZKP, requiring the agent to generate a proof demonstrating that its reported billing metrics are faithfully extracted from the encrypted TLS response record under the committed server-application traffic key $\textsf{tk}_\textsf{sapp}$.
}

% General-purpose TLS oracles~\cite{zhang2020deco} provide a foundation for verifiable data extraction by proving strings within well-defined context-free grammars. For instance, consider a tool response $R$ formatted as a JSON object where the target billing metric $\phi$ is ``\texttt{X}'', as illustrated in Figure~\ref{fig:ckt-compare}. 
% \sys must prove that the value of ``\texttt{X}'' is exactly ``\texttt{ValueX}'' and that its position within the payload is valid. 
% This validity check requires: 
% % is verified on two levels: 
% (1) \emph{Syntactic Integrity}, ensuring the key-value pair ``\texttt{"X":"ValueX"}'' resides at the correct logical depth, \ie it is a top-level member of the response payload $R$ rather than being nested within another field like ``\texttt{G}''; and (2) \emph{Ciphertext Alignment}, proving that the substring ``\texttt{"X":"ValueX"}'' locates at the precise byte offsets (\eg from $58_{th}$ to $69_{th}$ bytes) that correspond to the authenticated ciphertext blocks decrypted via AES.
% % \fixme{For instance, add an example} % (Here, while we use JSON as a representative example, these methods are naturally extensible to other structured languages such as HTML or XML.) 

\begin{figure}[t]
    \centering
    \includegraphics[width=0.65\columnwidth]{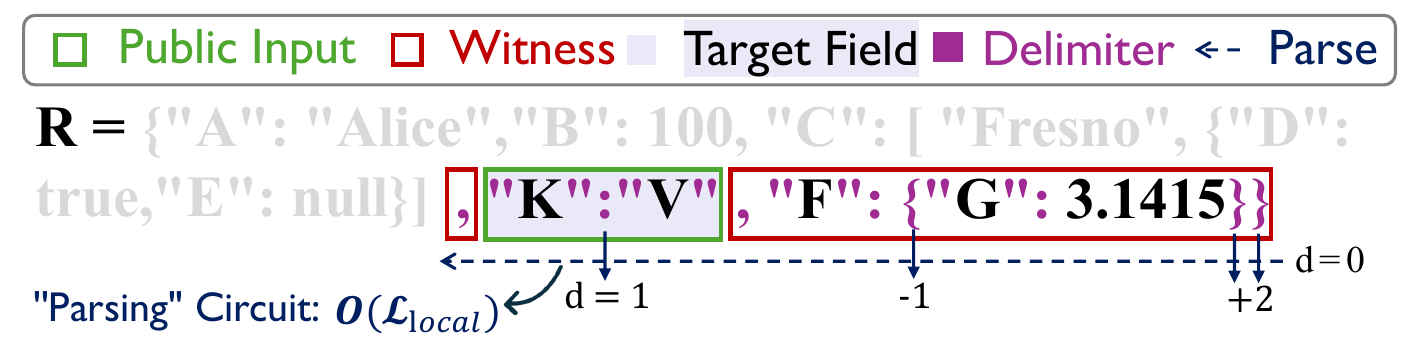}
    \caption{\revision{Overview of the localized parsing circuit for verifiable billing.}}
    \label{fig:ckt-compare}
\end{figure}

% However, existing methods incur prohibitive costs when proving complex MCP payloads. 
\revision{In our proving circuit, the target billing attributes consist of a predefined selector $\varphi$ and the agent-reported value $v$, both of which are treated as public inputs by the verifier (\ie ARI). 
Consequently, \sys only needs to assert two orthogonal properties in the ZKP circuit: 
\first \emph{structural isolation}: the proof must show that the reported value is the value associated with the predefined selector \(\varphi\) at the expected JSON path/depth, rather than a matching substring inside an unrelated nested field or string literal. 
For example, in Figure~\ref{fig:ckt-compare}, the selector ``K'' is a top-level member of the response payload rather than being nested within another field like ``G''.
\second \emph{cryptographic binding}: every byte used to satisfy this selector/value check must be the decryption of the public TLS response ciphertext under the committed server-application traffic key $\textsf{tk}_\textsf{sapp}$, \eg the selector ``K'' is located at byte offset 58.}

% As shown in Figure~\ref{fig:ckt-compare} (left), existing methods~\cite{zhang2020deco} %\emph{building circuits} 
% treat sensitive values and surrounding syntax as witnesses and public inputs chunks, respectively. 
% While these methods can easily verify syntactic integrity, it is costly to prove ciphertext alignment. 
% Because the verifier only observes the encrypted TLS packet and lacks the precise offsets of specific fields, the circuit must execute expensive \emph{``building''} logic (analogous to assembling a jigsaw puzzle) to interleave these chunks into a logically consistent plaintext. 
% Then the circuit must prove the decryption of the entire sequence to ensure ciphertext alignment, resulting in a computational overhead proportional to the full response length $\mathcal{L}_1$.

% \revision{To achieve this, existing TLS oracles rely on either global plaintext reconstruction~\cite{zhang2020deco}, which interleaves disjoint plaintext chunks inside the circuit, or full-text parsing tree derivation~\cite{angel2026coral}, which executes stack-automata constraints to recursively parse and check every syntactic node. 
% While these general paradigms successfully enforce structural isolation, they introduce excessive cryptographic constraints to maintain cryptographic binding, inevitably tying their proving complexity directly to the full response length $\mathcal{L}_{\text{full}}$.}
\revision{Existing TLS-oracle systems typically handle such checks by reconstructing selected plaintext chunks inside the circuit~\cite{zhang2020deco} or by deriving a full parsing structure over the response~\cite{angel2026coral}. 
These general-purpose designs are well-suited for flexible web attestation, yet their cryptographic binding cost scales with the amount of ciphertext that must be decrypted and parsed in-circuit. 
In our billing setting, the target field is predefined by the provider schema, allowing \sys to localize both decryption and parsing to a boundary-anchored window around \(\varphi:v\).}

\sys designs a localized \emph{parsing circuit} that enforces both properties only over a selected response window. 
As illustrated in Figure~\ref{fig:ckt-compare}, the circuit maintains arithmetized parser states for string boundaries, escape characters, and JSON depth. 
The depth accumulator \(d\) is updated only outside string literals, ensuring that braces appearing inside strings do not affect the structural check. 
The circuit then verifies that the target key-value pair \(\varphi:v\) appears at the expected syntactic depth.

To reduce both parsing and cryptographic decryption constraints, \sys uses an \emph{adaptive boundary window selection}. 
The window descriptor \(\omega=(o,\ell,\rho)\) specifies the local ciphertext window by its block offset \(o\), length \(\ell\), and parsing direction \(\rho \in \{\mathsf{fwd},\mathsf{rev}\}\). 
% For a prefix window, the circuit parses forward from the JSON start boundary to the target field; for a suffix window, it parses backward from the JSON terminal boundary to the target field using the corresponding reverse depth transition. 
\revision{The protocol considers two boundary-anchored windows that contain the target key-value pair $\varphi$:$v$: a prefix window parsed forward from the JSON start boundary, and a suffix window parsed backward from the JSON terminal boundary using the corresponding reverse depth transition.}
Because AES-GCM encryption is counter-mode based, the circuit only needs to prove decryption for the ciphertext blocks covered by \(\omega\), rather than for the full response.
\revision{
% Instead of evaluating the entire response, \sys selects the shorter boundary-anchored window that contains \(\varphi:v\). 
\sys then selects the shorter candidate window.
For example, if reaching the billing field requires decrypting and parsing five AES blocks from the start boundary but only three AES blocks from the terminal boundary, \sys selects the suffix window and reduces the effective proof length to \(\mathcal{L}_{\text{local}}\). }
Notably, by positioning the target data near the response boundaries---a configuration negotiable with the service provider---\sys maintains negligible proving overhead independent of the total response length.

\revision{
\parab{Formal Statement} 
We formalize billing attestation as an NP relation checked by the parsing circuit \(\textsf{Ckt}(\mathbb{X}, \mathbb{W})\) in Figure~\ref{fig:bill}. 
The public statement is $
\mathbb{X}=(\textsf{tk}^\textsf{R}_\textsf{sapp}, \textsf{com}_A, \hat{R}, r, \varphi, v, \omega, d_t, j, k)$,
where \(\hat{R}\) is the ARI-provided TLS response record, \(\omega\) identifies the local ciphertext window and parsing direction, \(d_t\) is the target JSON depth, and \([j,k]\) bounds the target key-value pair within the local plaintext. 
The private witness is \(\mathbb{W}=(\textsf{tk}^\textsf{A}_\textsf{sapp}, R_{\text{local}})\), consisting of the agent's key share and the decrypted local plaintext segment. Inside the circuit, \(\textsf{Ckt}\) checks the commitment \(\textsf{com}_A=\textsf{H}(\textsf{tk}^\textsf{A}_\textsf{sapp},r)\), reconstructs \(\textsf{tk}_\textsf{sapp}\), and constrains the selected ciphertext blocks \(\hat{R}[\omega]\) to decrypt to \(R_{\text{local}}\). 
It then tracks string, escape, and depth states in the selected parsing direction and enforces that \(R_{\text{local}}[j\dots k]\) encodes the public key-value pair \(\varphi:v\) at depth \(d_t\).
}

\parab{Completed Protocol} The verifiable billing protocol follows a standard NIZK lifecycle. The initial setup phase produces the proving key $\textsf{pk}$ and verification key $\textsf{vk}$ based on the definition of the circuit $\textsf{Ckt}$. To initiate billing verification, the agent executes the prove algorithm, taking the public inputs $\mathbb{X}$, its private witness $\mathbb{W}$, and $\textsf{pk}$ to generate a succinct proof $\pi$. This proof is then submitted to the ARI. The ARI concludes the process by running the verify algorithm using $\textsf{vk}$, $\pi$, and $\mathbb{X}$. If the algorithm outputs true, ARI is cryptographically assured that the billing metrics are authentic and structurally valid within the original TLS record, even without direct access to the plaintext response. \revision{Crucially, the agent can asynchronously generate proofs in parallel after collecting multiple responses for batched settlement, avoiding blocking the active service invocations.}

\input{algo/fig-circuit}
% \begin{figure}[t]
%     \centering
%     \includegraphics[width=\columnwidth]{figure/fig-circuit-big.pdf}
%     \caption{The verifiable billing protocol \(\Pi_{\text{bill}}\). The agent proves knowledge of \(\textsf{tk}^\textsf{A}_\textsf{sapp}\) and enforces selective decryption with header/payload checks for the targeted metering field.}
%     \label{fig:bill}
% \end{figure}

% \sys designs a payment protocol leveraging the x402 escrow scheme~\cite{x402_escrow_839} to achieve trustless and automated payment settlement. 
\sys can integrate the billing proof to an X402-style escrow settlement~\cite{x402_escrow_839,trustengine_sol}. 
% It follows a two-phase lifecycle:
The payment lifecycle follows two phases.
(i) a deposit and lock phase where funds are escrowed via the x402 engine, and (ii) a settle and distribute phase where the proof $\pi$ triggers the automated fee distribution based on authentic usage. Details are deferred to Appendix~\ref{sec:appendix:x402}.

%% file: handshake.tex
\section{ARI-Adapted Three-party Handshake}
\label{sec:handshake}

% Before the agent and ARI interact with a service provider, they first perform a one-time connection setup that establishes the TLS session and distributes key material according to ARI's role-specific interaction semantics. 
% In \sys, we upgrade this setup into an MCP-adapted three-party TLS handshake protocol to negotiate the TLS session keys with the target tools while enforcing \emph{asymmetric privileges} required in the MCP ecosystem.
% Specifically, the agent and MCP server jointly emulate a ``virtual TLS client'' to execute an MPC-based handshake with the intended tool services, while keeping the tool-side implementation unmodified. 
% Within this framework, our key schedule partitions session secrets based on the functional roles of different entities: the agent exclusively holds the secrets for tool authentication and response decryption, while the TLS client-side encryption keys are secret-shared between the agent and the MCP server. 
% This strategic distribution ensures the agent can independently authenticate the tool and obtain tool responses, whereas the MCP server only participates in collaborative query construction without gaining plaintext visibility or the ability to misroute agent's request to unintended tools. The handshake follows the standard TLS 1.3 structure, which we partition into the following four phases. 
\revision{Before the agent and ARI interact with a service provider, \sys first performs a one-time connection setup that binds the interaction to the intended provider and establishes a channel to be reused across subsequent ARI-mediated invocations.}
% derives key material to be reused across subsequent ARI-mediated invocations.
We instantiate this setup as an \emph{ARI-adapted three-party TLS handshake}, which allows the agent and ARI to jointly establish a standard TLS connection with the intended service provider while realizing ARI's role-specific interaction semantics.
Specifically, the agent and ARI jointly emulate a distributed ``virtual TLS client'' and, using secure multiparty computation, complete the handshake with the service provider without requiring any modification to the provider-side TLS implementation.
During this handshake, \sys distributes the derived TLS key material across protocol stages to align each party's cryptographic capabilities with its role in ARI-mediated invocations.
As a result, the agent can detect misrouting to unintended service providers, the agent and ARI can collaboratively construct service requests, and the agent ultimately obtains the complete provider response while ARI learns only the billing-related view.
Following the TLS 1.3 handshake stages, we describe how \sys distributes key material across the following four phases to realize these properties.

\parab{Key Exchange Phase} 
The agent and ARI first jointly emulate the client side of an ECDHE exchange with the service provider.
Specifically, the agent and ARI utilize $\textsf{ECtF}$~\cite{celi2023distefano} to generate a secret-shared input key $\langle\textsf{DHE}\rangle$, which is further expanded into the handshake secret $\langle\textsf{HS}\rangle$.
To reduce the overhead of 2PC computation, we incorporate a pre-computation and reuse strategy~\cite{xie2024lightweight} that caches intermediate SHA-256 compression states, significantly optimizing the 2PC-assisted HKDF expansions for deriving $\langle\textsf{dHS}\rangle$, $\langle\textsf{CHTS}\rangle$, and $\langle\textsf{SHTS}\rangle$. 
\revision{To bind the connection to the intended endpoint, ARI delivers its shares of the handshake traffic secrets (\ie $\textsf{CHTS}^\textsf{R}$ and $\textsf{SHTS}^\textsf{R}$) to the agent, enabling the agent to locally derive the handshake keys ($\textsf{tk}_\textsf{chs}, \textsf{tk}_\textsf{shs}$) 
% required for processing the encrypted handshake flight
% and eliminating the expensive 2PC computation.
, avoiding expensive 2PC for handshake-key derivation.
Crucially, since ARI never obtains the complete server-side handshake traffic secret $\textsf{SHTS}$, it cannot forge the server-authentication transcript or substitute an unintended service provider~\cite{lauinger2025janus}.
% , preserving the authenticity of the established session.
}

\parab{Service Provider Authentication Phase} 
% The ARI forwards the TLS handshake messages from the service provider to the agent, including its Certificate, \tls{SCV}, \tls{SF}, \etc. Utilizing the locally reconstructed $\textsf{SHTS}$, the agent independently expands the Server Finished key $\textsf{fk}_s$ to verify the certificate of the service provider and the authentication tag. This phase ensures that the agent and ARI are communicating with the intended, pre-declared service endpoint without ARI being able to intercept the authentication process.
ARI forwards the TLS handshake messages from the service provider to the agent, including the provider's Certificate, \tls{SCV}, \tls{SF}, \etc.
Using the locally reconstructed $\textsf{SHTS}$, the agent decrypts the server handshake flight, verifies the service provider's certificate and CertificateVerify message, and expands the Server Finished key $\textsf{fk}_s$ to validate the Finished MAC.
\revision{This phase ensures that the agent can independently authenticate the intended service endpoint, 
% without relying on ARI to perform or report the authentication result based on an expensive zero-knowledge proof existing TLS oracles~\cite{celi2023distefano}.
without relying on ARI to perform authentication and prove the result using expensive zero-knowledge proofs, as in existing TLS oracles~\cite{celi2023distefano}.
}

\parab{Application Key Generation Phase} 
% Following successful authentication, the two parties jointly derive the application-layer keys while maintaining the established privilege separation. The agent and ARI first extract the secret-shared master secret $\langle\textsf{MS}\rangle$ from $\langle\textsf{dHS}\rangle$. 
After the service provider is authenticated, the agent and ARI jointly derive the application traffic secrets while preserving the role-specific separation.
They first derive the secret-shared master secret $\langle\textsf{MS}\rangle$ from $\langle\textsf{dHS}\rangle$, and then expand it into the client-side and server-side application traffic secrets, \ie $\langle\textsf{CATS}\rangle$ and $\langle\textsf{SATS}\rangle$.
\revision{Here, the same pre-computation and caching strategy is applied to the HKDF expansions, reducing the 2PC overhead for deriving the application traffic secrets $\langle\textsf{CATS}\rangle$ and $\langle\textsf{SATS} \rangle$, application traffic keys $\langle\textsf{tk}_\textsf{capp}\rangle$ and $ \langle\textsf{tk}_\textsf{sapp}\rangle$, and their corresponding initialization vectors $\textsf{iv}_\textsf{capp}$ and $ \textsf{iv}_\textsf{sapp}$.}
The resulting keys are released according to their roles in the subsequent ARI-mediated interaction.
The client-side application key $\textsf{tk}_\textsf{capp}$ remains secret-shared between the agent and ARI, so that service requests can only be encrypted collaboratively.
In contrast, the server-side application key $\textsf{tk}_\textsf{sapp}$ is released only to the agent, so that the agent can decrypt the complete response from the service provider while ARI remains limited to its authorized billing-related view.
To ensure that this release remains bound to the authenticated TLS session, the agent first commits to its share $\textsf{tk}^{\textsf{A}}_\textsf{sapp}$ before ARI reveals its share $\textsf{tk}^{\textsf{R}}_\textsf{sapp}$.
This ordering prevents the agent from adaptively manipulating its key share after seeing ARI's share, thereby supporting later verification of billing-related claims while keeping ARI excluded from the full response plaintext.

\parab{Client Finished Phase} 
% To finalize the handshake, the agent locally expands the client finished key $\textsf{fk}_c$ and generates the Client Finished ($\textsf{CF}$) message using its exclusive hold of the complete $\textsf{CHTS}$. The ARI then forwards the $\textsf{CF}$ message to the tool to verify session integrity. 
% If the tool requires client authentication, the agent can encrypt and forward its certificate to the tool via the ARI as a blind relay.
To finalize the handshake, the agent locally expands the client finished key $\textsf{fk}_c$ using the reconstructed $\textsf{CHTS}$.
If the service provider requires client authentication, the agent first encrypts and sends the required client-authentication messages, such as its Certificate and CertificateVerify, through ARI as a blind relay.
The agent then generates the Client Finished ($\textsf{CF}$) message, which ARI forwards to the service provider to complete the TLS handshake.
% and confirm the integrity of the client-side transcript.

% At the end of the handshake, the application-layer secrets are distributed to establish the following security state: the client-side key $\textsf{tk}_\textsf{capp}$ remains secret-shared between the agent and ARI to enforce collaborative query encryption, while the server-side key $\textsf{tk}_\textsf{sapp}$ is exclusively held by the agent to ensure response confidentiality and independent decryption. We leave the detailed ARI-adapted three-party handshake protocol $\Pi_\textsf{ths}$ in Figure~\ref{prot:handshake} of Appendix~\ref{sec:appendix:handshake}.
% Crucially, the handshake protocol is a \emph{one-time cost} executed prior to runtime operations, allowing these TLS secrets to be cached and reused across subsequent queries. 
% Furthermore, this set can be executed in \emph{parallel across endpoints} when the agent should interact with multiple service providers. This ensures the cryptographic overhead of the handshake is thoroughly amortized and completely decoupled from the latency-sensitive runtime interaction loop.
\parab{Summary} At the end of the handshake, the application traffic keys establish the following security state: the client-side key $\textsf{tk}_\textsf{capp}$ remains secret-shared between the agent and ARI to enable collaborative request encryption, while the server-side key $\textsf{tk}_\textsf{sapp}$ is reconstructed only by the agent to ensure response confidentiality and integrity. %and independent decryption.
We provide the detailed ARI-adapted three-party handshake protocol $\Pi_\textsf{ths}$ in Appendix~\ref{sec:appendix:handshake}.
\revision{Crucially, the handshake is a \emph{one-time per-endpoint cost} incurred before runtime service invocations, allowing the TLS channel to be retained and reused across subsequent requests.
When the agent needs to interact with multiple service providers, these per-endpoint setups are executed in parallel.}

%% file: algo/fig-template.tex
\begin{Protocol}[!t]
    \begin{mdframed}[style=ProtocolFrame, align=center, font=\small]
        
        \begin{enumerate}[label=\textbf{\scriptsize \arabic*}, leftmargin=2.5ex,itemsep=0.25ex]
            \item[] \textbf{Protocol} $(\langle s_m \rangle, \langle g_m \rangle, h_m)$ $\gets$ $\Pi_\textsf{SHC}$($\langle s_l \rangle$, $\langle g_l \rangle$, $h_l$, $\langle s_r \rangle$,  $\langle g_r \rangle$,  $h_r$)
                  % \vspace{0.1cm}
                  \hrule
                  % \vspace{0.1cm}
            \item[] \textit{Input.} Two-party bitwise secret-shares of the left string $\langle s_l \rangle$, 
            the right string $\langle s_r \rangle$, 
            and arithmetic shares of the length of $s_l$'s content $\langle g_l \rangle$, the length of $s_r$'s content $\langle g_r \rangle$, respectively.
             The length of the padded string is $h_l$ ($h_r$), which is public for both parties.
            \item[] \textit{Output.} A secret-shared string $\langle s_m \rangle$ and its actual length $\langle g_m \rangle$ and public padded length $h_m$.
                  % \vspace{0.005cm}
                  \hrule
                  \vspace{0.005cm}
            \item $k \gets \lceil\log_2(h_l+h_r)\rceil$ \anno{Bitwidth.}
            \item 
            $\langle u_l \rangle \gets \langle s_l \rangle \parallel 0^{\,h_l - g_l}$;
            $\langle u_r \rangle \gets \langle s_r \rangle \parallel 0^{\,h_r - g_r}$. 
            \item 
            $ \langle v_a \rangle \gets \langle u_l \rangle \parallel 0^{\,h_r}$;$\langle v_0 \rangle \gets 0^{\,h_l} \parallel \langle u_r \rangle$.
            
            \item\label{itm:optm-sub} 
                  $\langle d \rangle_{\mathsf{arith}} \gets h_l - \langle g_l \rangle $; \anno{Locally compute shares.}
                  % \hfill{\anno{Form rotation offset in $\mathbb{Z}_{2^k}$.}}
            \item\label{itm:optm-a2b} $\langle d \rangle_{\mathsf{xor}} \gets \mathcal{F}_\textsf{A2B}(\langle d \rangle_{\mathsf{arith}}, k)$; \anno{Convert $d$ to XOR-share.}
            \item\label{itm:optm-rotate} $\langle v_b \rangle \gets \Pi_\textsf{BlindRotate}(\langle v_0 \rangle, \langle d \rangle_{\mathsf{xor}}, k)$; \anno{OT-based left shift.}
            % \item\label{itm:optm-sm} $\langle s_m \rangle \gets \langle v_a \rangle \oplus \langle v_b \rangle$; \anno{Local byte-wise/XOR merge of two masks.}
            % \item\label{itm:optm-gm} $\langle g_m \rangle \gets \langle g_l \rangle + \langle g_r \rangle \bmod 2^k$; \anno{Local arithmetic addition.}
            % \item $h_m = h_l + h_r$
            \item $\langle s_m \rangle \gets \langle v_a \rangle \oplus \langle v_b \rangle$;\ $\langle g_m \rangle \gets \langle g_l \rangle + \langle g_r \rangle$;\ $h_m = h_l + h_r$.
        \end{enumerate}

    \end{mdframed}
    % \caption{The OT-based Structure-hidden Concatenation.}
    \caption{The structure-hiding concatenation protocol.}
    \label{prot:concat-optm}
\end{Protocol}

%% file: algo/fig-query.tex
\begin{Protocol}[!t]
  \begin{mdframed}[style=ProtocolFrame, align=center, font=\small,userdefinedwidth=\columnwidth]
    \begin{enumerate}[label=\textbf{\scriptsize \arabic*}, leftmargin=2.5ex,itemsep=0.25ex]
      \item[] \textbf{Protocol} $(\hat{Q},\sigma) \gets \Pi_\textsf{query}(\mathcal{T}, \{E\}, \Omega, \langle \textsf{tk}_\textsf{capp} \rangle)$
      \vspace{0.1cm}
      \hrule
      \vspace{0.1cm}

      % \item[] \textit{Input.} \mcpclient and \mcpserver agree on a public template $\mathcal{T} = [C_1, \dots, C_L]$, where each chunk $C_i = (o_i, h_i)$ specifies its owner $o_i$ and padded length $h_i$. 
      \item[] \textit{Input.} \mcpclient and \mcpserver agree on a public template $\mathcal{T} =[\tau_1,\ldots,\tau_L]$, where each segment descriptor 
        \(\tau_i=(\theta_i,h_i)\) specifies the segment category and padded length. 
        The parties prepare initial segments  $\{E\} = [E_{[1,1]}, \dots, E_{[L,L]}]$, where each $E_{[i,i]} = (\theta_i, \langle s_i \rangle, \langle g_i \rangle, h_i)$ contains the padded content and its actual length.
        They hold an assembly plan $\Omega = [\omega_1, \omega_2, \dots]$ where each step $\omega = (i, k, j)$ defines a concatenation: $E_{[i,j]} = E_{[i,k]} \boxplus E_{[k+1,j]}$, merging the previously assembled range $[i, k]$ with $[k+1, j]$.
        They shared the session key $\langle \textsf{tk}_\textsf{capp} \rangle$.

      \item[] \textit{Output.} The encrypted query $\hat{Q}$ and its GMAC tag $\sigma_Q$.
      % , delivered to \toolserver by \mcpserver.

      \hrule

      \item\label{itm:mcpq-for} \textbf{for} each $\omega=(i,k,j)\in\Omega$ \textbf{do:}
      \anno{Iterative assembly.}

      \item\quad $E_l = E_{[i,k]}=(\theta_l,\langle s_l\rangle,\langle g_l\rangle,h_l)$.
      % \item \quad 
      \item\quad $E_r = E_{[k+1,j]}=(\theta_r,\langle s_r\rangle,\langle g_r\rangle,h_r)$.
      % \hfill{\anno{Load operands.}}

      \item\label{itm:mcpq-if-ldc} \quad \textbf{if} 
      % $g_l$ is a public metadata 
      $\theta_l \in \{\public, \mcpclient_f, \mcpserver_f,\shared_f\}$
      \textbf{then}
      \anno{$\boxplus$ instantiates $\Pi_\textsf{LDC}$.}

      \item\label{itm:mcpq-ldc} \quad\quad
      $\langle s_m\rangle \gets (\langle s_l \rangle \| \textsf{0x00}^{h_r}) \oplus (\textsf{0x00}^{h_l} \|\langle s_r \rangle $).
      % \langle s_l\rangle \,\|\,(\langle s_r\rangle \ll h_l)$;\;
      \item \quad\quad
      $\langle g_m\rangle \gets g_l+\langle g_r\rangle$ ;\; $h_m \gets h_l + h_r$.
      % \hfill{\anno{Local update.}}

      \item\label{itm:mcpq-else} \quad \textbf{else}
      \anno{$\boxplus$ instantiates $\Pi_\textsf{SHC}$.}

      % \item\label{itm:mcpq-shc-d} \quad\quad
      % $\langle d\rangle \gets h_l-\langle g_l\rangle$.
      % \hfill{\anno{Secret gap.}}
      \item \quad \quad $(\langle s_m \rangle, \langle g_m \rangle,h_m) \gets \Pi_\textsf{SHC}(\langle s_l \rangle, \langle g_l \rangle, h_l, \langle s_r \rangle,  \langle g_r \rangle,  h_r)$

      \item\label{itm:mcpq-set} \quad
      $E_m = E_{[i,j]}\gets(\theta_m,\langle s_m\rangle,\langle g_m\rangle,h_m)$.
      % \hfill{\anno{Write back.}}

      % \item\label{itm:mcpq-endfor} \textbf{end for}

      \item\label{itm:mcpq-Q} $\langle Q\rangle \gets E_{[1,L]}.\langle s\rangle$.
      \anno{Final content share.}

      \item\label{itm:mcpq-aead} $(\hat{Q},\sigma_Q)\gets \Pi_\textsf{2PC-AEAD}(\langle Q\rangle,\langle \textsf{tk}_\textsf{capp}\rangle)$.
      \anno{AEAD.}

      % \item\label{itm:mcpq-send} \mcpserver $\rightarrow$ \toolserver: $\hat{Q}$.
      % \hfill{\anno{Send ciphertext.}}

    \end{enumerate}
  \end{mdframed}
  \caption{\revision{The privacy-preserving query-construction.}}
  \label{prot:mcpquery}
\end{Protocol}

%% file: algo/fig-circuit.tex
\begin{figure}[t]
    \centering
    \begin{mdframed}[userdefinedwidth=\columnwidth]
    \begin{tikzpicture}
    \node[inner sep=0pt, text width=\columnwidth, align=justify] (content) {
    \small
    \setlength{\parindent}{0pt}
    \setlist[itemize]{leftmargin=*, label=$\bullet$, itemsep=0pt, parsep=0pt, topsep=1pt}

    \underline{{\bf Circuit} $\textsf{Ckt}(\mathbb{X}, \mathbb{W}):$}
\begin{itemize}
    \item $\mathbb{X} = (\textsf{tk}^\textsf{R}_\textsf{sapp}, \textsf{com}_A, \hat{R}, r, \varphi, v, \omega, d_t, j, k)$
    \item $\mathbb{W} = (\textsf{tk}^\textsf{A}_\textsf{sapp}, R_{\text{local}})$
    
    \item $\textsf{com}_A = \textsf{H}(\textsf{tk}^\textsf{A}_\textsf{sapp}, r)$, $\textsf{tk}_\textsf{sapp} = \textsf{tk}^\textsf{A}_\textsf{sapp} \oplus \textsf{tk}^\textsf{R}_\textsf{sapp}$
    \item $\textsf{Dec}_{\textsf{tk}_\textsf{sapp}}(\hat{R}, \omega) = R_{\text{local}}$, $|R_{\text{local}}| = \mathcal{L}_{\text{local}}$
    
    \item Set $b_0 \leftarrow 0$, $e_0 \leftarrow 0$, $d_0 \leftarrow 0$ \anno{string, escape, depth}
    
    \item $\forall i \in [1, \mathcal{L}_{\text{local}}]$ where $c_i \leftarrow R_{\text{local}}[i]$:
    \item[] $\kappa_{123} \leftarrow \mathbf{1}_{\{c_i = 123\}}$, $\kappa_{125} \leftarrow \mathbf{1}_{\{c_i = 125\}}$ \anno{Track `\{' and `\}'}
    \item[] $\kappa_{34} \leftarrow \mathbf{1}_{\{c_i = 34\}}$, $\kappa_{92} \leftarrow \mathbf{1}_{\{c_i = 92\}}$ \anno{Track `\texttt{"}' and `\texttt{\textbackslash}'}
    \item[] $e_i = (1 - e_{i-1}) \cdot b_{i-1} \cdot \kappa_{92}$ \anno{Track escape state}
    \item[] $b_i = e_{i-1} \cdot b_{i-1} + (1 - e_{i-1}) \cdot [b_{i-1} \cdot (1 - \kappa_{34}) + (1 - b_{i-1}) \cdot \kappa_{34}]$ \anno{Track string boundary}
    \item[] $d_i = d_{i-1} + (1 - b_{i-1}) \cdot (\kappa_{123} - \kappa_{125})$ \anno{Track depth}

    \item Enforce \(R_{\text{local}}[j \dots k]\) encodes field \(\varphi:v\) at depth \(d_t\).
\end{itemize}

    };
    \end{tikzpicture}
    % }
    \end{mdframed}
    % \caption{\revision{Arithmetic Constraint Specification for the Verifiable Billing Circuit $\textsf{Ckt}$.}}
    \caption{Arithmetic constraint specification for the verifiable billing circuit \(\textsf{Ckt}\) for a forward-parsing window.}
    \label{fig:bill}
\end{figure}

%% file: 5.evaluation.tex
\section{Evaluation}
\label{sec:eval}

Our evaluation is organized into two complementary dimensions: a formal security analysis of our protocols and a comprehensive experimental study.

\parab{Security Analysis} We integrate the handshake (§~\ref{sec:handshake}), query  (§~\ref{sec:query}), and billing (§~\ref{sec:response}) protocols into a unified protocol $\Pi_{\textsf{sys}}$. 
% We formally prove that $\Pi_{\sys}$ securely realizes the ideal functionality $\mathcal{F}_{\sys}$ in the $(\mathcal{F}_\textsf{ths}, \mathcal{F}_\textsf{query}, \mathcal{F}_\textsf{bill})$-hybrid model against a semi-honest adversary corrupting at most one party. 
% This is established through a standard hybrid argument, where the real phase transcripts are progressively replaced with simulated transcripts guaranteed by their corresponding realization theorems to demonstrate computational indistinguishability. 
\revision{
% Following the real-ideal paradigm, we formally prove that $\Pi_{\textsf{sys}}$ securely realizes the ideal functionality \fsys against a semi-honest adversary.
Due to the page limit, we provide a proof sketch that $\Pi_{\textsf{sys}}$ securely realizes the ideal functionality \fsys against a semi-honest adversary.}
We defer the complete protocol specification and the security proof to Appendix~\ref{sec:integrated}.

\parab{Experimental Evaluation} To evaluate the performance and deployability of \sys, we conduct extensive experiments on both real-world and synthetic service APIs. Our evaluations center around the following questions: the efficiency of our three core components—the ARI-adapted handshake (\textbf{RQ1}), the privacy-preserving query-construction protocol (\textbf{RQ2}), and the verifiable billing protocol (\textbf{RQ3}), as well as 
% \revision{the readily-deployability of \sys in real-world agentic AI ecosystems (\textbf{RQ4})}.
whether \sys is readily deployable without modifying service providers (\textbf{RQ4}).
% \revision{We release the artifact via an anonymous repository\footnote{https://anonymous.4open.science/r/trustedari-open-science-F7F8}.}

% % Focusing on our core contributions, this section extensively evaluates \sys from the following aspects:
% Our evaluations center around the following questions: 

% % \textbf{RQ1: The landscape of privacy risks in the MCP ecosystem:} What is the prevalence and severity of potential privacy leakages across MCP servers during both the authentication and tool-invocation stages?

% \textbf{RQ1 The efficiency of our MCP-adapted key schedule:} How does our MCP-adapted key scheduling compare to general three-party handshake protocols during connection setup in terms of running time and communication overhead?

% \textbf{RQ2 The efficiency of our privacy-preserving query protocol:} What is the performance breakdown of our protocol on real-world tool call APIs, and how do arithmetic sharing and the optimized segment concatenation plan contribute to its concrete efficiency?

% \textbf{RQ3 The efficiency of our verifiable billing protocol:} How do our ZKP-friendly circuits perform on real-world APIs compared to traditional oracle systems, and what is the specific efficiency of the JSON payload processing module?

% \textbf{RQ4 The deployability of \sys:} Is \sys readily deployable so that the integration of \sys with real-world LLMs will not affect the efficiency of tool calling compared to the standard MCP protocol?

\subsection{Implementation \& Setup}

% \subsubsection{Implementation}
\parab{Implementation}
\label{subsec:eval:impl}
% We implement a prototype of \sys based-on DiStefano's~\cite{celi2023distefano}. We extend about 9500 lines of C++ and 6500 lines of GoLang. Our system is built on the TLS 1.3 stack. The three-party handshake and privacy-preserving query protocol are developed in C++ using EMP-toolkit~\cite{emp-toolkit} with the integration of BoringSSL~\cite{boringssl}. The verifiable billing protocol is implemented on the Gnark library~\cite{gnark} with the Plonk proving scheme~\cite{gabizon2019plonk}. To ensure a fair comparison, we re-implement the non-open-sourced DECO~\cite{zhang2020deco} protocol and migrate the jsnark-based~\cite{jsnark} ZKMB~\cite{grubbs2022zero} to the same Plonk framework based on Ganrk library, eliminating performance discrepancies caused by differing cryptographic backends.
% \revision{We integrate our handshake (\S~\ref{sec:handshake}), query construction (\S~\ref{sec:query}), and verifiable billing (\S~\ref{sec:response}) protocols into the unified \sys protocol $\Pi_\sys$, as detailed in Appendix \ref{sec:integrated}.}
We implement a prototype of \sys based on DiStefano~\cite{celi2023distefano} and FreeAuth~\cite{fang2024freeauth}. 
We extend the codebase by approximately 9,000 lines of C++ and 3,000 lines of Go code. 
Our system is built on a TLS 1.3 protocol stack. The handshake (\S \ref{sec:handshake}) and query-construction (\S \ref{sec:query}) protocols are implemented in C++ using EMP-toolkit~\cite{emp-toolkit}, and integrated with BoringSSL~\cite{boringssl}.
\revision{
We reproduce its pre-computation and reuse optimization~\cite{xie2024lightweight} since the official implementation is closed-source.
% We implement and integrate the closed-source pre-computation and reuse strategy~\cite{xie2024lightweight}. 
Our query-construction protocol (\S \ref{sec:query}) incorporates IKNP OT extension~\cite{keller2015actively}, which avoids heavy preprocessing and decreases communication overhead.}
\revision{The verifiable billing protocol (\S \ref{sec:response}) is implemented using the Gnark library~\cite{gnark} with the Plonk~\cite{gabizon2019plonk} proving scheme over the BN254 scalar field, utilizing the ZK-friendly MiMC~\cite{albrecht2016mimc} hash function.}
% For a fair comparison, we re-implement the closed-source DECO~\cite{zhang2020deco} and port the jsnark-based~\cite{jsnark} ZKMB~\cite{grubbs2022zero} implementation to the same Plonk framework based on the Gnark library, thereby eliminating performance discrepancies due to different cryptographic backends.
\revision{
% For a fair comparison, we re-implement the closed-source DECO~\cite{zhang2020deco} and port the jsnark-based~\cite{jsnark} ZKMB~\cite{grubbs2022zero} implementation to the same Plonk framework based on the Gnark library. This baseline mirrors state-of-the-art oracle deployments where ZKMB proves packet headers and DECO proves payloads. 
We construct a composite baseline from TLS oracle systems: DECO~\cite{zhang2020deco} for proving TLS payloads and ZKMB~\cite{grubbs2022zero} for proving packet headers. 
Specifically, we re-implement the closed-source DECO and port the jsnark-based ZKMB implementation to the same Gnark/Plonk framework for a fair comparison.
}

% baselines

% \subsubsection{Datasets} 
% \parab{Datasets}
% %To comprehensively evaluate the performance of \sys, 
% We construct two datasets to evaluate \sys: a real-world tool-call API dataset and a synthetic API dataset. %for more fine-grained benchmarking.

\emph{Real-world Service Dataset.} 
% We select 10 diverse and representative service APIs used in ARI-mediated workflows, spanning multiple categories such as source code management, database retrieval, and LLM inference, including GitHub, Google, OpenAI.
We select 10 diverse and representative service APIs used in ARI-mediated workflows, including GitHub, Google, and OpenAI, and covering categories such as source code management, database retrieval, and LLM inference.
\revision{The dataset has an average query length of 542.8 bytes, ranging from 3 to 49 segments (averaging 20 segments per query); and an average response length of 723.4 bytes. }
%providing a realistic testbed for evaluating the end-to-end efficiency of our query and billing protocols.
This dataset is used to evaluate the end-to-end efficiency of our query construction and billing protocol in \sys. 
A detailed breakdown of the API categories is provided in Appendix~\ref{sec:appendix:api}.
\parab{Setup}
\label{subsec:eval:setup}
Our evaluation environment is a Linux server equipped with a multi-core x86\_64 Intel CPU at 2.60GHz. 
% The network environments, \ie bandwidth and latency of multiple parties, are simulated using the \textsf{tc} command and network namespace provided by Linux. The LAN environment is of $(1000Mbps, 1ms)$, while the WAN environment is of $(200Mbps, 10ms)$. Each experiment is run at least five times, and we take the average of the running durations.
% To evaluate the performance across various network conditions, we leverage the Linux tc command and network namespaces to simulate three distinct environments: (i) Network I: $5000\text{ Mbps}$ bandwidth with negligible latency; (ii) Network II: $1000\text{ Mbps}$ bandwidth with $1\text{ ms}$ latency; and (iii) Network III: $200\text{ Mbps}$ bandwidth with $10\text{ ms}$ latency. All results are averaged over five independent runs to minimize measurement noise.
To evaluate performance under diverse conditions, \revision{we utilize \textsf{tc} and network namespaces to simulate several environments: Network I (5000 Mbps, $\leq$ 1 ms), Network II (1000 Mbps, 1 ms), Network III (200 Mbps, 10 ms), and Network IV (200 Mbps, 50 ms).} Results are averaged over five runs to minimize measurement noise.

% \subsection{Q1: Potential Privacy Leakages in MCP Servers Ecosystem}

\subsection{RQ1: Efficient Three-Party Handshake}
\label{subsec:q1}

\begin{table}[t]
    \centering
    \begin{adjustbox}{max width=\columnwidth}
    \begin{tabular}{lcccc}
    \toprule
        & \multicolumn{2}{c}{\textbf{DiStefano~\cite{celi2023distefano}}}
        & \multicolumn{2}{c}{\textbf{\sys}} \\
    \cmidrule(lr){2-3} \cmidrule(lr){4-5}
        & \textbf{Agent-ARI} % \textbf{Agent-to-Server} 
        & \makecell[c]{\textbf{TLS} \\ \textbf{Online}} %\textbf{TLS Online} 
        % & \textbf{Total}
        & \textbf{Agent-ARI} % \textbf{Agent-to-Server} 
        & \makecell[c]{\textbf{TLS} \\ \textbf{Online}} \\ %\textbf{TLS Online} 
        % & \textbf{Total} \\
    \midrule

    Network I (s) 
          & $12.27 \pm 0.34$ & $1.14 \pm 0.09$ %& $13.41$
          & $5.73 \pm 0.10$ & $1.00 \pm 0.10$ \\%& $6.73$ \\ 

    Network II (s)   
          & $14.14 \pm 0.42$ & $1.32 \pm 0.11$ %& $15.46$
          & $6.46 \pm 0.10$ & $1.20 \pm 0.04$ \\%& $7.66$ \\ 

    Network III (s)   
          & $24.64 \pm 0.14$ & $3.51 \pm 0.08$ %& $28.16$
          & $13.52 \pm 0.28$ & $3.38 \pm 0.10$ \\%& $16.90$ \\ 

    Network IV (s)   
          & $41.79 \pm 0.41$ & $8.13 \pm 0.07$ %& $49.91$
          & $28.18 \pm 0.47$ & $7.40 \pm 0.07$ \\%& $35.58$ \\ 

    Comm. (MB)  
          & $220.48$ & $0.39$ %& $220.87$
          & $133.62$ & $0.36$ \\%& $133.98$ \\ 
    \bottomrule
    \end{tabular}
    \end{adjustbox}
    % \caption{\revision{End-to-end running time and communication of various schemes in the Three-party Handshake protocol.}}
    \caption{Phase-wise running time and communication overhead of the three-party handshake protocol.}
    \label{tab:cost-handshake}
\end{table}

To demonstrate the efficiency gains achieved through our ARI-specific handshake design, we evaluate the performance of our ARI-adapted key scheduling against the state-of-the-art general third-party TLS handshake protocol DiStefano~\cite{celi2023distefano}. Table~\ref{tab:cost-handshake} summarizes the running time and communication across various network environments. 
% Compared to DiStefano, \sys reduces the running time by approximately 57\%, and this reduction ratio remains stable across three network settings. Regarding communication overhead, \sys achieves a 39.50\% reduction, decreasing from 220.87 MB to 133.62 MB. The superior efficiency of \sys stems from replacing expensive two-party computation with local key generation through strategic key scheduling while employing a precomputation-and-reuse mechanism to eliminate redundant computation costs during the key expansion.
We benchmark the protocol across two phases: (i) the \textit{Agent-ARI} phase, which captures the precomputation between the Agent and ARI and can be executed asynchronously before the TLS Server is involved, and (ii) the \textit{TLS Online} phase, which represents the synchronous cryptographic operations requiring real-time interaction with the service provider. 

As shown in Table~\ref{tab:cost-handshake}, our ARI-specific handshake protocol achieves a substantial latency reduction ranging from $32.57$\% to $53.32$\% in the Agent-ARI phase, 
translating to an end-to-end speedup of $28.72\%$ to $50.47\%$ across various network settings.
Regarding the TLS Online phase, the online overhead ranges from $1.00$ to $7.40$ seconds, 
% which adheres to the standard industrial TLS handshake timeout threshold (typically $10$ seconds). 
which remains below a 10-second practical timeout budget for online TLS connection establishment.
Furthermore, \sys reduces the total communication cost by 39.34\%, from $220.87$ MB to $133.98$ MB. 

\revision{
% We emphasize that the handshake overhead constitutes a one-time setup cost during tool registration. When the agent initializes, the handshakes for multiple service providers can be processed in parallel. 
We clarify that the handshake is a one-time per-endpoint cost incurred before runtime service invocation, and setups for different providers run in parallel.
Once the connection is established, subsequent queries can reuse the established TLS channel.}

\subsection{RQ2: Efficient Query Construction}
\label{subsec:q2}

\subsubsection{Performance on Real-world Services}
% We evaluate the performance of our privacy-preserving protocol using 10 representative real-world tool call APIs. Table~\ref{tab:api:query} summarizes the breakdown of running time and communication into structure-hidden and content-hidden processing. 
We evaluate the performance of our privacy-preserving query-construction protocol on our real-world service dataset. Table~\ref{tab:api:query} summarizes the breakdown of local computation time and communication cost into structure-hiding and content-hiding processing, excluding network latency. 
Specifically, the structure-hiding processing refers to the iterative assembly of all segments $\{E\}$, after which parties extract the final secret-shared query string $\langle Q \rangle$ from the content part of the last segment $E[1,L]$. In contrast, the content-hiding processing involves the subsequent encryption of $\langle Q \rangle$ into ciphertext $\hat{Q}$ and a GMAC $\sigma_Q$ using a 2PC-AEAD protocol, 
% which introduces identical overhead to the general oracle systems and the naive method. 
which is the same 2PC-AEAD step required by general TLS-oracle systems and the content-only baseline.
% The results indicate that, compared to the naive method that only protects query content, our design, which provides comprehensive protection for both content and structure, introduces only minimal additional cost. Specifically, the structure-hidden part accounts for an average of only 0.49 seconds running time (27.8\% of total time) and 0.58 MB communication (1.36\% of total communication). This demonstrates that our protocol achieves high-level structural privacy with near-zero marginal overhead relative to naive content-hidden encryption.
The results indicate that, compared to a naive approach that only protects query content, our design introduces modest additional cost to provide comprehensive protection for both content and structure. Specifically, the structure-hiding part accounts for an average of only 0.19 seconds of computation time (14.29\% of total time) and 0.58 MB of communication (1.36\% of total communication). 
% This demonstrates that our protocol achieves high-level structural privacy with acceptable marginal overhead relative to naive content-hidden encryption.
These results show that hiding query structure incurs only a small marginal overhead beyond content-hiding encryption.

\begin{table}[t]
  \centering
  \begin{adjustbox}{max width=\columnwidth}
  \begin{tabular}{lcccccc} % 1 + 3 + 3 = 7 列
    \toprule
    \multirow{2}{*}{\textbf{ID}} & \multicolumn{3}{c}{\textbf{Computation Time (s)}} & \multicolumn{3}{c}{\textbf{Communication (MB)}} \\ % 删除了末尾多余的 &
    \cmidrule(lr){2-4} \cmidrule(lr){5-7} 
    & \textbf{Structure} & \textbf{Content} & \textbf{Total} & \textbf{Structure} & \textbf{Content} & \textbf{Total}  \\
    \midrule
    1  & $0.07$ & $0.96$ & $1.03$  & 0.19& 36.92& 37.11\\
    2  & $0.18$ & $0.91$ & $1.09$  & 0.47 & 33.04 & 33.52\\
    3  & $0.46$ & $0.97$ & $1.43$  & 1.39 & 37.08 & 38.47\\
    4  & $0.07$ & $0.46$ & $0.53$  & 0.16 & 18.86 & 19.03\\
    5  & $0.18$ & $1.52$ & $1.70$  & 0.54 & 54.10 & 54.64\\
    6  & $0.15$ & $0.81$ & $0.96$  & 0.63 & 29.79 & 30.42\\
    7  & $0.18$ & $1.44$ & $1.62$  & 0.66 & 55.14 & 55.80\\
    8  & $0.21$ & $1.53$ & $1.74$  & 0.56 & 57.54 & 58.10\\
    9  & $0.30$ & $1.42$ & $1.72$  & 0.87 & 49.30 & 50.17\\
    10 & $0.14$ & $1.34$ & $1.48$  & 0.38 & 48.86 & 49.24\\
    \midrule 
    \textbf{AVG} & $0.19$ & $1.13$ & $1.32$ & 0.58 & 42.06& 42.65\\ 
    \bottomrule
  \end{tabular}
  \end{adjustbox}
  % \caption{\revision{Average computation time per party and communication breakdown of $\Pi_\textsf{query}$ for various tool APIs.}}
  \caption{Per-party computation time and communication cost breakdown of $\Pi_\textsf{query}$ on real-world downstream APIs.}
  \label{tab:api:query}
\end{table}

\begin{figure}[t]
    \centering
    \includegraphics[width=0.8\columnwidth]{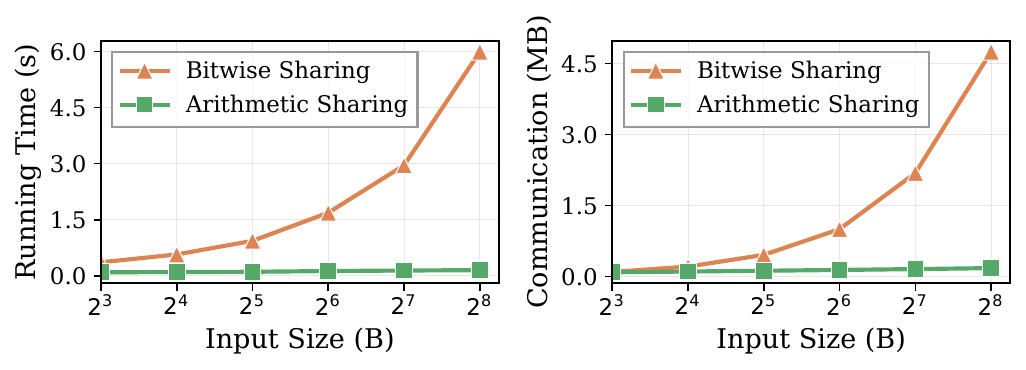}
    % \vspace{-6pt}
    \caption{Running time and communication of $\Pi_\textsf{SHC}$ under arithmetic and bitwise sharing (in Network II).}
    \label{fig:efficiency:concat}
\end{figure}
\begin{figure}[t]
    \centering
    \includegraphics[width=0.8\columnwidth]{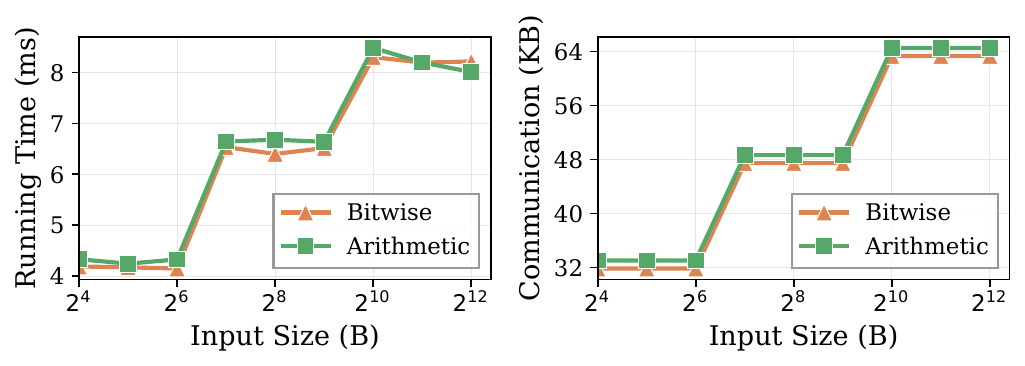}
    % \vspace{-6pt}
    \caption{Running time and communication of $\Pi_\textsf{I2S}$ under arithmetic and bitwise sharing (in Network I).}
    \label{fig:efficiency:postprocess}
\end{figure}

\subsubsection{Impact of Arithmetic Sharing} 
\label{subsubsec:arith}
To validate our design choice of representing the content length as arithmetic shares (as discussed in \S~\ref{subsec:query:optm}), we compare the performance of $\Pi_\textsf{SHC}$ with a baseline that employs bitwise sharing. 
Our arithmetic sharing variant incorporates byte-level processing and IKNP~\cite{keller2015actively} OT extension to decrease communication.
Figure~\ref{fig:efficiency:concat} shows the running time and communication with varying input sizes (from 8 to 256 bytes) in Network II. We set $\alpha = 0.1$, meaning that the padding occupies 10\% of the total input length. 

Compared to the bitwise sharing scheme, when the input size is 8 bytes, 
% arithmetic sharing reduces the running time by 3.87$\times$ (from 0.360 seconds to 0.093 seconds).
arithmetic sharing reduces the running time from $0.360$ seconds to $0.093$ seconds, a $3.87\times$ improvement.
When the input size increases to 256 bytes, arithmetic sharing reduces the running time by 39.96$\times$ from 5.995 seconds to 0.150 seconds. As for the communication, arithmetic sharing reduces the baseline’s overhead by 1.09$\times$ when the input size is 8 bytes, and by 26.85$\times$ when the input size reaches 256 bytes. 
% The reduction ratio of both the running time and communication grows with the input size.
% increases drastically as the input size grows. 
Both reductions grow with the input size.
% This is in conformity with our complexity analysis that the cost of bitwise sharing scales $\mathcal{O}(n\log n)$ with the input string length, whereas our arithmetic sharing-based design achieves a much more efficient logarithmic growth, validating its superior scalability for handling large-scale private inputs.

% \begin{figure}[t!]
%     \centering
%     \includegraphics[width=\columnwidth]{figure/postprocess_hin_sweep.pdf}
%     % \vspace{-6pt}
%     \caption{Running time and communication of the I2S protocol when sweeping the bitwidth $k$ with a fixed input string length $h_{in}$ (in Network II).}
%     \label{fig:efficiency-scale}
% \end{figure}

Figure~\ref{fig:efficiency:postprocess} shows the running time and communication of different sharing schemes for $\Pi_\textsf{I2S}$ when sweeping the input string length (from 16 to 4096 bytes) with a fixed bit width $k=20$. This experiment is performed in Network I to eliminate the disproportionate impact of network latency on millisecond-level operations. 
% Each algorithm is run for 1000 iterations. 
% We observed that the running time and communication exhibit a staircase-like growth. This is because $\Pi_\textsf{I2S}$ converts a secret-shared numerical value into its corresponding decimal ASCII string sharing. The output length $g_{out}$ (\ie the number of digits in $g_{in}$) only increases when $g_{in}$ crosses a power-of-ten threshold (\eg $g_{out} = 2$ for $g_{in} \in [10, 99]$ and $g_{out} = 3$ for $g_{in} \in [100, 999]$). 
We observed a staircase-like growth in both metrics, as the output length $g_{out}$ only increments 
% when the numerical value $g_{in}$ crosses power-of-ten thresholds.
when $g_{\mathsf{body}}$ crosses decimal digit-length thresholds.
Evaluation results indicate that arithmetic sharing maintains comparable performance to the bitwise baseline in $\Pi_\textsf{I2S}$, with a marginal overhead of only 2\% on average.
% In this phase, arithmetic sharing is slightly slower than the bitwise sharing baseline. On average, the running time of arithmetic sharing is 1.02$\times$ that of bitwise sharing.
% and this ratio remains nearly constant regardless of the input size $h_{in}$. Specifically, when $h_{in} = 1024$ bytes, arithmetic sharing requires 8.479 ms and 64.54 KB, while bitwise sharing requires 8.299 ms and 63.35 KB.

Despite this marginal disadvantage in $\Pi_\textsf{I2S}$, arithmetic sharing remains the superior choice for the overall protocol. The millisecond-level cost in $\Pi_\textsf{I2S}$ occurs at most once per invocation and is negligible compared to the substantial gains in multiple $\Pi_\textsf{SHC}$ rounds, where arithmetic sharing yields about $40\times$ speedup and saves megabytes of communication.

\begin{figure}[t]
    \centering
    \includegraphics[width=0.8\columnwidth]{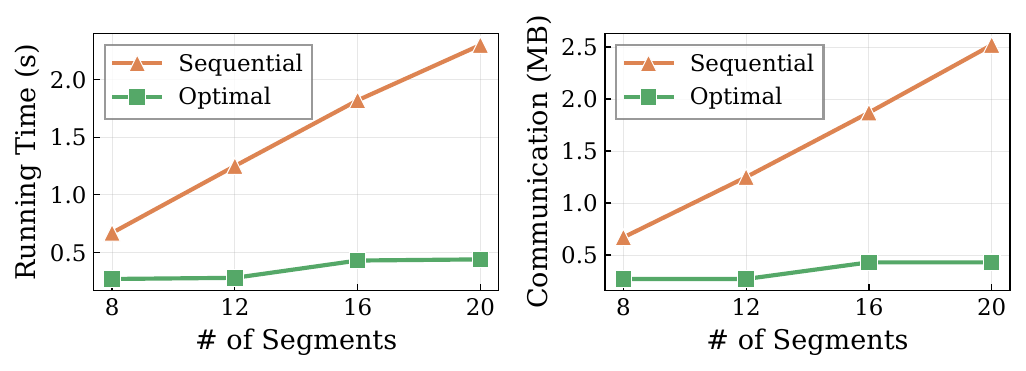}
    % \vspace{-6pt}
    % \caption{Running time and communication of various concatenation plans (in Network II).}
    \caption{Running time and communication cost of sequential and optimized concatenation plans (in Network II).}
    \label{fig:efficiency:dp}
\end{figure}

\begin{table}[t]
  \centering
  % \begin{adjustbox}{max width=\columnwidth}
  
  \begin{tabular}{lcccccc} % 1 + 2*3 = 7 列
    \toprule
    \multirow{2}{*}{\textbf{ID}} & \multicolumn{2}{c}{\textbf{\# of Constraints ($10^6$)}} & \multicolumn{2}{c}{\textbf{Proving Time (s)}} & \multicolumn{2}{c}{\textbf{Verifying Time (ms)}} \\
    \cmidrule(lr){2-3} \cmidrule(lr){4-5} \cmidrule(lr){6-7}
    & \makecell[c]{\textbf{DECO} \\ \textbf{+ZKMB}} & \textbf{Ours} & \makecell[c]{\textbf{DECO} \\ \textbf{+ZKMB}} & \textbf{Ours} & \makecell[c]{\textbf{DECO} \\ \textbf{+ZKMB}} & \textbf{Ours} \\
    \midrule
    1&22.28 &1.57 &108.00 &8.28 &9.08 &5.46 \\
    2&6.53 &0.89 &28.84 &4.17 &4.19 &3.10 \\
    3&8.56 &0.60 &50.43 &3.84 &2.73 &5.57 \\
    4&57.59 &0.80 &215.49 &4.13 &5.49 &2.81 \\
    5&9.33 &0.38 &53.27 &2.15 &6.28 &3.09 \\
    6&1.28 &0.13 &8.61 &0.61 &2.08 &2.11 \\
    7&38.13 &0.47 &236.01 &2.24 &3.64 &2.59\\ 
    8&24.90 &0.39 &125.00 &3.11 &5.01 &3.20 \\
    9&9.88 &0.67 &52.30 &4.40 &5.03 &3.81 \\
    10&27.76 &0.30 &109.04 &2.04 &5.43 &2.64\\ 
    \midrule
    \textbf{AVG} & 20.62 & 0.62 & 98.70 & 3.50 & 4.89 & 3.44\\ 
    \bottomrule
  \end{tabular}
  % \end{adjustbox}
  % \caption{\revision{Number of constraints, proving and verifying time of the baseline method and our $\Pi_\textsf{bill}$ for various APIs.}}
  \caption{Constraint count, proving time, and verification time on real-world downstream APIs.}
  \label{tab:response:zk}
\end{table}

\subsubsection{Effectiveness of Optimized Concatenation Plan} 
%To evaluate the efficacy of our optimization in mitigating the ``structural contamination'' effect, 
We now evaluate the effectiveness of our optimal segment concatenation planning introduced in \S~\ref{subsec:query:optm}. 
We compare our approach %optimal concatenation planning 
against a sequential concatenation plan, where all segments are concatenated strictly following their original sequence. Figure~\ref{fig:efficiency:dp} illustrates the running time and communication for different concatenation plans as the number of segments increases (from 8 to 20) in Network II. To simulate real-world packet structures, 
% we construct requests starting with a sequence of public and structure-hiding segments ($\public, \mcpclient_s, \public, \mcpserver_s$) group, followed by several groups of fixed-length segments ($\public, \mcpclient_f, \public, \mcpserver_f$). 
we construct requests starting with one group of public and structure-hiding segments ($\public, \mcpclient_s, \public, \mcpserver_s$), followed by several groups of fixed-length segments ($\public, \mcpclient_f, \public, \mcpserver_f$).

The core efficiency gain of our optimized plan stems from strategically ordering the assembly of segments to maximize the number of invocations of $\Pi_\textsf{LDC}$ while minimizing the number of $\Pi_\textsf{SHC}$ invocations. 
As shown in Figure~\ref{fig:efficiency:dp}, 
% our optimized plan demonstrates superior scalability as the request complexity grows. 
our optimized plan scales better as the number of segments increases.
Compared to the sequential plan, when there are 8 segments, our concatenation plan reduces the running time 
and communication by 2.48$\times$,
from 0.67 s to 0.27 s and improves communication by 2.48$\times$.
When the number of segments increases to 20, the speedup for running time reaches 5.23$\times$ (from 2.30 s to 0.44 s), while the communication overhead is reduced by 5.86$\times$ (from 2.52 MB to 0.43 MB). These results validate that our segment concatenation plan effectively eliminates redundant 2PC concatenations, 
% ensuring high efficiency even for complex, multi-segment service queries.
maintaining low overhead for complex, multi-segment service queries.

\subsection{RQ3: Efficient Verifiable Billing Protocol}
\label{subsec:q3}

\subsubsection{End-to-End Performance on Real-world APIs}
We evaluate the performance of our verifiable billing protocol using the real-world API dataset. Table~\ref{tab:response:zk} presents a comparative analysis between \sys and a hybrid baseline combining ZKMB~\cite{grubbs2022zero} and DECO~\cite{zhang2020deco}.
% representing existing oracle techniques, where ZKMB~\cite{grubbs2022zero} is employed to prove the packet headers and DECO~\cite{zhang2020deco} is used to prove the payloads.
% Our protocol achieves significant reductions in computational overhead across all metrics by avoiding the heavy ``building'' circuits required by the baseline. On average, \sys reduces the number of constraints by 41.5$\times$ (from 16,162K to 389K). This optimized circuit design translates directly to substantial gains in proof generation: the average proving time is reduced from 81.53 seconds to 2.05 seconds, representing a 39.77$\times$ speedup. Meanwhile, the verifying time remains highly efficient and comparable to the baseline at approximately 2.63 milliseconds. These results demonstrate that \sys can make real-time verifiable billing practical for high-frequency tool calls.
% Our protocol achieves significant reductions in computational overhead across all metrics by avoiding proving the entire packet required by the baseline. 
Our protocol substantially reduces the constraint count and proving time,  by avoiding proving the entire packet required by the baseline, while keeping verification time within a few milliseconds.
On average, \sys reduces the number of constraints by 33.26$\times$, from $20.62 \times 10^6$ to $0.62 \times 10^6$. 
% This optimized circuit design translates directly to substantial gains in
This reduction directly improves
proof generation, where the average proving time is reduced from 98.70 seconds to 3.50 seconds, representing a 28.20$\times$ speedup. Meanwhile, the average verifying time remains efficient at 3.44 milliseconds.
% Crucially, the proof generation could be executed asynchronously and decoupled from the synchronous agent execution flow, thereby introducing zero blocking latency to subsequent requests.}
\revision{In practice, the agent can collect multiple responses and generate their billing proofs asynchronously for batched settlement, so proof generation does not block the active service-invocation path.
% Crucially, the proof generation can be executed asynchronously and kept off the synchronous agent execution path, so it need not block subsequent requests.
}

% \begin{figure}[t!]
%     \centering
%     \includegraphics[width=\columnwidth]{figure/results_analysis.pdf}
% \end{figure}

% \begin{figure}[t!]
%     \centering
%     \includegraphics[width=\columnwidth]{figure/results_analysis_verifying.pdf}
%     % \vspace{-6pt}
%     \caption{Verifying time.}
%     \label{fig:efficiency-scale}
% \end{figure}

\subsubsection{Efficient Proving of JSON Payload} 

% To evaluate the proving circuit efficiency of \sys across diverse scenarios, we constructed a benchmark based on APIBank~\cite{li2023apibank}. We formatted the tool calls results as JSON strings to serve as payloads of the tool response. Specifically, to evaluate proving efficiency under different target locations, we embedded four checkpoints (CP1--CP4) at fixed structural positions within each JSON string. 
% This dataset comprises 388 instances with a mean length of 430 bytes (ranging from 236 bytes to 1,386 bytes). 
% This dataset comprises 378 instances with a mean length of 407 bytes (ranging from 236 bytes to 842 bytes).
% As illustrated in Figure~\ref{fig:dataset:api}, we measured the average byte offsets of these checkpoints from both the forward and backward directions to analyze the transition points for our bidirectional scanning optimization.
To evaluate the proving circuit efficiency across diverse scenarios, we construct a benchmark based on APIBank~\cite{li2023apibank}, where the service responses are formatted as JSON strings. We embed four checkpoints (CP1--CP4) at fixed structural positions within each payload to represent different target locations. 

\begin{figure}[t]
    \centering
    \includegraphics[width=0.8\columnwidth]{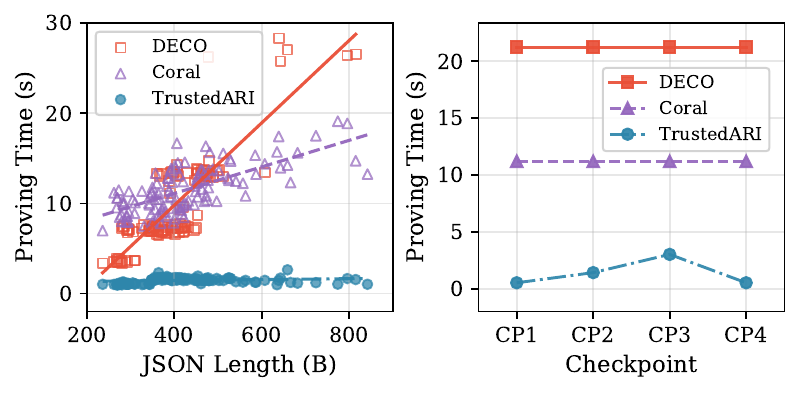}
    % \vspace{-6pt}
    % \caption{\revision{The proving time of various methods on our tool-response benchmark.}}
    \caption{Proving time of \sys and baselines under varying JSON payload lengths and target-field locations.}
    \label{fig:proving}
\end{figure}

% To evaluate the practical impact of our targeted parsing and bidirectional decryption, we benchmarked \sys against the ``building'' circuit baseline. The experimental results in Figure~\ref{fig:proving} highlight the superior efficiency and scalability of our design. Specifically, Figure~\ref{fig:proving} (left) illustrates the proving time relative to the total JSON length, where our scheme exhibits a significantly flatter slope compared to the steep increase observed in the baseline. This confirms that while the baseline's overhead is primarily driven by the total length of the JSON structure, \sys remains resilient to large payloads by only focusing on the claimed key-value pairs. Furthermore, Figure~\ref{fig:proving} (right) demonstrates that our method is highly sensitive to the target's proximity to string boundaries, as evidenced by the X-shaped performance curve of proving time CP1 and CP4. In contrast, the baseline's performance remains nearly constant across all checkpoints because it must exhaustively process the entire JSON regardless of the target's position. By selecting the shortest path via bidirectional scanning, \sys ensures that the verification cost is proportional to the distance from the nearest boundary rather than the total payload length, providing a scalable and high-performance solution for verifiable billing.
We benchmark \sys against two TLS-oracle baselines: DECO~\cite{zhang2020deco} and Coral~\cite{angel2026coral}. Figure~\ref{fig:proving} (left) illustrates the proving time relative to the total JSON length, where each data point represents the average performance across all four checkpoints for a specific payload. 
% As the payload scales, the baselines exhibit a steep linear increase 
As the payload scales, the baselines increase roughly linearly with the payload length
because they exhaustively process the full JSON structure. In contrast, \sys presents a significantly flatter slope, requiring an average of only 1.47 seconds of proving time (ranging from 0.94s to 2.64s). 
% This demonstrates that our scheme remains resilient to large payloads
This shows that the proving cost of \sys grows much more slowly with payload length
by only proving the boundary window containing the targeted fields.

Figure~\ref{fig:proving} (right) plots the average proving time at each specific checkpoint across the dataset to evaluate location sensitivity. For DECO and Coral, the execution curves remain flat across all checkpoints due to their full-parse design. In contrast, \sys exhibits a clear position-dependent curve owing to our adaptive boundary window selection. Consequently, \sys yields a substantial speedup ranging from $3.70\times$ to $20.95\times$ over Coral, and $7.01\times$ to $39.69\times$ over DECO, with the performance gain peaking 
% near the outer string boundaries.
when the target field is close to the beginning or end of the payload.
This confirms that the cryptographic cost scales with the distance to the nearest string boundary rather than the total payload length.

% \begin{figure}[t]
%     \centering
%     \includegraphics[width=\columnwidth]{figure/jsonbench_analysis.pdf}
%     % \vspace{-6pt}
%     \caption{Structural distribution and average byte offsets of checkpoints (CP1--CP4) within our microbenchmark.}
%     \label{fig:dataset:api}
% \end{figure}
% \begin{figure}[t]
%     \centering
%     % \includegraphics[width=\columnwidth]{figure/results_analysis.pdf}
%     \includegraphics[width=\columnwidth]{figure/results_analysis_improved.pdf}
%     % \vspace{-6pt}
%     \caption{The proving time of various methods on our tool-response benchmark.}
%     \label{fig:proving}
% \end{figure}

\subsection{RQ4: Accurate Service-Request Generation}
\label{subsec:q4}

% In this segment, we demonstrate that \sys is fully compatible with the existing LLM-powered agents by showing that state-of-the-art LLMs can accurately generate \sys-compatible tool calls. 

% To evaluate whether \sys-compatible tool calls impact the tool usage of agents, %introduce accuracy degradation, 
% we conduct a comparative study against standard MCP calls. We test three state-of-the-art models: Claude-sonnet-4-5, Gemini-3-Flash, and GPT-5.2. As shown in Table~\ref{tab:llm}, the performance gap between generating standard tool calls and \sys-compatible calls is negligible. For all models, the total accuracy fluctuations remain within approximately 1\%, with observed differences of only 1.89\% for Gemini-3-Flash, 1.55\% for GPT-5.2, and 0.45\% for Claude-sonnet-4-5. 
% Specifically, all models maintain high format accuracy (95.79\%–97.84\%) and semantic accuracy (91.67\%–92.73\%), indicating that the requirement for server-side parameters does not impair the models' reasoning or instruction-following capabilities.
% The adapter success rate is slightly below 100\% because the LLM occasionally generates content exceeding the predetermined maximum length $h$, preventing the adapter from mapping the input into a valid segment. These results confirm that \sys preserves the functional fidelity of LLMs.

% We construct another benchmark based on APIBank~\cite{li2023apibank} to evaluate the impact of LLM performance on generating \sys-compatible requests for Agent Tool Routing.
\revision{For LLM API Routing, generating \sys-compatible input is straightforward because the user prompt itself serves as the agent-owned content to be placed in the public query template. 
% We therefore focus on the more challenging Agent Tool Routing setting, where the LLM must synthesize tool-specific parameters that satisfy the \sys-compatible query template. 
We therefore focus on the more challenging Agent Tool Routing setting, where agents must synthesize tool-specific parameters that satisfy the \sys-compatible query template. 
To evaluate whether this template affects agent compatibility, we construct another benchmark based on APIBank~\cite{li2023apibank}.
% We modified 53 APIs (including 132 input parameters) to match the \sys-compatible query template. 
Our evaluation uses 214 tool-invocation dialogues, each containing 9.98 turns on average, covering 53 APIs and 132 input parameters adapted to the \sys-compatible query template.}
Specifically, we partition the parameters based on data ownership: 22 parameters are designated as ARI-owned, where the agent is expected to output null values, while the remaining 110 parameters are agent-owned, requiring the agent to generate semantic values. 
% For each parameter, we define a maximum length constraint $h$ to simulate the segmented query structure. 

% ——deployable as a local MCP server or an Anthropic Skill~\cite{agent_skill}——
We also implement an adapter in form of a   skill~\cite{agent_skill}
% to transform LLM-generated general requests into \sys-compatible inputs. 
that maps agent-generated \sys-compatible requests into the protocol-level input representation.
Specifically, this adapter maps each parameter into a 4-tuple segment $E_i = \{\theta_i,s_i,g_i,h_i\}$ (described in \S~\ref{subsec:query:ours}), which then serves as the input for the privacy-preserving query-construction protocol (\S~\ref{sec:query}), collaboratively executed by the agent and ARI to produce the final encrypted query.
% (\ie each parameter follows the 4-tuple format $E_i = \{\theta_i,s_i,g_i,h_i\}$ described in \S \ref{subsec:query:ours}). In addition, the agent and the MCP server could collaboratively execute our privacy-preserving query construction protocol.

We evaluate the compatibility of \sys with existing agents by comparing the performance of generating \sys-compatible requests against standard requests~\cite{li2023apibank}. 
We use three metrics: \first Format, which assesses whether the generated tool calls strictly adhere to the API definition, including correct parameter names, types, and the mandatory ``null'' assignment for ARI-owned parameters; \second Content, which measures the correctness of generated values relative to the conversation context; and \third Adapter, for \sys-compatible requests, which validates the successful transformation of agent outputs into valid 4-tuple segments. 
% Any invocation of an incorrect API or failure to follow the defined schema fails all metrics.
Any invocation of an incorrect API or failure to follow the defined schema is counted as a failure for all metrics.
% We test on three representative models: Gemini-3-Flash (Gemini), GPT-5.2 (GPT), and Claude-sonnet-4-5 (Claude). 
We evaluate the compatibility using agents powered by three moderately-capable LLMs: Gemini-3-Flash (Gemini), GPT-5.2 (GPT), and Claude-Sonnet-4.5 (Claude). 

\begin{table}[t]
\centering
\begin{adjustbox}{max width=\columnwidth}
\begin{tabular}{lccccc}
\hline
Agent                   & Template  & Format(\%)                                                 & Content(\%)                                               & Adapter(\%)                                                & Total(\%)                                                  \\ \hline
\multirow{2}{*}{\makecell{Gemini\\-based\\Agent}} & \begin{tabular}[c]{@{}c@{}}APIBank\\ \cite{li2023apibank}\end{tabular} & \begin{tabular}[c]{@{}c@{}}95.54\\ $\pm$0.76\end{tabular} & \begin{tabular}[c]{@{}c@{}}88.62\\ $\pm$0.95\end{tabular} & -                                                      & \begin{tabular}[c]{@{}c@{}}88.62\\ $\pm$1.01\end{tabular} \\ \cline{2-6} 
                        & \makecell[c]{\sys- \\ compatible}   & \begin{tabular}[c]{@{}c@{}}96.34\\ $\pm$1.38\end{tabular} & \begin{tabular}[c]{@{}c@{}}91.53\\ $\pm$0.86\end{tabular} & \begin{tabular}[c]{@{}c@{}}97.29\\ $\pm$0.54\end{tabular} & \begin{tabular}[c]{@{}c@{}}90.43\\ $\pm$1.48\end{tabular} \\ \hline
\multirow{2}{*}{\makecell{GPT\\-based\\Agent}} & \begin{tabular}[c]{@{}c@{}}APIBank\\ \cite{li2023apibank}\end{tabular} & \begin{tabular}[c]{@{}c@{}}97.39\\ $\pm$0.22\end{tabular} & \begin{tabular}[c]{@{}c@{}}91.18\\ $\pm$0.37\end{tabular} & -                                                      & \begin{tabular}[c]{@{}c@{}}91.18\\ $\pm$0.37\end{tabular} \\ \cline{2-6} 
                        & \makecell[c]{\sys- \\ compatible}    & \begin{tabular}[c]{@{}c@{}}97.84\\ $\pm$0.49\end{tabular} & \begin{tabular}[c]{@{}c@{}}92.73\\ $\pm$0.47\end{tabular} & \begin{tabular}[c]{@{}c@{}}97.59\\ $\pm$0.49\end{tabular} & \begin{tabular}[c]{@{}c@{}}92.73\\ $\pm$0.47\end{tabular} \\ \hline
\multirow{2}{*}{\makecell{Claude\\-based\\Agent}} & \begin{tabular}[c]{@{}c@{}}APIBank\\ \cite{li2023apibank}\end{tabular} & \begin{tabular}[c]{@{}c@{}}97.34\\ $\pm$0.14\end{tabular} & \begin{tabular}[c]{@{}c@{}}91.33\\ $\pm$0.22\end{tabular} & -                                                      & \begin{tabular}[c]{@{}c@{}}91.33\\ $\pm$0.22\end{tabular} \\ \cline{2-6} 
                        & \makecell[c]{\sys- \\ compatible}    & \begin{tabular}[c]{@{}c@{}}95.79\\ $\pm$1.35\end{tabular} & \begin{tabular}[c]{@{}c@{}}92.33\\ $\pm$0.68\end{tabular} & \begin{tabular}[c]{@{}c@{}}96.99\\ $\pm$0.59\end{tabular} & \begin{tabular}[c]{@{}c@{}}90.88\\ $\pm$1.46\end{tabular} \\ \hline
\end{tabular}
\end{adjustbox}
% \caption{LLM tool-calling accuracy comparison. The ``format'' and ``content'' assess structure adherence and parameter correctness, respectively. The ``adapter'' measures the success rate of transforming into \sys-compatible requests.}
\caption{Service-request accuracy under the standard APIBank format and the \sys-compatible template.}
\label{tab:llm}
\end{table}

% As shown in Table~\ref{tab:llm}, the performance gap between the two formats is negligible, with total accuracy differing by at most $1.81$\% across the two formats.
As shown in Table~\ref{tab:llm}, the \sys-compatible format preserves request-generation accuracy, with the largest observed drop in total accuracy being only $0.45$ percentage points.
% The high format and content accuracy across all agents indicate that the \sys-compatible template does not materially degrade the accuracy of agent-generated tool calls.
This trend is also reflected in the consistently high format and content accuracy across all agents, indicating that agents can reliably generate \sys-compatible tool calls.
The slightly higher content accuracy under \sys stems from the designation of 22 parameters as ARI-owned, such that agents are only required to output ``null'' for these segments, reducing error probabilities. The adapter success rate is slightly below 100\% because agents occasionally generate content exceeding the predetermined maximum length $h$, which prevents the adapter from mapping it into a valid segment. 
% These results confirm that \sys effectively preserves the functional fidelity of LLMs while ensuring compatibility with complex tool-calling queries.
% These results confirm that \sys preserves LLM functional fidelity while ensuring compatibility with complex service-requests workflows.
Our findings demonstrate that integration with our protocol is straightforward, given that even mid-tier LLM agents are fully capable of generating the required requests. 
% Therefore, \sys remains compatible with existing agent tool-calling workflows while preserving request-generation accuracy.
Overall, these results confirm the compatibility of \sys with existing agent tool-calling workflows.

%% file: 6.discussion.tex
\section{Discussion}
\label{sec:discussion}

\parab{Service Adaptation}
In real-world deployment, an agent needs to adopt a \sys skill~\cite{agent_skill} that realizing agent-side \sys protocols when accessing ARI. 
The skill maps agent inputs into template-specified request segments, which serve as inputs to the privacy-preserving query-construction protocol. 
After receiving a response, the skill decrypts it locally and extracts the required fields. 
To handle heterogeneous services, the ARI provisions service-specific request and response templates to the skill and keeps their versions synchronized across both sides. 
Thus, \sys preserves the service-adaptation capability of existing ARIs: agents still program against a unified interface, e.g., a common LLM API that abstracts over provider-specific endpoints, parameters, and return schemas.

\parab{Billing-Field Offset Leakage}
The verifiable billing protocol reveals the byte offset of the attested billing field, because ARI specifies the local response window used by the proof.
% However, the leakage is narrowly scoped: it identifies the location of the pre-defined billing field used for settlement, but does not reveal the contents of the remaining response or the boundaries of other variable-length fields. 
This leakage is limited to the settlement metadata: it does not reveal the remaining response contents or the boundaries of other variable-length fields.
In particular, when the response contains multiple interleaved variable-length fields, one attested field offset does not, by itself, determine the layout of the remaining payload.

\parab{Malicious Security} 
% \fixme{Malicious adversaries can provide arbitrary inputs that does not satisfy the segment format defined in the public template.}
% \sys can be instantiated in the malicious-security model by combining malicious-secure backends with input-validity checks following the garble-then-prove paradigm~\cite{xie2024lightweight}.
% \sys can support the malicious-security model by instantiating each protocol component with malicious-secure mechanisms.
% The handshake already uses malicious-secure authenticated garbled circuits, and the billing protocol relies on standard NIZK soundness against malicious provers.
Extending \sys to the malicious security model requires malicious-security guarantees for all three protocols. Our handshake protocol follows a malicious-secure three-party construction (see Appendix~\ref{sec:appendix:handshake}), and the verifiable billing protocol relies on NIZK soundness against malicious provers. 
% For query-construction, the remaining requirement is to ensure that each party's segment inputs satisfy the template-defined structure before collaborative assembly and encryption.
% For query construction, a malicious party may submit malformed segments, deviate from the assembly or encryption procedure, or produce inconsistent ciphertext/tag outputs. 
% Most such deviations lead to TLS authentication failure, malformed service requests, or denial-of-service; 
% they do not reveal private field values to the other party or let an adversary overwrite an unknown field with a chosen value.
% The remaining leakage is application-level: valid inputs may reveal coarse predicates of low-entropy fields, \eg the agent may learn whether ARI's hidden coupon applies to a chosen item category, but not the coupon string itself.
The remaining semi-honest component is the query construction protocol. 
We clarify that a malicious adversary gains only limited additional advantage even against a semi-honest query construction protocol: malformed or template-nonconforming segments, or incorrect assembly/encryption mainly lead to failed queries or denial of service, \eg through TLS authentication failure or malformed service requests, rather than revealing private field values or enabling an adversary to overwrite unknown fields with chosen values. 
% The remaining component is query construction, which would require adding input well-formedness checks and replacing the semi-honest backend with a malicious-secure backend,  following the garble-then-prove paradigm~\cite{xie2024lightweight}.
% These checks ensure that each party's segment inputs satisfy the template-defined structure before collaborative assembly and encryption.
% A malicious-secure variant would add input validation for template conformance and segment well-formedness, and use a malicious-secure backend for computation correctness, following the garble-then-prove paradigm~\cite{xie2024lightweight}.
% This increases computation and communication due to additional validation and proof overhead.
With additional overhead, it is possible to augment the query construction protocol with malicious-security guarantees with the garble-then-prove paradigm~\cite{xie2024lightweight}, which adds input validation for template conformance and segment well-formedness, and uses a malicious-secure backend for computation correctness. 

%% file: 7.related_work.tex
\section{Related Work}
\label{sec:related}

% \subsection{MCP Ecosystem and Security Risks}

% MCP has rapidly scaled into a large ecosystem with thousands of third-party servers~\cite{wu2025mcpzoo}, but a large-scale measurement \cite{guo2025measurement} shows that most MCP servers are developed by independent parties rather than the downstream tool providers themselves. 
% This decoupled supply chain yields highly uneven implementation quality and introduces substantial security risk.
% \fixme{add OAuth}
% Researchers found that malicious MCP servers can mount active attacks against both the host environment and agent’s decision loop. Li et al.~\cite{li2025we} highlight risks of privilege escalation and misinformation/phishing in MCP servers, while Zhao et al.~\cite{zhao2025mcp} show that malicious servers can perform metadata poisoning and output manipulation. Beyond active attacks, recent studies have further revealed that MCP servers can cause privacy leakage even without compromising functionality. Zhao et al.~\cite{zhao2025mcp} highlight information over-collection, while Yao et al.~\cite{yao2025intentminer} demonstrate intent inference from tool-call traces.
% Consequently, there are two mainstream defense directions: reactive server-side scanning~\cite{xing2025mcp-guard, radosevich2025mcpsafetyaudit, narajala2025securing} and runtime monitoring~\cite{bhatt2025etdi,kumar2025mcp, jing2025mcip}. However, these defenses remain heuristic and reactive, without any cryptographic guarantees for correct tool execution, privacy preservation, or verifiable metering under an untrusted MCP server.

\subsection{ARI Ecosystem and Security Enhancements}
\label{subsec:related:mcp}
ARI has become an intermediary layer for agents to access external services.
% , covering both LLM API Routing and Agentic Tool Routing. 
% For the latter line, MCP~\cite{mcp_document} provides a standardized instantiation and has rapidly evolved into an open ecosystem~\cite{wu2025mcpzoo,guo2025measurement}, making MCP-based deployments a concrete setting for studying ARI-mediated security~\cite{zhao2025mcp}.
For Agentic Tool Routing, 
% For the latter line, 
MCP~\cite{mcp_document} provides a standardized instantiation and has evolved into an open ecosystem~\cite{wu2025mcpzoo}, making MCP-based deployments a concrete setting for studying ARI-mediated security~\cite{zhao2025mcp}.

% ARI suffers from two fundamental trust issues: privacy leakage and lack of integrity guarantees.
Existing ARI security work has attempted to mitigate privacy leakage and integrity risks.
For privacy, the current best practice is to use OAuth~\cite{rfc6749}, which reduces the need to expose long-term service credentials to ARI. However, OAuth does not protect runtime requests or responses when ARI mediates the invocation, and may introduce privilege-abuse or unauthorized-access risks when delegated permissions are overly broad or mishandled~\cite{rfc6819}. To protect runtime privacy, Zhao et al.~\cite{zhao2026anonymization} replace personally identifiable information (PII) with deterministic, type-preserving placeholders before invocation, but their protection is limited to query prompts in LLM API routing. For integrity, existing security mechanisms mainly follow two directions: reactive server-side scanning~\cite{xing2025mcp-guard,radosevich2025mcpsafetyaudit,narajala2025securing}, which inspects ARI implementations before invocation; and runtime monitoring~\cite{bhatt2025etdi,kumar2025mcp,jing2025mcip}, which interposes auditors, gateways, or guard models to inspect interactions.
These mechanisms remain heuristic and require visibility into the ARI implementation or deployment path, making them primarily applicable to open-source or locally deployed ARI.
Overall, existing security enhancement mechanisms cannot provide cryptographic guarantees for request and response privacy, end-to-end integrity, or fair billing under an untrusted ARI.

\subsection{TLS-Oracle}
\label{subsec:related:oracle}
TLS-oracle systems extend a standard TLS connection between a client and an unmodified service provider with a proof mechanism, enabling the client to convince an external verifier of statements about selected response fields
% , such as proving that an authenticated age field is above 18 
(\eg the age field is above 18)
without revealing other information.
Existing TLS-oracle systems largely fall into two categories: notary mode~\cite{zhang2020deco, xie2024lightweight, celi2023distefano, tan2023mpcauth} and proxy mode~\cite{grubbs2022zero, zhang2024zombie, ernstberger2025origo}.
In notary mode, the prover communicates directly with the TLS server and later convinces an external verifier about selected properties of the response.
% However, in ARI-mediated interactions, the notary mechanisms would either force ARI to see plaintext to perform request construction~\cite{zhang2020deco, xie2024lightweight, celi2023distefano}, or fail to support the structural text concatenation required for multi-field query building~\cite{tan2023mpcauth}.
However, in ARI-mediated interactions, notary mechanisms would either force ARI to see plaintext for request construction~\cite{zhang2020deco, xie2024lightweight, celi2023distefano}, or fail to support the structural text concatenation required for multi-field query building~\cite{tan2023mpcauth}.
Alternatively, applying them to ARI-mediated interactions would require proving large transcript portions (\ie the entire query and response), where strict ciphertext alignment incurs prohibitive cryptographic overhead.
Moreover, recent variants rely on restrictive assumptions such as distributed provers~\cite{tan2023mpcauth} or service-provider signatures~\cite{della2026acts}, which do not hold in the ARI setting.
By contrast, proxy-mode TLS-oracle targets network-middlebox settings, where an intermediary relays TLS traffic and checks whether encrypted packets satisfy network policies without learning the plaintext payload~\cite{grubbs2022zero,zhang2024zombie,ernstberger2025origo}.
This design preserves confidentiality, but it confines the intermediary to ciphertext relay, which conflicts with ARI's role in mediating heterogeneous service interfaces and fragmented provider subscriptions.
In particular, ARI must use service-access metadata and support usage-based billing during the invocation, which proxy-mode systems do not provide.
Overall, existing TLS-oracle designs do not support the role-specific interaction semantics required between the agent and ARI, whereas \sys addresses this gap with several ARI-specific designs. 

%% file: 8.conclusion.tex
\section{Conclusion}

% In this paper, we propose \sys, a privacy-preserving and verifiable MCP protocol for agentic AI. 
% \sys enables the agent to validate the identity of tools via an MCP-specific key scheduling without altering the tool’s view. Benefiting from the privacy-preserving query protocol, \sys can prevent private information leakage to the MCP server when invoking downstream tools. Furthermore, \sys incorporates a verifiable billing protocol that allows agents to achieve a transparent payment for services, without disclosing any personal data. We implemented a prototype of \sys and extensively evaluated it on real-world and synthetic API datasets. The experimental results across extensive settings confirm that \sys is superior to general TLS oracles and is readily deployable in real-world LLM ecosystems.

To address privacy leakage and the lack of integrity guarantees in ARI-mediated agentic interactions, we propose \sys, a trust-native agentic routing infrastructure for agentic AI.
\sys introduces an ARI-adapted three-party TLS handshake that enables the agent and ARI to establish a standard TLS session with a service provider, while distributing key material according to role-specific interaction semantics and allowing the agent to validate the intended endpoint.
Building on this handshake, \sys provides a privacy-preserving query-construction protocol that allows the agent and ARI to collaboratively construct the TLS-authenticated request without revealing private inputs to each other.
Finally, \sys incorporates a verifiable billing protocol that preserves the confidentiality and integrity of the agent's received response while supporting existing usage-based settlement.
The experimental results across extensive settings confirm that \sys is readily deployable without any modification to service providers.

%% file: appendix.tex
\section{Zero-Knowledge Arguments}
\label{sec:zkp}

% An argument system for an NP relationship $\mathcal{R}$ is a protocol between a computationally-bounded prover $\mathcal{P}$ and a verifier $\mathcal{V}$. At the end of the protocol, $\mathcal{V}$ is convinced by $\mathcal{P}$ that there exists a witness $w$ such that $(x; w) \in \mathcal{R}$ for some input $x$. We focus on arguments of knowledge which have the stronger property that if the prover convinces the verifier of the statement's validity, then the prover must know $w$.
We give the formal definition of the zero-knowledge argument of knowledge used in this paper. Let $\mathcal{G}$ denote the setup algorithm that generates public parameters $pp$, and assume the NP relation $\mathcal{R}$ is known to both $\mathcal{P}$ and $\mathcal{V}$.

\noindent\textit{Definition} Let $\mathcal{R}$ be an NP relation. A tuple of algorithms $(\mathcal{G}, \mathcal{P}, \mathcal{V})$ is a zero-knowledge argument of knowledge for $\mathcal{R}$ if the following holds.

\begin{itemize}[leftmargin=*]
    \item \textbf{Correctness.} For $pp$ output by $\mathcal{G}(1^\lambda)$ and $(x, w) \in \mathcal{R}$,
    \[
        \langle \mathcal{P}(pp, w), \mathcal{V}(pp) \rangle (x) = 1
    \]
    \item \textbf{Knowledge Soundness.} For any PPT prover $\mathcal{P}^*$, there exists a PPT extractor $\mathcal{X}$ such that given the access to the entire executing process and the randomness of $\mathcal{P}^*$, $\mathcal{X}$ can extract a witness $w$ such that $\text{pp} \leftarrow \mathcal{G}(1^\lambda)$, $\pi^* \leftarrow \mathcal{P}^*(x, \text{pp})$ and $w \leftarrow \mathcal{X}^{\mathcal{P}^*}(\text{pp}, x, \pi^*)$, the following probability is $\text{negl}(\lambda)$:
    \[
        \Pr[(x; w) \notin \mathcal{R} \wedge \mathcal{V}(x, \pi^*, \text{pp}) = 1]
    \]
    \item \textbf{Zero knowledge.} There exists a PPT simulator \simu such that for any PPT algorithm $\mathcal{V}^*$, auxiliary input $z \in \{0,1\}^*$, $(x; w) \in \mathcal{R}$, pp output by $\mathcal{G}(1^\lambda)$, it holds that
    \[
        \text{View}(\langle \mathcal{P}(\text{pp}, w), \mathcal{V}^*(z, \text{pp}) \rangle (x)) \approx \mathcal{S}^{\mathcal{V}^*}(x, z)
    \]
\end{itemize}

We say that $(\mathcal{G}, \mathcal{P}, \mathcal{V})$ is a succinct argument system if the total communication between $\mathcal{P}$ and $\mathcal{V}$ (proof size) are $\text{poly}(\lambda, |x|, \log |w|)$.
In the definition of zero knowledge, $\mathcal{S}^{\mathcal{V}^*}$ denotes that the simulator \simu is given the randomness of $\mathcal{V}^*$ sampled from a polynomial-size space.

\input{algo/fig-handshake}

\section{The Detailed Handshake Protocol $\Pi_\textsf{ths}$}
\label{sec:appendix:handshake}

We present the detailed ARI-adapted Three-party handshake protocol $\Pi_\textsf{ths}$ in Figure~\ref{prot:handshake}. Specifically, in Protocol~\ref{prot:handshake:sub}, $\textsf{IV}_1 = f_{\textsf{H}}(\textsf{IV}_0, \langle X \rangle \oplus \textsf{ipad})$ via a 2PC computation and make it public, while $\textsf{IV}_2 = f_{\textsf{H}}(\textsf{IV}_0,  \langle X \rangle \oplus \textsf{opad})$ remains secret.

% \parab{Subprotocol Details} As illustrated in the lower part of Figure~\ref{prot:handshake}, our protocol relies on two subroutines. They provide the algorithmic realization of our 2PC-optimized key schedule, specifically targeting scenarios where multiple derivative keys share a common parent secret. Subprotocol 1 ($\textsf{HKDF.ExpandOptm}$) is invoked whenever multiple keys originate from a single shared secret $\langle X \rangle$; by pre-computing the intermediate state $PC_X = (IV_1, \langle IV_2 \rangle)$ once, it allows subsequent expansions to reuse these cached values. This strategy is applied systematically throughout the protocol: first during the Key Exchange Phase to derive $\langle\textsf{CHTS}\rangle$, $\langle\textsf{SHTS}\rangle$, and $\langle\textsf{dHS}\rangle$ from $PC_{HS}$, and subsequently during the Application Key Generation Phase to derive $\langle\textsf{CATS}\rangle$ and $\langle\textsf{SATS}\rangle$ from $PC_{MS}$. Following the same principle, Subprotocol 2 ($\textsf{DeriveTKoptm}$) acts as a specialized wrapper that leverages a common pre-computed state (e.g., $PC_{CATS}$ or $PC_{SATS}$) to simultaneously derive both the secret-shared traffic key $\langle \textsf{tk} \rangle$ and the public initialization vector $\textsf{iv}$. By bundling these co-derived parameters into a single logical execution, the protocol minimizes expensive 2PC-assisted $f_{\textsf{H}}$ compressions and ensures that the transport parameters are established with minimal interaction rounds.

\parab{Security Proof Sketch} The security of $\Pi_{\textsf{ths}}$ is established within the Multi-Stage Key Exchange (MSKE) model~\cite{abram2021oblivious}, following the formal framework and security games defined in Appendix E.1 of DiStefano~\cite{celi2023distefano}. Since our protocol maintains the core transcript binding and key dependency structure of the TLS 1.3 handshake, its security properties (e.g., Key Secrecy and Forward Secrecy) can be reduced to the same cryptographic primitives. We refer the readers to the full-version of DiStefano~\cite{celi2023distefano} for the detailed reduction and formal verification of these security properties.

\section{The Detailed Blind Rotate Protocol $\Pi_\textsf{BlindRotate}$}
\label{sec:appendix:br}

The protocol~\ref{prot:br} executes an oblivious left-shift on an XOR-shared vector without disclosing its content or the precise shift distance.
Specifically, the protocol takes an XOR-shared length-$n$ vector $\langle v_0 \rangle$ and an XOR-shared bit decomposition of a secret shift amount $\langle d \rangle_{\mathsf{xor}}$ as inputs. To prevent wrap-around artifacts and handle potential overflow via zero-padding, the vector length is implicitly extended in a shift network bounded by $L \gets \max(k, \lceil \log_2(2n) \rceil)$. The computation then proceeds bit-by-bit through an iterative layer-by-layer multiplexing cascade. At each stage $\ell$, for every element index $i$, the protocol obliviously fetches the unshifted share $\langle x_0 \rangle$ and the conditionally shifted share $\langle x_1 \rangle$ (which evaluates to a zero-share $\langle 0 \rangle$ upon boundary overflow). The state is then updated by invoking the underlying 2-input oblivious multiplexer functionality $\mathcal{F}_{\textsf{MUX}_2}(\langle d_\ell \rangle, \langle x_1 \rangle, \langle x_0 \rangle)$. This ensures that the conditional shift is applied strictly within the encrypted domain if and only if the secret bit $d_\ell = 1$, ultimately delivering a securely rotated, XOR-shared output vector $\langle v_b \rangle$ to the participants while maintaining zero information leakage.

\begin{Protocol}[!t]
    \begin{mdframed}[style=ProtocolFrame, align=center, font=\small]
        
        \begin{enumerate}[label=\textbf{\scriptsize \arabic*}, leftmargin=2.5ex,itemsep=0.25ex]
            \item[] \textbf{Protocol} $\langle v_b \rangle \gets \Pi_\textsf{BlindRotate}(\langle v_0 \rangle, \langle d \rangle_{\mathsf{xor}}, k)$
                  % \vspace{0.1cm}
                  \hrule
                  \vspace{0.1cm}
            \item[] \textit{Input.} An XOR-shared vector $\langle v_0 \rangle$ of length $n$, and an XOR-shared bit
            decomposition
            $\langle d \rangle_{\mathsf{xor}} = (\langle d_0 \rangle, \ldots, \langle d_{k-1} \rangle)$
            of a secret shift amount $d \in \{0,\ldots,2^k-1\}$.
            \item[] \textit{Output.} An XOR-shared vector $\langle v_b \rangle$ representing $v_0$ left-shifted by
            $d$ positions, computed obliviously.
                  \vspace{0.1cm}
                  \hrule
                  \vspace{0.1cm}

            \item\label{itm:brb-extend}
          Let $L \gets \max\!\bigl(k,\lceil \log_2(2n)\rceil\bigr)$ and
          extend $\langle d \rangle_{\mathsf{xor}}$ with zeros to length $L$.
          % \anno{We choose $L \geq \lceil \log_2(2n)\rceil$ so that the shift network can be viewed as operating over a zero-padded length-$2n$ vector, which supports truncating shifts without wrap-around.}
        
          \item\label{itm:brb-init}
          Set $\langle \mathsf{cur} \rangle \gets \langle v_0 \rangle$.
          \anno{Current XOR-shared vector.}
        
          \item\label{itm:brb-loop}
          \textit{for} $\ell = 0$ \textit{to} $L-1$:
        
          \item\label{itm:brb-delta}
          \quad Let $\Delta \gets 2^\ell$.
          \anno{Stage-$\ell$ shift distance.}
        
          \item\label{itm:brb-inner}
          \quad \textit{for} $i = 0$ \textit{to} $n-1$:
        
          \item\label{itm:brb-x0x1}
          \quad\quad Let $\langle x_0 \rangle \gets \langle \mathsf{cur}[i] \rangle$.
          
          % \item \quad\quad Let $\langle x_1 \rangle \gets
            % \begin{cases}
            %   \langle \mathsf{cur}[i+\Delta] \rangle, & \text{if } i+\Delta < n, \\ 
            %   \langle 0 \rangle, & \text{otherwise}.
            % \end{cases}$
          
          % \hfill{\anno{Fetch shifted element (zero-padding on overflow).}}
          \item\label{itm:brb-x1}
            \quad\quad Let $\langle x_1 \rangle \gets
            \left\{
            \begin{array}{ll}
              \langle \mathsf{cur}[i+\Delta] \rangle, & \text{if } i+\Delta < n \\
              % \hfill{\anno{In-range.}}\\
              \langle 0 \rangle, & \text{otherwise}
              % \hfill{\anno{Zero-padding.}}
            \end{array}
            \right.$
        
          \item\label{itm:brb-mux}
          \quad\quad Update $\langle \mathsf{cur}[i] \rangle \gets
            \mathcal{F}_{\textsf{MUX}_2}\ \!\bigl(\langle d_\ell \rangle,\ \langle x_1 \rangle,\ \langle x_0 \rangle\bigr)$.
          % \hfill{\anno{Obliviously apply stage-$\ell$ shift if $d_\ell=1$.}}
        
          \item\label{itm:brb-out}
          Set $\langle v_b \rangle \gets \langle \mathsf{cur} \rangle$.
          \anno{Final XOR-shared, $d$-shifted vector.}
            
        \end{enumerate}
    
    \end{mdframed}
    % \caption{The OT-based Structure-hidden Concatenation.}
    \caption{The blind rotate protocol.}
    \label{prot:br}
\end{Protocol}

\section{The Detailed Integer-to-String Protocol $\Pi_\textsf{I2S}$}
\label{sec:appendix:i2s}

% \begin{figure}[t]
%     \centering
%     \includegraphics[width=\columnwidth]{figure/fig-i2s-big.pdf}
%     \caption{Integer-to-String (I2S) Conversion Protocol $\Pi_\textsf{I2S}$: arithmetic-shared integer to XOR-shared ASCII string and arithmetic-shared length based on GC.}
%     \label{prot:i2s}
% \end{figure}
\input{algo/fig-i2s}

% The protocol~\ref{prot:i2s} aims to securely convert an arithmetic-shared integer into an XOR-shared ASCII string without leaking intermediate values or padding lengths. To achieve this, the two parties execute a multi-stage deterministic circuit pipeline. First, the input integer is resized to prevent potential signed-division or modular wrap-around artifacts. The protocol then iteratively extracts ASCII representations and tracking quotients in an LSD-first manner. To handle variable-length outputs while protecting structural privacy, the circuit dynamically scans the running quotients via $\Pi_\textsf{FindTrueLen}$ to anchor the exact position where the quotient drops to zero, explicitly accommodating the zero-input boundary condition using a privacy-preserving multiplexer. Finally, $\Pi_\textsf{PackDigits}$ aligns the extracted digits into a standard most-significant-digit-first format, dynamically truncating or padding trailing bytes with zeros to match the public upper bound $h$. The resulting structured text string and its exact byte length are securely secret-shared into the XOR and arithmetic domains, respectively.

The protocol~\ref{prot:i2s} securely converts an arithmetic-shared non-negative integer into an XOR-shared ASCII decimal string without revealing the integer value or its decimal length. 
The implementation is a fixed-size Boolean circuit whose loop bounds are public and determined by the maximum digit length $h$. 
First, the input is resized to a working width large enough to avoid wrap-around during division and ASCII encoding. 
Then, $\Pi_\textsf{ExtractDigits}$ performs a fixed $h$-round base-10 decomposition: in each round, it divides the running value by $10$, emits the ASCII-encoded remainder in least-significant-digit-first order, and records the corresponding quotient. 
Next, $\Pi_\textsf{FindTrueLen}$ obliviously scans the quotient array and uses $\mathcal{F}_{\textsf{MUX}_2}$ to commit once to the first position where the quotient becomes zero, while handling the zero-input case by setting the length to $1$. 
Finally, $\Pi_\textsf{PackDigits}$ uses multiplexer-based oblivious selection to reverse the LSD-first digits into MSD-first order and right-pad the unused suffix with \texttt{0x00}. 
Since all loops run for the public bound $h$ and all value-dependent choices are implemented with multiplexers, the protocol reveals neither the input value nor its decimal length beyond the secret-shared outputs.

\input{security-proof}

\section{The Payment Protocol based on X402}
\label{sec:appendix:x402}

The x402 protocol~\cite{x402_escrow_839} is an internet-native payment standard designed for autonomous machine-to-machine transactions via a pre-funded settlement model. Utilizing escrow smart contracts like TrustEngine~\cite{trustengine_sol}, it enables secure fund locking and atomic reconciliation based on post-execution usage evidence. This provides a decentralized, trustless financial foundation for usage-based agentic AI.
% x402 is an internet-native payment protocol for autonomous machine-to-machine transactions, where a client pre-funds a payment session and later settles the final charge based on execution evidence~\cite{x402_escrow_839}. 
% When combined with escrow contracts such as TrustEngine~\cite{trustengine_sol}, this model naturally supports usage-based ARI billing: funds are locked before execution, while the final fee is released only after verifiable billing evidence is produced.

% \sys designs a verifiable billing protocol to enable automated and trustless payment settlement based on the x402 escrow scheme~\cite{x402_escrow_839}. 
Since ARI also adopts similar models where the exact costs are only determinable after execution, we connect the billing proof of \sys to this escrow workflow. 

\parab{Deposit and Lock Phase} To initiate an active session, the agent invokes the prepare phase, prompting the client to commit a security deposit to the TrustEngine~\cite{trustengine_sol} escrow vault covering the maximum anticipated budget. 
The smart contract permanently logs the session metadata and locks the escrowed assets, preventing unilateral asset withdrawal by either party and eliminating potential payment default during tool execution.

\parab{Settle and Distribute Phase} This phase reconciles post-execution consumption with the locked deposit via the cryptographic proof $\pi$.
Upon receiving the tool's response, the agent generates an off-chain NIZK proof $\pi$, demonstrating that the billing metrics were faithfully extracted from the TLS-authenticated ciphertext without privacy leakage. 
The agent submits $\pi$ to the on-chain verification function; the contract verifies the proof, transfers the calculated fee to ARI, and automatically refunds the residual balance back to the agent.

\input{integrate}

\section{Details of the Real-world API Dataset}
\label{sec:appendix:api}

\begin{table}[t]
  \centering
  \caption{The Structure of Real Tool-call APIs.}
  \label{tab:api}
  \begin{adjustbox}{max width=\columnwidth}
  \begin{tabular}{cccccc}
    \toprule
    \textbf{ID} & \textbf{Category} & \textbf{Tool}
    & \multicolumn{2}{c}{\textbf{Query}}
    & \textbf{Response} \\
    \cmidrule(lr){4-5} \cmidrule(lr){6-6}
    & & &
    \textbf{Segments (\#)}
    & \textbf{Length (B)}
    & \textbf{Length (B)} \\
    \midrule
    1  & SCM & Github    & 15 & 488 & 824  \\
    2  & DKR & Google    & 11 & 419 & 713  \\
    3  & DKR & Qdrant    & 17 & 473 & 592  \\
    4  & PM  & Jira      & 7 & 234 & 1074 \\
    5  & PC  & Slack     & 25 & 704 & 556  \\
    6  & PC  & Gmail     & 3 & 382 & 245  \\
    7  & WS  & Tavily    & 37 & 715 & 617  \\
    8  & WS  & Firecrawl & 49 & 747 & 1102 \\
    9  & LLM & OpenAI    & 23 & 640 & 608  \\
    10 & LLM & Anthropic & 13 & 626 & 541  \\
    \midrule
    \textbf{AVG} & - & - & 20 & 542.8 & 687.2 \\
    \bottomrule
  \end{tabular}
  \end{adjustbox}
\end{table}

We curate a dataset comprising 10 real-world tool-call APIs across six categories in Table~\ref{tab:api}: 
% Each category introduces unique structural characteristics to evaluate our protocol's robustness: 
(i) Source Code Management (SCM, e.g., GitHub) exhibits highly nested JSON objects; 
(ii) Database \& Knowledge Retrieval (DKR, e.g., BigQuery, Qdrant) covers high-throughput queries and vector search fields; 
(iii) Project Management (PM, e.g., Jira) features long issue-tracking text responses; 
(iv) Productivity \& Communication (PC, e.g., Slack, Gmail) contains highly private, human-centric metadata; 
(v) Web Search (WS, e.g., Tavily, Firecrawl) requires substantial query parameter stretches; and 
(vi) LLM Inference (LLM, e.g., OpenAI, Anthropic) represents inter-model interactions critical for evaluating billing-related fields. 
This diversity validates our protocols across heterogeneous industrial APIs.

% \parab{Structural Characteristics} The dataset exhibits a wide range of payload sizes to stress-test our protocols. The Query Length varies from 234 bytes to 747 bytes (avg. 542.8 bytes), reflecting the varied complexity of tool parameters. The Response Length ranges from 245 bytes to 1,425 bytes (avg. 723.4 bytes), which validates the scalability of our verifiable billing protocol. 

%% file: algo/fig-handshake.tex
\begin{Protocol}[h]
    % \begin{adjustbox}{width=\linewidth}
    \begin{mdframed}[style=ProtocolFrame, align=center, font=\normalsize,userdefinedwidth=\columnwidth]
        \textbf{Protocol} ARI-Adapted Three-party handshake $\Pi_\textsf{ths}$
        \vspace{0.1cm}
        \hrule
        \vspace{0.1cm}
        \centering
        \begin{tikzpicture}[
            node distance=0.5cm,
            font=\small,
            >=stealth,
            every node/.style={inner sep=1pt},
            box/.style={draw, rectangle, minimum width=1.8cm, minimum height=0.6cm, align=center, font=\bfseries},
            msgblock/.style={align=left},
            centerblock/.style={align=center},
            arrow/.style={->, thin},
            arrowback/.style={<-, thin},
            dashedarrow/.style={->, dashed, procgray},
            dashedarrowback/.style={<-, dashed, procgray}
        ]
        
        % Width settings
        \def\w{6} % Half width of the diagram
        \def\cw{7} % Text width constraint
        
        % Headers
        \node[box] (client) at (-\w+1, 0) {Agent \mcpclient};
        \node[box] (server) at (\w-1, 0) {ARI \mcpserver};
        
        % --- Block 1: Client/Server Hello ---
        \node[msgblock, anchor=north west] (b1) at (-\w, -0.4) {
            $r_{\textsf{A}} \leftarrow\$ \mathbb{Z}^q$, $Z_{\textsf{A}} \leftarrow g^{r_{\textsf{A}}}$ 
        };
        
        \node[align=right, anchor=north east] (b8) at (\w, -0.4) {
            $r_{\textsf{R}} \leftarrow\$ \mathbb{Z}^q, Z_{\textsf{R}} \leftarrow g^{r_{\textsf{R}}}$
        };
        
        % Stage 1
        \coordinate (y_s1) at (0, -1);
        \draw[dotted, thick, procblue] (-\w, -1) -- (\w, -1);
        \node[fill=white, text=procblue, inner sep=2pt] at (0, -1) {\mcpserver sends $\textsf{CKS} \leftarrow Z_{\textsf{A}} + Z_{\textsf{R}}$ to \toolserver};
        % \node[text=procblue, anchor=east] at (\w, -1) {stage 1};

        % Stage 2
        \coordinate (y_s2) at (0, -1.4);
        \draw[dotted, thick, procblue] (-\w, -1.4) -- (\w, -1.4);
        \node[fill=white, text=procblue, inner sep=2pt] at (0, -1.4) {\mcpclient and \mcpserver accept \textsf{SKS} $\leftarrow Y_t$ from \toolserver};
        % \node[text=procblue, anchor=east] at (\w, -1.4) {stage 2};
        
        % --- Block 2: DHE ---
        
        \node[msgblock, anchor=north west] (b1) at (-\w, -1.5) {
            $\textsf{ssk}^\textsf{A} \leftarrow\$ Y_t^{r_c}$ \\
            $\textsf{DHE}^\textsf{A} \leftarrow $ ECtF$(\textsf{ssk}^\textsf{A})$ 
        };
        
        \node[align=right, anchor=north east] (b8) at (\w, -1.5) {
            $\textsf{ssk}^\textsf{R} \leftarrow\$ Y_t^{r_s}$\\
            $\textsf{DHE}^\textsf{R} \leftarrow $ ECtF$(\textsf{ssk}^\textsf{R})$ 
        };
        
        % --- Block 4: HKDFs ---
        \node[centerblock, anchor=north] (hkdf1) at (0, -2.25) {
            $\textsf{HS}^\textsf{A} \oplus \textsf{HS}^\textsf{R} \leftarrow$ HKDF.Extract($\phi$, $\textsf{DHE}^\textsf{A} \oplus \textsf{DHE}^\textsf{R}$) \\

            \textcolor{procgreen}{$\textsf{PC}_{\textsf{HS}} \gets (\textsf{IV}_1,(\textsf{IV}_2^\textsf{A} \oplus \textsf{IV}_2^\textsf{R}) \leftarrow$ PreCompute($\textsf{HS}^\textsf{A} \oplus \textsf{HS}^\textsf{R}$)}\\

            \textcolor{procgreen}{$\textsf{CHTS}^\textsf{A} \oplus \textsf{CHTS}^\textsf{R} \leftarrow$ HKDF.ExpandOptm($\textsf{PC}_{\textsf{HS}}, \text{Label}_1 \| \textsf{H}_0$)} \\
            \textcolor{procgreen}{$\textsf{SHTS}^\textsf{A} \oplus \textsf{SHTS}^\textsf{R} \leftarrow$ HKDF.ExpandOptm($\textsf{PC}_{\textsf{HS}}, \text{Label}_2 \| \textsf{H}_0$)} \\
            \textcolor{procgreen}{$\textsf{dHS}^\textsf{A} \oplus \textsf{dHS}^\textsf{R} \leftarrow$ HKDF.ExpandOptm($\textsf{PC}_{\textsf{HS}}, \text{Label}_3 \| \textsf{H}_1$)}\\
        };
        
        % Stage 3
        \coordinate (y_s1) at (0, -4.6);
        \draw[dotted, thick, procblue] (-\w, -4.6) -- (\w, -4.6);
        \node[fill=white, text=procblue, inner sep=2pt] at (0, -4.6) {\mcpclient accepts $\textsf{CHTS}^\textsf{R}$ and $\textsf{SHTS}^\textsf{R}$ for \mcpserver};
        % \node[text=procblue, anchor=east] at (\w, -4.4) {stage 1};
        
        % Stage 4
        \coordinate (y_s2) at (0, -5);
        \draw[dotted, thick, procblue] (-\w, -5) -- (\w, -5);
        \node[fill=white, text=procblue, inner sep=2pt] at (0, -5) {\mcpserver and \mcpclient accept encrypted \textsf{EE},\textsf{SCV},\textsf{SF} \etc from \toolserver};
        % \node[text=procblue, anchor=east] at (\w, -4.7) {stage 2};

        \node[msgblock, anchor=north west] (b1) at (-\w, -5.2) {
            \textcolor{procgreen}{$\textsf{tk}_{\textsf{chs}} \leftarrow \textsf{DeriveTK(CHTS)}$} \\
            \textcolor{procgreen}{$\textsf{tk}_{\textsf{shs}} \leftarrow \textsf{DeriveTK(SHTS)}$} \\
            \textcolor{procgreen}{$\textsf{fk}_{\textsf{s}} \leftarrow \textsf{DeriveTK(SHTS, Label}_4 \parallel \textsf{H}_{\epsilon})$} \\
            \textbf{abort} if Auth($\text{pk}_{\toolserver}, \text{Label}_{11} \| \textsf{H}_6$, \textsf{SCV}) $\neq$ 1 \\
            \textbf{abort} if \textsf{SF} $\neq$ HMAC(fk$_\textsf{s}, \textsf{H}_7$)
        };
        
        % --- Block 5: HKDFs ---
        \node[centerblock, anchor=north] (hkdf2) at (0, -6.95) {
            $\textsf{MS}^\textsf{A} \oplus \textsf{MS}^\textsf{R} \leftarrow$ HKDF.Extract($\textsf{dHE}^\textsf{A} \oplus \textsf{dHE}^\textsf{R}$,$\phi$) \\
            \textcolor{procgreen}{$\textsf{PC}_{\textsf{MS}} \gets (\textsf{IV}_1,(\textsf{IV}_2^\textsf{A} \oplus \textsf{IV}_2^\textsf{R})) \leftarrow$ PreCompute($\textsf{MS}^\textsf{A} \oplus \textsf{MS}^\textsf{R}$)}\\
            \textcolor{procgreen}{$\textsf{CATS}^\textsf{A} \oplus \textsf{CATS}^\textsf{R} \leftarrow$ HKDF.ExpandOptm($\textsf{PC}_{\textsf{MS}}, \text{Label}_5 \| \textsf{H}_2$)} \\
            \textcolor{procgreen}{$\textsf{SATS}^\textsf{A} \oplus \textsf{SATS}^\textsf{R} \leftarrow$ HKDF.ExpandOptm($\textsf{PC}_{\textsf{MS}}, \text{Label}_6 \| \textsf{H}_2$)} \\
            \textcolor{procgreen}{$\textsf{PC}_\textsf{CATS} \gets (\textsf{IV}_1,(\textsf{IV}_2^\textsf{A} \oplus \textsf{IV}_2^\textsf{R})) \leftarrow$ PreCompute($\textsf{CATS}^\textsf{A} \oplus \textsf{CATS}^\textsf{R}$)}\\
            \textcolor{procgreen}{$(\textsf{tk}_\textsf{capp}^{\textsf{A}} \oplus \textsf{tk}_\textsf{capp}^\textsf{R},\, \textsf{iv}_\textsf{capp}) \leftarrow$ DeriveTKOptm($\textsf{PC}_{\textsf{CATS}}$)} \\
            \textcolor{procgreen}{$\textsf{PC}_{\textsf{SATS}} \gets (\textsf{IV}_1,(\textsf{IV}_2^\textsf{A} \oplus \textsf{IV}_2^\textsf{R})) \leftarrow$ PreCompute($\textsf{SATS}^\textsf{A} \oplus \textsf{SATS}^\textsf{R}$)}\\
            \textcolor{procgreen}{$(\textsf{tk}_\textsf{sapp}^\textsf{A} \oplus \textsf{tk}_\textsf{sapp}^\textsf{R},\, \textsf{iv}_\textsf{sapp}) \leftarrow$ DeriveTKOptm($\textsf{PC}_\textsf{SATS}$)} \\
            
        };

        % Stage 3
        \coordinate (y_s3) at (0, -10.5);
        \draw[dotted, thick, procblue] (-\w, -10.5) -- (\w, -10.5);
        \node[fill=white, text=procblue, inner sep=2pt] at (0, -10.5) {\mcpclient commits $\textsf{tk}_\textsf{sapp}^\textsf{A}$ to \mcpserver};
        % \node[text=procblue, anchor=east] at (\w, -9.8) {stage 3};
        
        % Stage 4
        \coordinate (y_s4) at (0, -10.9);
        \draw[dotted, thick, procblue] (-\w, -10.9) -- (\w, -10.9);
        \node[fill=white, text=procblue, inner sep=2pt] at (0, -10.9) {\mcpclient accepts $\textsf{tk}_\textsf{sapp}^\textsf{R}$ from \mcpserver};
        % \node[text=procblue, anchor=east] at (\w, -10.5) {stage 4};
        
        \node[msgblock, anchor=north west] (b1) at (-\w, -10.9) {
            $\textsf{tk}_{\textsf{sapp}} \leftarrow \textsf{tk}_\textsf{sapp}^\textsf{A} \oplus \textsf{tk}_\textsf{sapp}^\textsf{R}$ \\
            $\textsf{fk}_\textsf{A} \leftarrow \textsf{DeriveTK(CHTS, Label}_4 \parallel \textsf{H}_{\epsilon})$ \\
            \textsf{CF} $\leftarrow$ \textsf{HMAC}($\textsf{fk}_\textsf{A},\textsf{H}_6$)
        };
        
        % Stage 5
        \coordinate (y_s5) at (0, -12.1);
        \draw[dotted, thick, procblue] (-\w, -12.1) -- (\w, -12.1);
        \node[fill=white, text=procblue, inner sep=2pt] at (0, -12.1) {\mcpclient sends \textsf{CF} to \toolserver through \mcpserver };
        % \node[text=procblue, anchor=east] at (\w, -11.5) {stage 5};
        
        % Gray Arrow 1
        \draw[dashedarrow] (-\w, -12.7) -- (\w, -12.7) node[midway, above, text=procgray, font=\footnotesize] {record layer, \mcpclient and \mcpserver encrypt data with $\textsf{tk}_\textsf{capp}^\textsf{A} \oplus \textsf{tk}_\textsf{capp}^\textsf{R}$ by AEAD};
        
        % Gray Arrow 2
        \draw[dashedarrowback] (-\w, -13.2) -- (\w, -13.2) node[midway, above, text=procgray, font=\footnotesize] {record layer, \mcpclient decrypt data with key \textsf{tk}$_{\textsf{sapp}}$ by AEAD};
        
        % blank
        % \node[fill=white, text=procblue, inner sep=2pt] at (0, -13){};
        
        \end{tikzpicture}
    \vspace{0.1cm}

    \end{mdframed}
    % \end{adjustbox}
    
    \caption{The handshake key schedule and message flow in \sys. Blue indicates message transmission; green shows optimizations over \cite{celi2023distefano}.}\label{fig:handshake-flow} 
    \label{prot:handshake}
    
\end{Protocol}

% \end{document}

\begin{Protocol}[!t]
  \begin{mdframed}[style=ProtocolFrame, align=center, font=\small,userdefinedwidth=\columnwidth]
    \begin{enumerate}[label=\textbf{\scriptsize \arabic*}, leftmargin=2.5ex,itemsep=0.25ex]
        
        \item[] \refstepcounter{subprot}\label{subprot:expand-optm} \textbf{Subprotocol \thesubprot} $\langle \textsf{K} \rangle \gets$ \textsf{HKDF.ExpandOptm}$(\textsf{PC},\, \textsf{info})$
              % \vspace{0.1cm}
              \hrule
              % \vspace{0.1cm}
        \item[] \textit{Input.} $\textsf{PC}=(\textsf{IV}_1,\, \textsf{IV}_2^\textsf{A}\oplus \textsf{IV}_2^\textsf{R})$ and context string $\textsf{info}$.
        \item[] \textit{Output.} Secret-shared output $\langle \textsf{K} \rangle = \textsf{K}^\textsf{A} \oplus \textsf{K}^\textsf{R}$.
              \vspace{0.1cm}
              \hrule
              \vspace{0.1cm}
    
        \item\label{itm:expandoptm-precompute} \textbf{if} $\textsf{PC}$ not given: $(\textsf{IV}_1, \textsf{IV}_2^\textsf{A}\oplus \textsf{IV}_2^\textsf{R}) \gets$ \textsf{PreCompute}$(\langle \textsf{X} \rangle)$; 
            \hfill{\anno{Compute once per shared input $\langle X\rangle$ and reuse.}}
        \item\label{itm:expandoptm-expand} $\langle \textsf{K} \rangle \gets$ \textsf{HKDF.Expand}$(\textsf{IV}_1,\, \textsf{IV}_2^\textsf{A}\oplus \textsf{IV}_2^\textsf{R},\, \textsf{info})$.
            \hfill{\anno{2PC-assisted expansion using cached $(\textsf{IV}_1,\textsf{IV}_2)$.}}
    \end{enumerate}
    
    \hrule
    \begin{enumerate}[label=\textbf{\scriptsize \arabic*}, leftmargin=2.7ex,itemsep=0.25ex]
        \item[] \refstepcounter{subprot}\label{subprot:derivetk-optm}%
        \textbf{Subprotocol \thesubprot} $(\langle \textsf{tk} \rangle, \textsf{iv}) \gets$ \textsf{DeriveTKOptm}$(\textsf{PC}_X)$
              % \vspace{0.1cm}
              \hrule
              % \vspace{0.1cm}
        \item[] \textit{Input.} $\textsf{PC}_X=(\textsf{IV}_1,\, \textsf{IV}_2^\textsf{A}\oplus \textsf{IV}_2^\textsf{R})$ precomputed from a shared secret $\langle X\rangle$.
        \item[] \textit{Output.} Secret-shared traffic key $\langle \textsf{tk}\rangle$ and public IV $\textsf{iv}$.
              \vspace{0.1cm}
              \hrule
              \vspace{0.1cm}
        \anno{Wrapper: two expansions over the same $\textsf{PC}_X$.}
        \item\label{itm:derivetkoptm-wrapper}
            $\langle \textsf{tk} \rangle \gets$ \textsf{HKDF.ExpandOptm}$(\textsf{PC}_X,\, \textsf{Label}_{k}\|H_\epsilon)$;\;
        \item  $\textsf{iv} \gets$ \textsf{HKDF.ExpandOptm}$(\textsf{PC}_X,\, \textsf{Label}_{iv}\|H_\epsilon)$. 
    \end{enumerate}
    \end{mdframed}
    \caption{The subprotocols used in the ARI-adapted three-party handshake protocol.}
    \label{prot:handshake:sub}
\end{Protocol}

%% file: algo/fig-i2s.tex
\begin{Protocol}[!t]
  \begin{mdframed}[style=ProtocolFrame, align=center, font=\small]
    \begin{enumerate}[label=\textbf{\scriptsize \arabic*}, leftmargin=2.5ex,itemsep=0.25ex]
      \item[] \textbf{Protocol} $\langle T\rangle_{\mathsf{xor}}, \langle g_{\textsf{out}}\rangle \gets
      \Pi_\textsf{I2S}(\langle g_{\textsf{body}}\rangle, h)$
      \vspace{0.1cm}
      \hrule
      \vspace{0.1cm}

      \item[] \textit{Input.} An arithmetic-shared integer $\langle g_{\textsf{body}}\rangle \in \mathbb{Z}_{2^k}$ and a public bound $h$
      (maximum number of decimal digits).

      \item[] \textit{Output.} An XOR-shared bitstring $\langle T\rangle_{\mathsf{xor}} \in \{0,1\}^{8h}$ encoding an $h$-byte
      ASCII string (MSD-first, right-padded with \texttt{0x00}), and an arithmetic-shared length
      $\langle g_{\textsf{out}}\rangle \in \mathbb{Z}_{2^k}$ equal to the true ASCII length in bytes.

      \vspace{0.1cm} \hrule \vspace{0.1cm}

      \item\label{itm:clg-recon}
      Let $g \gets \langle g_{\textsf{body}}\rangle$ in a working width $w$.
      % \hfill{\anno{Interpret the input shares as a single secret value for computation.}}

      \item\label{itm:clg-width}
      % Set $w \gets \max(k+1, 7)$ and $g \gets \textsf{Resize}(g,w)$.
      % \hfill{\anno{Prevent signed-division artifacts and ensure ASCII range.}}
      $g \gets \textsf{Resize}(g,\max(k+1,7))$
      \anno{Avoid modular wrap-around.}

      \item\label{itm:clg-digits}
      $(\textsf{digit}, \textsf{quot}) \gets \Pi_\textsf{ExtractDigits}(g,h)$.
      \anno{$\textsf{digit}[i]$ are ASCII bytes (LSD-first); $\textsf{quot}[i]$ are running quotients.}

      \item\label{itm:clg-zero}
      $\textsf{isZero} \gets [g == 0]$;\quad $\ell_0 \gets \mathcal{F}_{\textsf{MUX}_2}(\textsf{isZero}, 1, h)$.
      \anno{If $g=0$, the string is \texttt{"0"} (length is $1$).}

      \item\label{itm:clg-len}
      $\langle g_{\textsf{out}}\rangle \gets \Pi_\textsf{FindTrueLen}(\textsf{quot}, h, \ell_0, \textsf{isZero})$.
      \anno{Compute the first position where the quotient becomes $0$.}

      \item\label{itm:clg-pack}
      $\langle T\rangle_{\mathsf{xor}} \gets \Pi_\textsf{PackDigits}(\textsf{digit}, \langle g_{\textsf{out}}\rangle, h)$.
      % \anno{Produce MSD-first ASCII of length, pad remaining bytes with \texttt{0x00}.}
      \anno{Produce MSD-first ASCII and right-pad unused bytes with \texttt{0x00}.}

    \end{enumerate}
  \end{mdframed}
  % \caption{Integer-to-String (I2S): arithmetic-shared integer to XOR-shared ASCII string conversion.}
  \caption{Integer-to-string conversion from an arithmetic-shared integer to an XOR-shared ASCII decimal string.}
  \label{prot:i2s}
\end{Protocol}

%% file: security-proof.tex
\section{Security Proof of $\Pi_{\textsf{query}}$}
\label{sec:appendix:query-proof}

We prove that $\Pi_{\mathsf{query}}$ securely realizes the Query branch of \fsys. We write $\mathcal{E}_{\textsf{A}}$ and $\mathcal{E}_{\textsf{R}}$ for the abstract private field contents of the agent and ARI in Functionality~\ref{func:fsys}.

\parab{The SHC Functionality \(\mathcal{F}_{\mathsf{SHC}}\)}
We first define the ideal functionality for structure-hiding concatenation.
A padded segment is written as \(E=(\theta,\langle s\rangle,\langle g\rangle,h)\), where \(s\in\{0,1\}^{8h}\) is a padded byte string, \(g\le h\) is the true byte length, and \(h\) is the public padding bound.
For a string \(s\) and length \(g\), let \(\mathsf{trim}(s,g)\) denote the first \(g\) bytes of \(s\).

The functionality \(\mathcal{F}_{\mathsf{SHC}}\) takes shares of two adjacent segments $E_l=(\theta_l,\langle s_l\rangle,\langle g_l\rangle,h_l)$, and $E_r=(\theta_r,\langle s_r\rangle,\langle g_r\rangle,h_r)$. It reconstructs \(s_l,s_r,g_l,g_r\), computes $
s_m =
\mathsf{pad}_{h_l+h_r}
(\mathsf{trim}(s_l,g_l) \,\|\, \mathsf{trim}(s_r,g_r))$,$g_m=g_l+g_r,h_m=h_l+h_r$,
and returns fresh shares of $E_m=(\theta_m,\langle s_m\rangle,\langle g_m\rangle,h_m)$ to the parties, where \(\theta_m\) is determined by the public segment descriptors.
The functionality leaks only the public descriptors and padding bounds \((\theta_l,\theta_r,h_l,h_r)\), but not \((s_l,s_r,g_l,g_r)\).

\parab{Lemma 1}
Assuming the underlying 2PC primitive securely evaluates the structure-hiding compaction function in the semi-honest model, \(\Pi_{\mathsf{SHC}}\) securely realizes \(\mathcal{F}_{\mathsf{SHC}}\).

\parab{Proof}
Consider a semi-honest adversary corrupting either party.
All local values held by the corrupted party are either its input shares, public descriptors, or output shares.
The only operation depending on the hidden boundary \(g_l\) and private strings \(s_l,s_r\) is the secure 2PC evaluation of the compaction function.
By the security of the underlying 2PC primitive, this interaction can be simulated from the corrupted party's input and output shares and the public bounds.
The output shares returned by \(\mathcal{F}_{\mathsf{SHC}}\) are freshly randomized shares of the same merged segment, and are therefore distributed identically to the real protocol outputs.
Hence, the real and ideal executions are indistinguishable.

\parab{Metadata Preparation}
Functionality~\ref{func:fsys} abstracts the query inputs as party-owned private field contents $\mathcal{E}_{\textsf{A}}$ and $\mathcal{E}_{\textsf{R}}$. 
As described in Section~\ref{subsec:query:ours}, $\Pi_{\mathsf{query}}$ represents these abstract fields as initial padded segments under the public template $\mathcal T$. We denote this protocol-level representation by
$\{E\} \leftarrow \mathsf{Encode}_{\mathcal T}(\mathcal{E}_{\textsf{A}},\mathcal{E}_{\textsf{R}})$.
Here, each segment follows the padded secret-shared representation defined in Section~\ref{subsec:query:ours}, and any derived metadata segment is computed by the corresponding secure subprotocol. Thus, $\{E\}$ is computable from $\mathcal {T},\mathcal{E}_{\textsf{A}},\mathcal{E}_{\textsf{R}}$ and the parties' secret-sharing randomness; it is not an additional input to the ideal functionality.

\parab{Assembly Plan} As described in Section~\ref{subsec:query:optm}, \(\Pi_{\mathsf{query}}\) computes an assembly plan
$\Omega \leftarrow \mathsf{Plan}(\mathcal T)$
using only the public template.
The plan specifies the order in which adjacent segments are concatenated.
Since \(\mathsf{Plan}\) depends only on public segment descriptors and padding bounds, \(\Omega\) is independent of either party's private field contents and hidden lengths.
Therefore, \(\Omega\) is only a public execution schedule and introduces no leakage beyond \(\mathcal T\).
Any valid plan preserves the abstract assembled request \(\mathsf{Assemble}(\mathcal T,\mathcal{E}_{\textsf{A}},\mathcal{E}_{\textsf{R}})\), because it changes only the order of semantically equivalent adjacent concatenations.

% \begin{theorem}
\parab{Theorem 1} Assume that the subprotocols for derived metadata segments securely realize their corresponding ideal functionalities, that \(\Pi_{\mathsf{SHC}}\) securely realizes \(\mathcal{F}_{\mathsf{SHC}}\), and that \(\Pi_{\mathsf{2PC\text{-}AEAD}}\) securely realizes \(\mathcal{F}_{\mathsf{AEAD}}\), all in the semi-honest model.
Then \(\Pi_{\mathsf{query}}\) securely realizes the Query branch of \fsys against any semi-honest adversary corrupting either the agent or ARI.

\parab{Proof} Let \(\mathcal A\) corrupt \(P_i\in\{\mcpclient,\mcpserver\}\).
We construct a simulator \simu that receives \(P_i\)'s input, the public template \(\mathcal T\), and the ideal output \((\hat Q,\sigma_Q)\).
\simu computes \(\Omega\leftarrow\mathsf{Plan}(\mathcal T)\).
For public slots and slots owned by the corrupted party, \simu follows the real protocol.
For honest-owned slots, it samples the corrupted party's string and length shares uniformly from the corresponding sharing domains, subject only to the public padding bounds in \(\mathcal T\).
For derived metadata segments, \simu invokes the simulators of their corresponding secure subprotocols.
For each \(\Pi_{\mathsf{LDC}}\) step, \simu applies the same local update as in the real protocol.
For each \(\Pi_{\mathsf{SHC}}\) step, it invokes 
the simulator guaranteed by Lemma~1.
% the simulator guaranteed by the security of \(\Pi_{\mathsf{SHC}}\) with respect to \(\mathcal{F}_{\mathsf{SHC}}\).
Finally, it simulates \(\Pi_{\mathsf{2PC\text{-}AEAD}}\) by the simulator for \(\mathcal{F}_{\mathsf{AEAD}}\), programmed with \((\hat Q,\sigma_Q)\). We prove indistinguishability by hybrids.

\parab{Hyb0}
This is the real execution of \(\Pi_{\mathsf{query}}\).

\parab{Hyb1}
Replace the corrupted party's shares of honest-owned initial segments with uniformly random shares over the same domains and public padding bounds.
This hybrid is identically distributed to \(\mathsf{Hyb}_0\) by the privacy of XOR and additive secret sharing.

\parab{Hyb2}
Replace each subprotocol for derived metadata with its ideal functionality and simulator.
By the assumed security of these subprotocols, \(\mathsf{Hyb}_2\) is computationally indistinguishable from \(\mathsf{Hyb}_1\).

\parab{Hyb3}
Replace every invocation of \(\Pi_{\mathsf{SHC}}\) with \(\mathcal{F}_{\mathsf{SHC}}\) and its simulator.
The \(\Pi_{\mathsf{LDC}}\) steps remain unchanged because they are local deterministic computations over shares and public offsets.
By Lemma~1, \(\mathsf{Hyb}_3\) is computationally indistinguishable from \(\mathsf{Hyb}_2\).

\parab{Hyb4}
Replace the final \(\Pi_{\mathsf{2PC\text{-}AEAD}}\) execution with \(\mathcal{F}_{\mathsf{AEAD}}\) and its simulator, programmed with the ideal output \((\hat Q,\sigma_Q)\).
By the assumed security of \(\Pi_{\mathsf{2PC\text{-}AEAD}}\), \(\mathsf{Hyb}_4\) is computationally indistinguishable from \(\mathsf{Hyb}_3\).

\(\mathsf{Hyb}_4\) is exactly the simulated view of the Query branch of \fsys. The proof is completed.

%% file: integrate.tex
\section{The Integrated Protocol $\Pi_{\textsf{sys}}$ }
\label{sec:integrated}

\input{algo/fig-integrate}

% Taking all parts together, the final \sys protocol is shown % as shown in 
% Protocol~\ref{prot:integrated}.
% By integrating the ARI-specific three-party handshake, the privacy-preserving query protocol, and the verifiable billing protocol, we present the complete \sys protocol $\Pi_{\sys}$ in Figure~\ref{prot:integrated}.
Figure~\ref{prot:integrated} presents the complete protocol $\Pi_{\sys}$, which composes the three protocols.
% the ARI-specific three-party handshake, privacy-preserving query construction, response forwarding, and verifiable billing. 
The protocol first establishes the TLS states between three parties, while splitting the provider-facing application secret between the client and ARI through a commit-and-reveal step. Given a query template $\mathcal{T}$ and encrypted segments ${E}$, the client and ARI derive a query plan and jointly construct an encrypted request $(\hat{Q},\sigma_Q)$, which the ARI forwards to the tool server. After receiving the encrypted response $(\hat{R},\sigma_R)$, the ARI relays it to the client, who decrypts the response locally using the reconstructed application secret. Finally, the client extracts the billing witness from the plaintext response, while the ARI extracts the corresponding public inputs from the encrypted transcript, enabling the client to prove the pre-declared billing metric $\varphi$ and value $v$ without revealing the response content to the ARI.

\parab{Security Proof Sketch}
The security of $\Pi_{\textsf{sys}}$ follows by sequentially composing the security of its three phases. The handshake phase realizes the Handshake branch of \fsys by the semi-honest security of $\Pi_{\mathsf{ths}}$ proven in Appendix~\ref{sec:appendix:handshake}. The query phase implements the Query branch of \fsys, as proven secure by $\Pi_{\mathsf{query}}$ in Appendix~\ref{sec:appendix:query-proof}.

For the billing phase, $\Pi_{\mathsf{bill}}$ is instantiated by a standard NIZK proof system for the relation specified by the billing circuit in \S~\ref{sec:response}. Completeness ensures that an honest agent can prove the declared billing value extracted from the authenticated response. Zero-knowledge hides all response contents except the public statement, including the declared value and the billing descriptor. Knowledge soundness ensures that any accepting proof must correspond to a valid witness, namely a response plaintext and key share that are cryptographically bound to the TLS-authenticated ciphertext and satisfy the claimed billing relation. Therefore, $\Pi_{\mathsf{bill}}$ securely realizes the Response branch of \fsys.

Since the three phases are executed sequentially and the only values passed between phases are exactly the session state and public outputs, the standard sequential composition argument yields that the integrated protocol realizes \fsys against a semi-honest adversary.

%% file: algo/fig-integrate.tex
\begin{Protocol}[!t]
  % \begin{mdframed}[style=ProtocolFrame, align=center, font=\footnotesize]
  \begin{mdframed}[style=ProtocolFrame, align=center, font=\small,userdefinedwidth=\columnwidth]
    \begin{enumerate}[label=\textbf{\scriptsize \arabic*}, leftmargin=1.5ex,itemsep=0.2ex]
      \item[] \textbf{Protocol} $(\hat{Q}, \sigma_Q, \hat{R}, \sigma_R, v, \pi,\textsf{acc}) \gets \Pi_{\textsf{sys}}(\mathcal{T},\varphi,\{E\})$
      % \vspace{0.1cm}
      \hrule
      % \vspace{0.1cm}

      \item[] \textit{Input.} Query template $\mathcal{T}$, billing metrics $\varphi$, segments $\{E\}$.
      \item[] \textit{Output.} Encrypted request $(\hat{Q}, \sigma_Q)$ and response $(\hat{R},\sigma_R)$, the proof $\pi$ for the pre-declared $\varphi$ and its value $v$, and the verification result \textsf{acc}.

      \vspace{0.1cm} \hrule \vspace{0.1cm}

      \item\label{itm:int-hs}
      \textbf{(Handshake)}
      % $(\textsf{tr}, tk_{capp}, tk_{sapp}) \gets \Pi_{\textsf{hs}}(\mcpclient,\mcpserver,\toolserver)$;\;
      % $\mcpclient:\ \textsf{Verify}(\textsf{cert}_{\toolserver},\textsf{tr})=1$.
      % \anno{$\textsf{tr}$: transcript.}
      $(\textsf{tk}_\textsf{chs},\textsf{tk}_\textsf{shs},\langle \textsf{tk}_\textsf{capp}\rangle, \langle\textsf{tk}_\textsf{sapp}\rangle) \gets \Pi_{\textsf{ths}}$(\mcpclient,\mcpserver,\toolserver);\;
      $\mcpclient:\ \textsf{Auth}$($\textsf{id}_{\toolserver}$,$\textsf{tk}_\textsf{shs})=1$.
      % \anno{$\textsf{tr}$: transcript.}

      \item\label{itm:int-share-sapp}
      \textbf{(Commit})
      $\mcpclient: r \gets \mathbb{Z}^l$,$\mathsf{com}_A \gets \textsf{H}(\textsf{tk}^\textsf{A}_\textsf{sapp}, r)$.
      $\mcpclient \rightarrow \mcpserver:\mathsf{com}_A$;\;
      $\mcpserver \rightarrow \mcpclient:\textsf{tk}^\textsf{R}_\textsf{sapp}$. $\mcpclient: \textsf{tk}_\textsf{sapp} \gets \textsf{tk}^\textsf{A}_\textsf{sapp} \oplus \textsf{tk}^\textsf{R}_\textsf{sapp}$.

      \item\label{itm:int-q}
      \textbf{(Query)} \mcpclient and \mcpserver: $\Omega \gets \Pi_{\textsf{plan}}(\mathcal{T})$;
      $(\hat{Q},\sigma_Q) \gets \Pi_\textsf{Query}$($\mathcal{T}$, $\{E\}$, $\Omega$, $\langle\textsf{tk}_\textsf{capp}\rangle)$;\ $\mcpserver \rightarrow \toolserver:(\hat{Q},\sigma_Q)$.

      \item\label{itm:int-hat-r}
      \textbf{(Response)}
      $\toolserver:(\hat{R},\sigma_R) \gets \textsf{Tool}(\hat{Q})$;\;
      $\toolserver \rightarrow \mcpserver: (\hat{R},\sigma_R)$;\;
      $\mcpserver\ \rightarrow \mcpclient:\ (\hat{R},\sigma_R)$;\;
      $\mcpclient:\ R \gets \Pi_{\textsf{tls-dec}}\big(\textsf{tk}_\textsf{sapp},\hat{R} \big)$.
      % ($k_{\textsf{tgt}}$, $v_{\textsf{tgt}}$, $i_{\textsf{colon}}$, $d$, $\textsf{hdr}_\textsf{p}$)$\gets \Pi_\textsf{preprocess}(R)$.

      \item\label{itm:int-zkx}
      \textbf{(Setup)}
      $(\textsf{pk},\textsf{vk})\leftarrow\text{Gen}(\textsf{Ckt})$;\;
      % $\textsf{zkx} \gets$ ($\textsf{tk}^\textsf{R}_\textsf{sapp}$, $\mathsf{com}_A$, $\hat{R}$, $k_{\textsf{tgt}}$, $v_{\textsf{tgt}}$, $i_{\textsf{colon}}$, $d$, $\textsf{hdr}_\textsf{p}$,$r$);\;
      % $\textsf{zkw} \gets (\textsf{tk}^\textsf{A}_\textsf{sapp})$.
      $\mathbb{W} = (\textsf{tk}^\textsf{A}_\textsf{sapp}, R_{\text{local}})$;\;
      $\mathbb{X} = (\textsf{tk}^\textsf{R}_\textsf{sapp}$, $\hat{R}$, $r$, $\varphi$, $v$, $\omega$, $d_t$, $j$, $k$, $\textsf{com}_A)$; $\mcpclient:\ (\mathbb{W},\mathbb{X})\leftarrow\Pi_\textsf{ext}(R)$; $\mcpserver:\ (\mathbb{X})\leftarrow\Pi_\textsf{ext}(\hat{R})$.

      \item\label{itm:int-proof}
      \textbf{(Billing)}
      $\mcpclient:\ \pi \gets \Pi_{\textsf{bill}}.\textsf{Prove}(\textsf{pk},\mathbb{W},\mathbb{X})$;\;
      $\mcpclient \rightarrow \mcpserver:\ \pi$;\;
      $\mcpserver:\ \textsf{acc}\gets \Pi_{\textsf{bill}}.\textsf{Verify}(\textsf{vk},\pi,\mathbb{X})$.

    \end{enumerate}
  \end{mdframed}
  \caption{The Integrated \sys Protocol.}
  \label{prot:integrated}
\end{Protocol}

% \end{document}

%% file: ref.bib
@article{wang2024survey,
  title={A survey on large language model based autonomous agents},
  author={Wang, Lei and Ma, Chen and Feng, Xueyang and Zhang, Zeyu and Yang, Hao and Zhang, Jingsen and Chen, Zhiyuan and Tang, Jiakai and Chen, Xu and Lin, Yankai and others},
  journal={Frontiers of Computer Science},
  volume={18},
  number={6},
  pages={186345},
  year={2024},
  publisher={Springer}
}

@misc{tlsadoption, 
    title = {Adoption \& Usage},
    url={https://radar.cloudflare.com/adoption-and-usage?#tls-12-vs-tls-13-vs-quic}, 
    author={Cloudflare}, 
    year={2025}, 
    month={Oct} 
}

@misc{rfc9000,
    series =    {Request for Comments},
    number =    9000,
    howpublished =  {RFC 9000},
    publisher = {RFC Editor},
    doi =       {10.17487/RFC9000},
    url =       {https://www.rfc-editor.org/info/rfc9000},
    author =    {Jana Iyengar and Martin Thomson},
    title =     {{QUIC: A UDP-Based Multiplexed and Secure Transport}},
    pagetotal = 151,
    year =      2021,
    month =     may,
}

@misc{rfc8446,
    series =    {Request for Comments},
    number =    8446,
    howpublished =  {RFC 8446},
    publisher = {RFC Editor},
    doi =       {10.17487/RFC8446},
    url =       {https://www.rfc-editor.org/info/rfc8446},
    author =    {Eric Rescorla},
    title =     {{The Transport Layer Security (TLS) Protocol Version 1.3}},
    pagetotal = 160,
    year =      2018,
    month =     aug,
}

@misc{rfc6749,
    series =    {Request for Comments},
    number =    6749,
    howpublished =  {RFC 6749},
    publisher = {RFC Editor},
    doi =       {10.17487/RFC6749},
    url =       {https://www.rfc-editor.org/info/rfc6749},
    author =    {Dick Hardt},
    title =     {{The OAuth 2.0 Authorization Framework}},
    pagetotal = 76,
    year =      2012,
    month =     oct,
}

@misc{rfc6819,
    series =    {Request for Comments},
    number =    6819,
    howpublished =  {RFC 6819},
    publisher = {RFC Editor},
    doi =       {10.17487/RFC6819},
    url =       {https://www.rfc-editor.org/info/rfc6819},
    author =    {Torsten Lodderstedt and Mark McGloin and Phil Hunt},
    title =     {{OAuth 2.0 Threat Model and Security Considerations}},
    pagetotal = 71,
    year =      2013,
    month =     jan,
}

@inproceedings{rathee2020cryptflow2,
  title={Cryptflow2: Practical 2-party secure inference},
  author={Rathee, Deevashwer and Rathee, Mayank and Kumar, Nishant and Chandran, Nishanth and Gupta, Divya and Rastogi, Aseem and Sharma, Rahul},
  booktitle={Proceedings of the 2020 ACM SIGSAC conference on computer and communications security},
  pages={325--342},
  year={2020}
}

@misc{openrouter_2023, 
    title={OpenRouter: One API for Any Model. Access all major models through a single, unified interface.},
    url={https://openrouter.ai/}, 
    author={OpenRouter}, 
    year={2023} 
}

@article{xing2025mcp-guard,
  title={MCP-Guard: A Defense Framework for Model Context Protocol Integrity in Large Language Model Applications},
  author={Xing, Wenpeng and Qi, Zhonghao and Qin, Yupeng and Li, Yilin and Chang, Caini and Yu, Jiahui and Lin, Changting and Xie, Zhenzhen and Han, Meng},
  journal={arXiv preprint arXiv:2508.10991},
  year={2025}
}

@article{narajala2025securing,
  title={Securing genai multi-agent systems against tool squatting: A zero trust registry-based approach},
  author={Narajala, Vineeth Sai and Huang, Ken and Habler, Idan},
  journal={arXiv preprint arXiv:2504.19951},
  year={2025}
}

@article{bhatt2025etdi,
  title={{Etdi: Mitigating Tool Squatting and Rug Pull Attacks in Model Context Protocol (MCP) by Using Oauth-enhanced Tool Definitions and Policy-based Access Control}},
  author={Bhatt, Manish and Narajala, Vineeth Sai and Habler, Idan},
  journal={arXiv preprint arXiv:2506.01333},
  year={2025}
}

@article{kumar2025mcp,
  title={Mcp guardian: A security-first layer for safeguarding mcp-based ai system},
  author={Kumar, Sonu and Girdhar, Anubhav and Patil, Ritesh and Tripathi, Divyansh},
  journal={arXiv preprint arXiv:2504.12757},
  year={2025}
}

@article{dowling2021cryptographic,
  title={A cryptographic analysis of the TLS 1.3 handshake protocol},
  author={Dowling, Benjamin and Fischlin, Marc and G{\"u}nther, Felix and Stebila, Douglas},
  journal={Journal of Cryptology},
  volume={34},
  number={4},
  pages={37},
  year={2021},
  publisher={Springer}
}

@inproceedings{zhang2020deco,
  title={Deco: Liberating web data using decentralized oracles for tls},
  author={Zhang, Fan and Maram, Deepak and Malvai, Harjasleen and Goldfeder, Steven and Juels, Ari},
  booktitle={Proceedings of the 2020 ACM SIGSAC Conference on Computer and Communications Security},
  pages={1919--1938},
  year={2020}
}

@article{celi2023distefano,
  title={Distefano: Decentralized infrastructure for sharing trusted encrypted facts and nothing more},
  author={Celi, Sof{\'\i}a and Davidson, Alex and Haddadi, Hamed and Pestana, Gon{\c{c}}alo and Rowell, Joe},
  journal={Cryptology ePrint Archive},
  year={2023}
}

@inproceedings{xie2024lightweight,
  title={Lightweight authentication of web data via garble-then-prove},
  author={Xie, Xiang and Yang, Kang and Wang, Xiao and Yu, Yu},
  booktitle={Proceedings of the 33rd USENIX Conference on Security Symposium},
  pages={1957--1974},
  year={2024}
}

@article{ernstberger2025origo,
  title={Origo: Proving provenance of sensitive data with constant communication},
  author={Ernstberger, Jens and Lauinger, Jan and Wu, Yinnan and Gervais, Arthur and Steinhorst, Sebastian},
  journal={Proceedings on Privacy Enhancing Technologies},
  year={2025}
}

@inproceedings{grubbs2022zero,
  title={{Zero-Knowledge Middleboxes}},
  author={Grubbs, Paul and Arun, Arasu and Zhang, Ye and Bonneau, Joseph and Walfish, Michael},
  booktitle={31st USENIX Security Symposium (USENIX Security 22)},
  pages={4255--4272},
  year={2022}
}

@inproceedings{zhang2024zombie,
  title={{Zombie: Middleboxes that Don’t snoop}},
  author={Zhang, Collin and DeStefano, Zachary and Arun, Arasu and Bonneau, Joseph and Grubbs, Paul and Walfish, Michael},
  booktitle={21st USENIX Symposium on Networked Systems Design and Implementation (NSDI 24)},
  pages={1917--1936},
  year={2024}
}

@article{lauinger2025janus,
  title={Janus: Fast Privacy-Preserving Data Provenance For TLS},
  author={Lauinger, Jan and Ernstberger, Jens and Finkenzeller, Andreas and Steinhorst, Sebastian},
  journal={Proceedings on Privacy Enhancing Technologies},
  year={2025}
}

@misc{emp-toolkit,
      author = {Xiao Wang and Alex J. Malozemoff and Jonathan Katz},
      title = {{EMP-toolkit: Efficient MultiParty computation toolkit}},
      howpublished = {\url{https://github.com/emp-toolkit}},
      year={2016}
    }

@misc{boringssl,
      author = {Google},
      title = {{BoringSSL}},
      howpublished = {\url{https://boringssl.googlesource.com/boringssl}},
      year={2026}
    }

@misc{gnark,
  author       = {Gautam Botrel and
                  Thomas Piellard and
                  Youssef El Housni and
                  Ivo Kubjas and
                  Arya Tabaie},
  title        = {Consensys/gnark: v0.14.0},
  month        = jun,
  year         = 2025,
  publisher    = {Zenodo},
  version      = {v0.14.0},
  doi          = {10.5281/zenodo.5819104},
  url          = {https://doi.org/10.5281/zenodo.5819104}
}

@article{gabizon2019plonk,
  title={Plonk: Permutations over lagrange-bases for oecumenical noninteractive arguments of knowledge},
  author={Gabizon, Ariel and Williamson, Zachary J and Ciobotaru, Oana},
  journal={Cryptology ePrint Archive},
  year={2019}
}

@misc{li2023apibank,
      title={API-Bank: A Benchmark for Tool-Augmented LLMs}, 
      author={Minghao Li and Feifan Song and Bowen Yu and Haiyang Yu and Zhoujun Li and Fei Huang and Yongbin Li},
      year={2023},
      eprint={2304.08244},
      archivePrefix={arXiv},
      primaryClass={cs.CL}
}

@article{mo2025livemcpbench,
  title={Livemcpbench: Can agents navigate an ocean of mcp tools?},
  author={Mo, Guozhao and Zhong, Wenliang and Chen, Jiawei and Chen, Xuanang and Lu, Yaojie and Lin, Hongyu and He, Ben and Han, Xianpei and Sun, Le},
  journal={arXiv preprint arXiv:2508.01780},
  year={2025}
}

@article{yao2025intentminer,
  title={IntentMiner: Intent Inversion Attack via Tool Call Analysis in the Model Context Protocol},
  author={Yao, Yunhao and Wang, Zhiqiang and Cheng, Haoran and Cheng, Yihang and Du, Haohua and Li, Xiang-Yang},
  journal={arXiv preprint arXiv:2512.14166},
  year={2025}
}

@article{zhao2025mcp,
  title={When mcp servers attack: Taxonomy, feasibility, and mitigation},
  author={Zhao, Weibo and Liu, Jiahao and Ruan, Bonan and Li, Shaofei and Liang, Zhenkai},
  journal={arXiv preprint arXiv:2509.24272},
  year={2025}
}

@article{li2025we,
  title={We urgently need privilege management in mcp: A measurement of api usage in mcp ecosystems},
  author={Li, Zhihao and Li, Kun and Ma, Boyang and Xu, Minghui and Zhang, Yue and Cheng, Xiuzhen},
  journal={arXiv preprint arXiv:2507.06250},
  year={2025}
}

@misc{Pipedream, 
    title={Pipedream: The AI toolkit for integrations},
    url={https://mcp.pipedream.com/}, 
    year={2026} 
}

@misc{Zapier, 
    title={Zapier: Automate AI Workflows, Agents, and Apps},
    url={https://zapier.com/},
    year = {2026}
}

@misc{litellm,
    title={LiteLLM: Al Gateway to provide model access, fallbacks and spend tracking across 100+ LLMs.}, 
    url={https://www.litellm.ai/}, 
    year= {2026}
}

@misc{mcp_document,
    title = {Understanding Remote MCP Servers},
    url = {https://modelcontextprotocol.io/docs/develop/connect-remote-servers},
    author = {Anthropic},
    year = {2025}
}

@article{wu2025mcpzoo,
  title={MCPZoo: A Large-Scale Dataset of Runnable Model Context Protocol Servers for AI Agent},
  author={Wu, Mengying and Chen, Pei and Hong, Geng and An, Aichao and Chen, Jinsong and Wan, Binwang and Pan, Xudong and Dai, Jiarun and Yang, Min},
  journal={arXiv preprint arXiv:2512.15144},
  year={2025}
}

@article{radosevich2025mcpsafetyaudit,
  title={Mcp safety audit: Llms with the model context protocol allow major security exploits},
  author={Radosevich, Brandon and Halloran, John},
  journal={arXiv preprint arXiv:2504.03767},
  year={2025}
}

@inproceedings{jing2025mcip,
  title={Mcip: Protecting mcp safety via model contextual integrity protocol},
  author={Jing, Huihao and Li, Haoran and Hu, Wenbin and Hu, Qi and Heli, Xu and Chu, Tianshu and Hu, Peizhao and Song, Yangqiu},
  booktitle={Proceedings of the 2025 Conference on Empirical Methods in Natural Language Processing},
  pages={1177--1194},
  year={2025}
}

@inproceedings{yao1982protocols,
  title={Protocols for secure computations},
  author={Yao, Andrew C},
  booktitle={23rd annual symposium on foundations of computer science (sfcs 1982)},
  pages={160--164},
  year={1982},
  organization={IEEE}
}

@inproceedings{yao1986generate,
  title={How to generate and exchange secrets},
  author={Yao, Andrew Chi-Chih},
  booktitle={27th annual symposium on foundations of computer science (Sfcs 1986)},
  pages={162--167},
  year={1986},
  organization={IEEE}
}

@inproceedings{ishai2003extending,
  title={Extending oblivious transfers efficiently},
  author={Ishai, Yuval and Kilian, Joe and Nissim, Kobbi and Petrank, Erez},
  booktitle={Annual International Cryptology Conference},
  pages={145--161},
  year={2003},
  organization={Springer}
}

@misc{x402_escrow_839,
  author       = {Faulk, Lloyd and others},
  title        = {{x402 Escrow Scheme for Pre-funded, Usage-Based Payments}},
  howpublished = {GitHub Issue \#839, coinbase/x402},
  year         = {2025},
  url          = {https://github.com/coinbase/x402/issues/839}
}

@misc{trustengine_sol,
  author       = {Lloyd Faulk},
  title        = {{TrustEngine.sol: Escrow Smart Contract for Pre-funded, Usage-Based Payments}},
  howpublished = {\url{https://github.com/cartdotfun/evm-contracts/blob/main/contracts/TrustEngine.sol}},
  year         = {2025},
}

@misc{agent_skill,
    author = {Anthropic},
    title = {{Agent Skills}},
    howpublished = {\url{https://platform.claude.com/docs/en/agents-and-tools/agent-skills/overview}},
    year = {2025},
}

@article{chang2026overcoming,
  title={Overcoming the Retrieval Barrier: Indirect Prompt Injection in the Wild for LLM Systems},
  author={Chang, Hongyan and Bao, Ergute and Luo, Xinjian and Yu, Ting},
  journal={arXiv preprint arXiv:2601.07072},
  year={2026}
}

@article{wang2025mcptox,
  title={MCPTox: A benchmark for tool poisoning attack on real-world MCP servers},
  author={Wang, Zhiqiang and Gao, Yichao and Wang, Yanting and Liu, Suyuan and Sun, Haifeng and Cheng, Haoran and Shi, Guanquan and Du, Haohua and Li, Xiangyang},
  journal={arXiv preprint arXiv:2508.14925},
  year={2025}
}

@inproceedings{keller2015actively,
  title={Actively secure OT extension with optimal overhead},
  author={Keller, Marcel and Orsini, Emmanuela and Scholl, Peter},
  booktitle={Annual Cryptology Conference},
  pages={724--741},
  year={2015},
  organization={Springer}
}

@inproceedings{abram2021oblivious,
  title={{Oblivious TLS via Multi-party Computation}},
  author={Abram, Damiano and Damg{\aa}rd, Ivan and Scholl, Peter and Trieflinger, Sven},
  booktitle={Cryptographers’ Track at the RSA Conference},
  pages={51--74},
  year={2021},
  organization={Springer}
}

@inproceedings{fang2024freeauth,
  title={FreeAuth: Privacy-Preserving Email Ownership Authentication with Verification-Email-Free},
  author={Fang, Yijia and Li, Bingyu and Xiao, Jiale and Qin, Bo and Zhang, Zhijintong and Wu, Qianhong},
  booktitle={2024 Annual Computer Security Applications Conference (ACSAC)},
  pages={336--352},
  year={2024},
  organization={IEEE}
}

@inproceedings{della2026acts,
  title={ACTS: Attestations of Contents in TLS Sessions},
  author={Della Monica, Pierpaolo and Visconti, Ivan and Vitaletti, Andrea and Zecchini, Marco and others},
  booktitle={Proceedings of the Network and Distributed System Security (NDSS) Symposium 2026},
  year={2026}
}

@inproceedings{tan2023mpcauth,
  title={Mpcauth: Multi-factor authentication for distributed-trust systems},
  author={Tan, Sijun and Chen, Weikeng and Deng, Ryan and Popa, Raluca Ada},
  booktitle={2023 IEEE symposium on security and privacy (S\&P)},
  pages={829--847},
  year={2023},
  organization={IEEE}
}

@inproceedings{angel2026coral,
  title={Coral: Fast succinct non-interactive zero-knowledge CFG proofs},
  author={Angel, Sebastian and Celi, Sof{\'\i}a and Margolin, Elizabeth and Mishra, Pratyush and Sander, Martin and Woods, Jess},
  year={2026},
  booktitle={IEEE Symposium on Security and Privacy (S\&P)}
}

@inproceedings{albrecht2016mimc,
  title={MiMC: Efficient encryption and cryptographic hashing with minimal multiplicative complexity},
  author={Albrecht, Martin and Grassi, Lorenzo and Rechberger, Christian and Roy, Arnab and Tiessen, Tyge},
  booktitle={International Conference on the Theory and Application of Cryptology and Information Security},
  pages={191--219},
  year={2016},
  organization={Springer}
}

@article{zhao2026anonymization,
  title={Anonymization-Enhanced Privacy Protection for Mobile GUI Agents: Available but Invisible},
  author={Zhao, Lepeng and Zou, Zhenhua and Li, Shuo and Liu, Zhuotao},
  journal={arXiv preprint arXiv:2602.10139},
  year={2026}
}
